\documentclass[10pt,twocolumn,letterpaper]{article}

\usepackage[pagenumbers]{cvpr} 

\usepackage{graphicx}
\usepackage{amsmath}
\usepackage{amssymb}
\usepackage{booktabs}
\usepackage{comment}
\usepackage{multirow}
\usepackage{xcolor}

\usepackage[pagebackref,breaklinks,colorlinks]{hyperref}

\usepackage[capitalize]{cleveref}
\crefname{section}{Sec.}{Secs.}
\Crefname{section}{Section}{Sections}
\Crefname{table}{Table}{Tables}
\crefname{table}{Tab.}{Tabs.}

\def\cvprPaperID{13138} 
\def\confName{CVPR}
\def\confYear{2025}

\begin{document}

\title{LUCES-MV: A Multi-View Dataset for Near-Field Point Light Source Photometric Stereo}

\author{ Fotios Logothetis\\
Toshiba Europe \\
Cambridge UK\\
{\tt\small fotios.logothetis} \\ {\tt\small @toshiba.eu}
\and
 Ignas Budvytis\\
 Independent researcher\\
 Cambridge UK\\
{\tt\small ignas.budvytis} \\{ \tt\small @gmail.com}\\
\and
Stephan, Liwicki\\
Toshiba Europe \\
Cambridge UK\\
{\tt\small stephan.liwicki} \\ { \tt\small  @toshiba.eu}
\and
Roberto Cipolla \\
University of Cambridge\\
Cambridge UK\\
{\tt\small rc10001@cam.ac.uk}
}
\maketitle

\begin{abstract}

The biggest improvements in Photometric Stereo (PS) field has recently come from adoption of differentiable volumetric rendering techniques such as NeRF or Neural SDF achieving impressive reconstruction error of 0.2mm on DiLiGenT-MV benchmark. However, while there are sizeable datasets for environment lit objects such as Digital Twin Catalogue (DTS), there are only several small Photometric Stereo datasets which often lack challenging objects (simple, smooth, untextured) and practical, small form factor (near-field) light setup.

To address this, we propose LUCES-MV, the first real-world, multi-view dataset designed for near-field point light source photometric stereo. Our dataset includes 15 objects  with diverse materials, each imaged under varying light conditions from an array of 15 LEDs positioned 30 to 40 centimeters from the camera center. To facilitate transparent end-to-end evaluation, our dataset provides not only ground truth normals and ground truth object meshes and poses but also light and camera calibration images.

We evaluate state-of-the-art near-field photometric stereo algorithms, highlighting their strengths and limitations across different material and shape complexities. LUCES-MV dataset offers an important benchmark for developing more robust, accurate and scalable real-world Photometric Stereo based 3D reconstruction methods.

\end{abstract}

\section{Introduction}

\begin{figure}[ht]
\centering
 \includegraphics[width=\columnwidth]{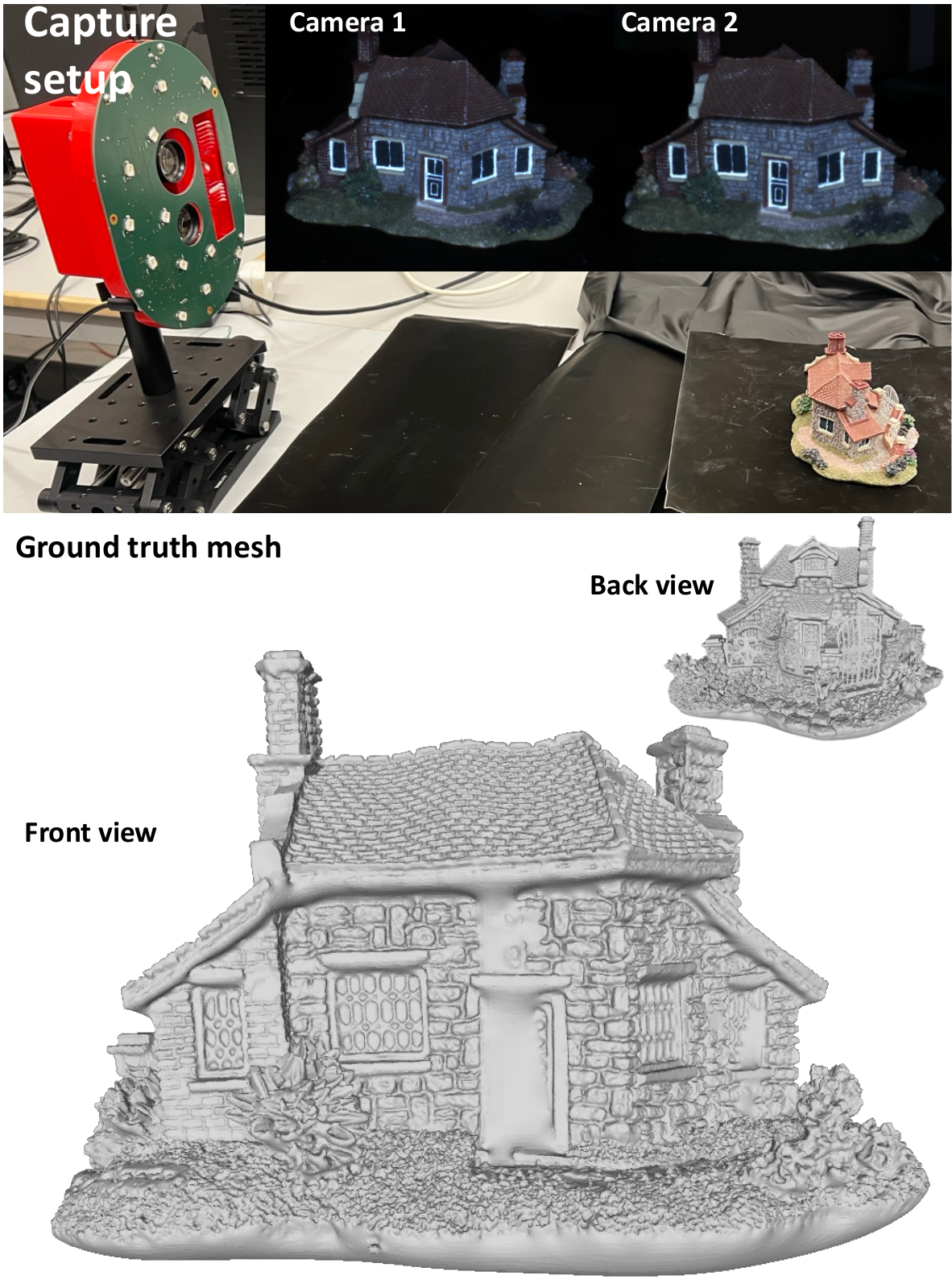}
 \caption{This figure illustrates our capture setup (top half). Capture device is enclosed by the red casing which houses a 15 LED lights and two RGB cameras. The object (a miniature \textit{House} in this case) is placed on a bluetooth controllable turntable covered by non-reflective material on the right. The average images from 15 lights of a particular view is shown on the top right corner. The ground truth mesh of the object scan is illustrated on the bottom of the figure. The shape of the \textit{House} is far more complex than the shape of any object in current Photometric Stereo datasets such as DiLiGenT-MV~\cite{LiZWSDT20}. See examples of other challenging objects in Figure~\ref{fig:objects}.}
 \label{fig:intro}
\end{figure}

\begin{figure}[t]
\centering
 \includegraphics[width=0.9\columnwidth]{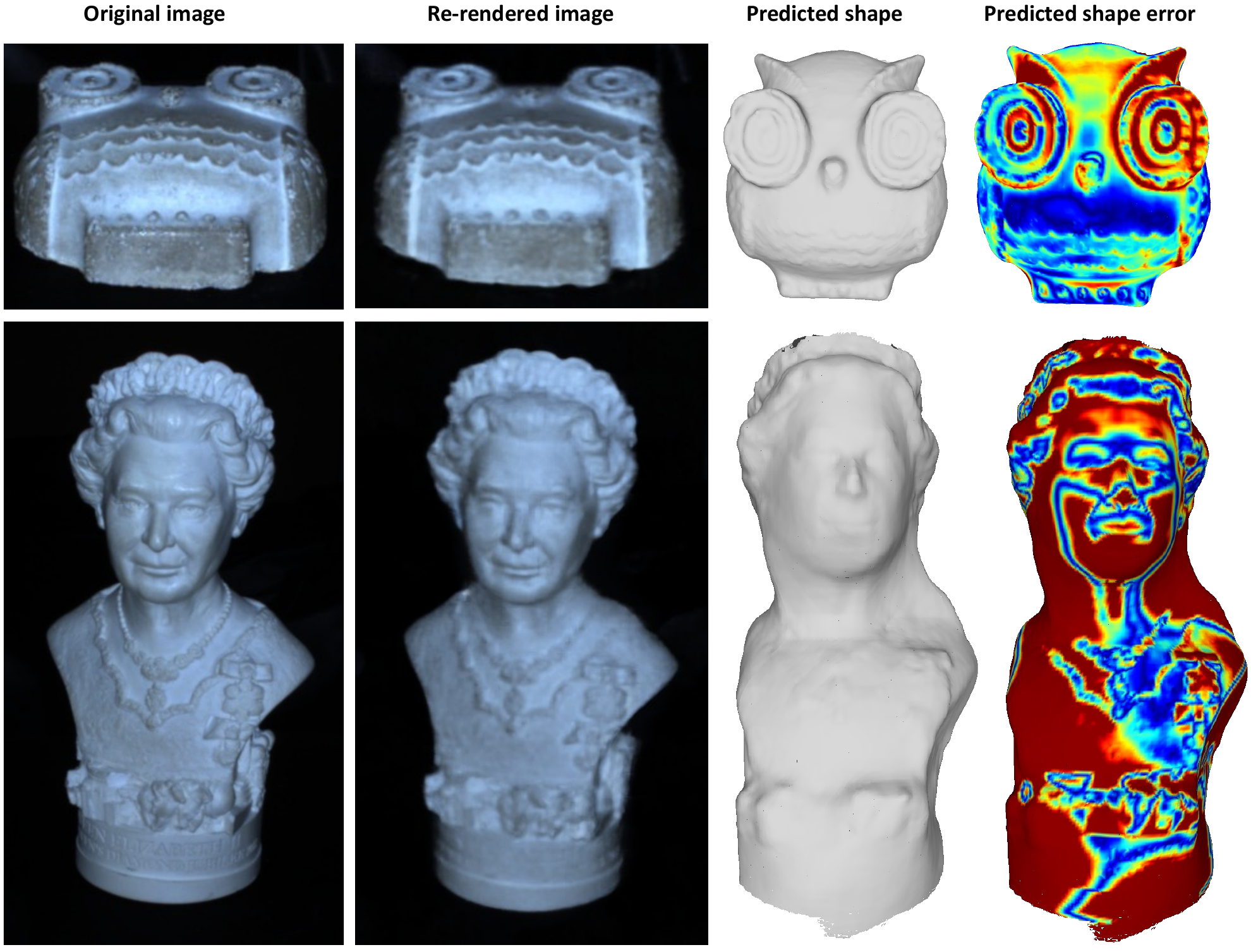}
 \caption{In this figure we show a reconstruction result of Neuralangelo~\cite{li2023neuralangelo} using 36 pairs (average images of 15 light images are used) of stereo views. Note that the re-rendered images match the original very well (PSNR of 32.8 and 38.8 for \textit{Owl} and  \textit{Queen} respectively). However, the predicted shape is significantly worse for  \textit{Queen} (1.68mm) than for  \textit{Owl} (0.44mm). This should be compared to results of RNb-NeuS~\cite{BrumentRNb24} ( \textit{Owl} - 0.37mm,  \textit{Queen} - 0.21mm), also shown qualitatively in Figure~\ref{fig:mainshape}.
 }
 \label{fig:neuralangelo}
\end{figure}

The quality of object shape reconstruction has improved rapidly over recent years. Major milestones include the adoption of neural shape representations such as NeRF~\cite{mildenhall2020nerf} and DeepSDF~\cite{deepsdf19} and subsequent structure in differentiable rendering process (e.g. Ref-NeRF~\cite{refnerf}, NeILF++~\cite{zhang2023neilfpp},  NERO~\cite{liu2023nero}). While improving neural re-rendering accuracy is believed to also improve the accuracy of the recovered shape, it is most often not evaluated. In fact, in practice we see a significant discrepancy between state-of-the-art neural rendering methods and photometric stereo methods targeted explicitly at shape reconstruction as illustrated in Figure~\ref{fig:neuralangelo} (to be contrasted with Figure~\ref{fig:mainshape}). 


In the Photometric Stereo community, shape reconstruction first utilized na\"ive volumetric rendering~\cite{Kaya2021}, and evolved to leverage normals as a loss for shape estimation~\cite{psnerf,cao2023supernormal}. Most recently both the rendering and normals signals, are employed for relatively accurate shape reconstruction~\cite{LogothetisWACV2025,Brahimi_2024_CVPR}. The true impact of this evolution, however, is hidden by evaluations on 
the saturated DiLiGenT-MV~\cite{LiZWSDT20} (as shown by~\cite{LogothetisWACV2025}), which is almost always the only benchmark used. We note, there is a clear need for new data, given the significant progress and promise of photometric stereo as powerful approach to shape reconstruction.


\begin{figure*}[h]
\centering
\includegraphics[width=0.9\textwidth]{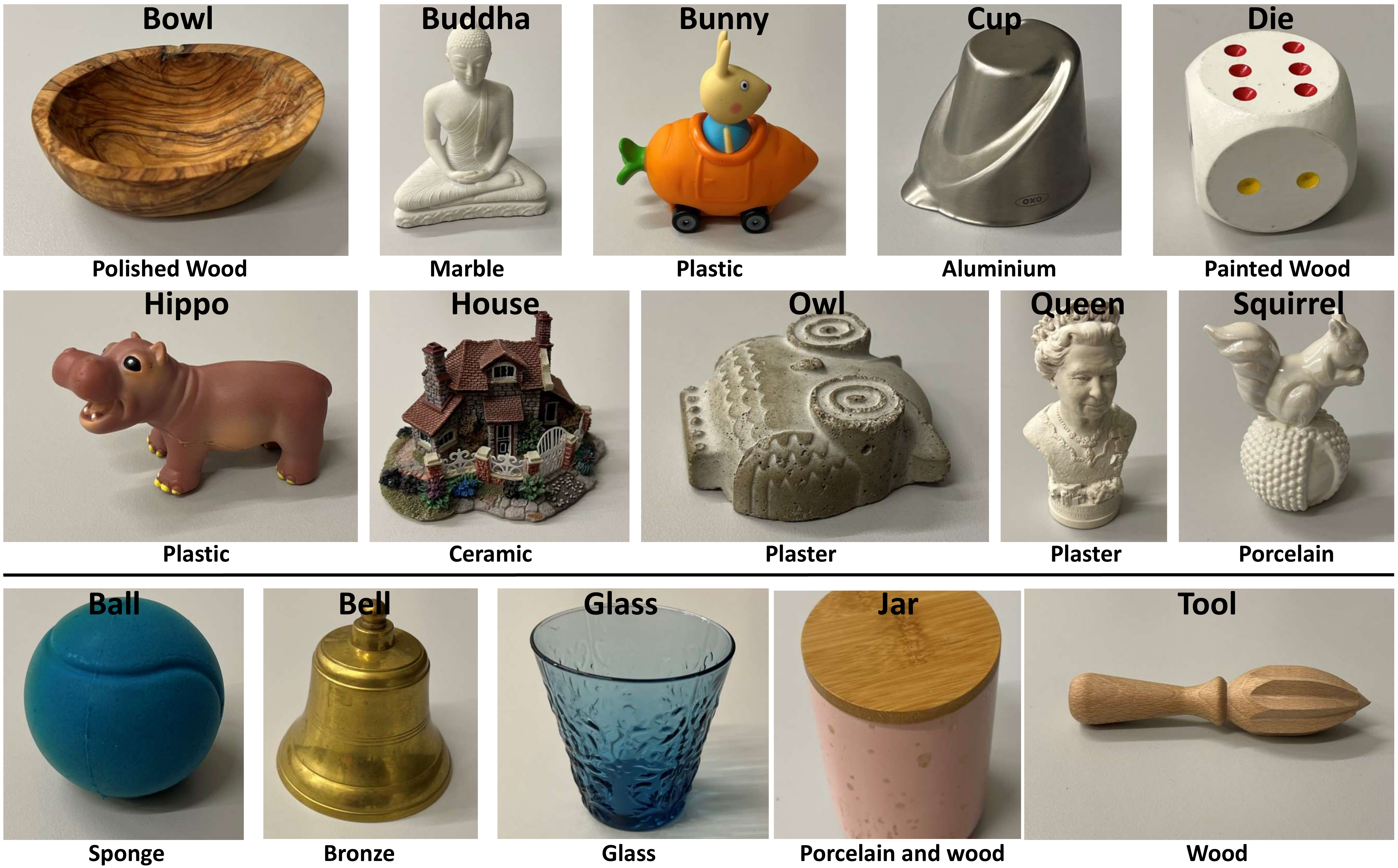}
\caption{The top part of figure shows iPhone images of 10 objects (first two rows) in the LUCES-MV dataset for which not only the photometric stereo images and ground truth meshes but also ground truth poses for 12 frames (6 pairs of stereo images) are provided. The final 5 objects (row 3) has all information except the alignment between images and ground truth meshes. We hope that future methods of unposed or weakly posed Photometric Stereo will be able to tackle the estimation and evaluation of the shape of these objects. }
\label{fig:objects}
\end{figure*}


To alleviate this problem, we propose our LUCES-MV dataset, an extension of LUCES~\cite{Mecca21luces}. Our dataset contains 10 objects\footnote{In fact we provide Photometric Stereo images from 36 viewpoints of which 6 are provided with accurate ground truth pose and ground truth meshes for 15 objects from original LUCES~\cite{Mecca21luces} benchmark. However, due to rotational symmetry, we do not include pose alignments or shape reconstruction accuracies for 4 objects:  \textit{Ball},  \textit{Bell},  \textit{Jar},  \textit{Tool} and we also add one more transparent object -  \textit{Glass}. We hope future unposed Photometric Stereo methods to be able to tackle such data. See more information about those objects in the supplementary material.} of varying materials and diverse shapes. Compared to DiLiGenT-MV~\cite{LiZWSDT20} we improve in several aspects:
\begin{itemize}
\item We capture high resolution ($2080 \times 1552$ vs $612\times 512$) images under true near-field lighting conditions (objects are within 30-40cm).
\item  We provide stereo pairs for all views to foster the improvement binocular photometric stereo methods~\cite{Logothetis24WACV}.
\item  We add a challenging metal object (\textit{Cup}) as well as objects with highly complex shapes such as \textit{House} or concave \textit{Bowl} (see Figure~\ref{fig:objects}).
\item We provide all captured images in its raw format (to encourage PS methods to avoid using demosaiced images affected by processing artifacts) and also provide light and camera calibration images so that complete recalibration can be performed, if required (common complaint~\cite{LogothetisWACV2025} with DiLiGenT-MV~\cite{LiZWSDT20} benchmark).
\end{itemize}

Our contributions are as follows: (a) we introduce a novel near-field multi-view PS dataset of 10 posed and 5 unposed objects (b) evaluate key SOTA algorithms for single view, binocular and multi-view Photometric Stereo, (c) identify the future challenges for the PS research.
The dataset and the evaluation of the methods is publicly available 
\footnote{\url{https://drive.google.com/drive/folders/1634yweYUpLvNPC1qEG8hRpmhtxVtFrLi?usp=sharing}}.

The rest of this paper is structured as follows. Section~\ref{sec:relatedworks} discusses the key approaches and datasets used in Photometric Stereo community. Section~\ref{sec:capsetup} describes our data capture process. Whereas Sections~\ref{sec:experiments} and~\ref{sec:results} describes the experiments performed and results obtained respectively.

\section{Related Work}
\label{sec:relatedworks}

In this section we review key single view and multi-view photometric stereo, neural surface reconstruction methods and datasets.\\
\noindent
\textbf{Single view photometric stereo.}  Early PS approaches \cite{Wolff1994ratio, Agrawal2006,MeccaFalcone2013,ChandrakerAK07,VogiatzisH10 } have employed classical optimizations for obtaining 2D normal maps and thus their applicability has been limited in objects of mostly diffuse reflection. More recently, deep learning-based approaches such as \cite{ikehata2018cnn,logothetis2021pxnet,YakunLearning2023} for the far-field problem and~\cite{santo2017deep,Logothetis22, guo2022edgepreserving} for the near field, have significantly improved the single view normal estimation accuracy. Additionally, large synthetic datasets in  conjunction with transformer models also allow tacking the weakly uncalibrated setting in works like \cite{chen2019SDPS_Net,li2022selfps,li2022neural,yang2022snerf}.  Finally,  \cite{Ikehata2022} introduced the universal PS concept with further improvements in \cite{ikehata2023sdmunips}  and \cite{hardyunips}.  \\ 
\noindent
\textbf{Multi-view photometric stereo.} Obtaining accurate shape from a single view has been challenging as well as ill-posed in many cases. Multiview PS methods have traditionally been employed to solve this issue by using information from multiple view and multiple lights.  Classical MVPS approaches applicable on diffuse reflectance objects include \cite{park17mvps,Logothetis19,NiessnerZIS13,ZollhoferDIWSTN15} and \cite{LiZWSDT20} which is applicable to general materials.

\noindent
\textbf{Neural surfaces.} Just like in the single view case, the introduction of deep methods, especially through the use of neural surfaces have revolutionized the multi-view reconstruction problem. The first seminal works are NeRF ~\cite{mildenhall2020nerf}, IDR~\cite{yariv2020multiview}, VolSDF~\cite{yariv2021volume} and NeuS~\cite{wang2021neus}
More recent methods that have advanced the sophistication of the neural rendering pipeline include Ref-NeRF~\cite{refnerf}, NERO~\cite{liu2023nero} and NeILF++~\cite{zhang2023neilfpp} and Neuralangelo~\cite{li2023neuralangelo} which has shown impressive reconstruction quality on large MV datasets.

The first Neural MVPS works \cite{Kaya2021,kaya2022uncertainty,kaya2023multi,Zhao_2023_ICCV,psnerf} have used 2D feature (normals, alebdo) rendering in order to maximize consistency with the output of single-view PS methods. These has been the case  even for the most recent methods including Supernormal~\cite{cao2023supernormal}, which renders patches of normals and RNb-NeuS \cite{BrumentRNb24} which uses normal and albedo maps to render virtual diffuse images. 

Very few methods have attempted to re-render the original PS images including the binocular photometric stereo method \cite{Logothetis24WACV}, Brahimi et. al~\cite{Brahimi_2024_CVPR} and the most recent  NPLMV-PS~\cite{LogothetisWACV2025}. We speculate that the lack of availability of calibrated MVPS data is responsible and that LUCES-MV dataset will push the research frontier forward.

\noindent
\textbf{Photometric Stereo datasets.} PS real datasets have been very limited for a long time with early methods including synthetic data evaluations and only qualitative real  \cite{MeccaQLC2016,logothetis2017semi}.  Some early works including some real data include \cite{Alldrin2008}, \cite{xiong2014shading}  and \cite{queau2015edge} however non of these data are extensive enough to be considered a trusted benchmark. Some early synthetic datasets include SculpturePS~\cite{santo2017deep}, CyclePS~\cite{ikehata2018cnn} and BlobbyPS~\cite{chen2018ps}.

The first widely used PS dataset of significant scale is DiLiGenT \cite{Shi_2016_CVPR} containing 10 objects light from 96 approximately directional lights. Although the image resolution was limited to $612\times512$px, DiLiGenT has been the gold standard benchmark for signle view PS since.  Further extension have significantly improved the single view PS data availability adding a lot more materials in  \cite{Ren_2022_CVPR}, high accuracy planar objects  \cite{wang2023diligent} and translucency \cite{Guo_2024_CVPR}. 
In addition, DiLiGenT-MV \cite{LiZWSDT20} is the only multi-view extension that however only contains 5 objects in the original $612\times512$ low resolution. One limitation of the DiLiGenT* datasets is that they are all in a far-field setting, containing approximate orthographic cameras and approximate directional lights. This minimizes the potential of evaluation near field methods. To mitigate this issue, the LUCES \cite{Mecca21luces} and LUCES-stereo \cite{Logothetis24WACV} datasets were introduced containing near cameras and point light sources. This work extends Luces* to the  MVPS setting.

Finally, note that there are extensive multi-view stereo datasets such as DTU \cite{jensen2014large}, Tanks and Temples and \cite{knapitsch2017tanks} and DTC~\cite{projectaria_dtc_dataset} by Meta, 
however, the last of controlled and varied illumination on these datasets makes PS reconstruction impossible.

\section{Data Capture}
\label{sec:capsetup}

\begin{figure*}[ht]
\centering
 \includegraphics[width=0.85\textwidth]{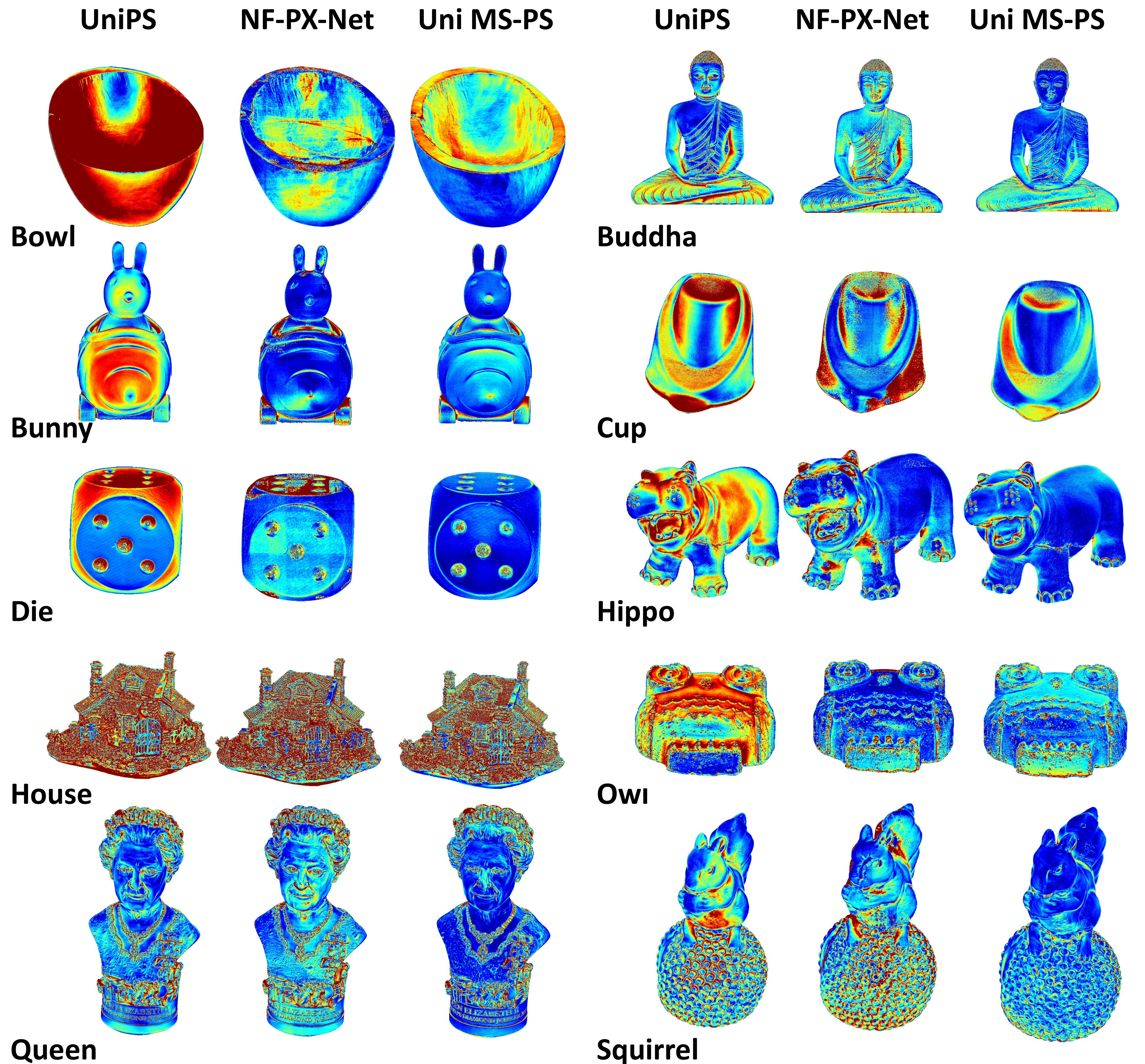}

 \caption{This figure shows normal error map predictions for three normal estimation methods. Uni MS-PS~\cite{hardyunips} significantly outperforms other methods. See corresponding quantitative results in Table~\ref{tab:normalresults}. Dark red color corresponds to an angluar error of 15 degrees and dark blue - to zero.}
 \label{fig:normals}
\end{figure*}

This section gives an overview of the calibration procedure, data capture as well as ground truth capture procedure. It also provides an overview of the contents of the dataset.

\subsection{Photometric Stereo Data Capture}

\paragraph{The Photometric Stereo setup.} Our photometric capture setup (see Figure \ref{fig:intro}) consists of a stereo pair of Flea3 FL3-U3-32S2C-CS 1/2.8” Color USB 3.0 pointgray cameras (resolution $2080\times1552$ px, mounting 8mm lenses ) rigidly attached to a custom circuit board containing 15 Golden Dragon OSRAM LEDs. The approximate dimentions of the circuit board are 14cm$\times$18cm with the camera baseline being 4.5cm. Objects are positioned on a computer controlled turntable around 40cm away from the capture device. Black absorbing material is used on the surface of the turntable and around the setup in order to minimise the effect of intereflections. The whole setup is completely automatic and allows repeatable capture of any object (with the turntable rotation accuracy of around $0.1^o$. Black, polyurethane-coated nylon fabric is used to cover the background.

Camera focus, aperture and exposure were kept constant to make sure the calibration was consistent for the whole dataset. In particular, the aperture was set to the minimum to best approximate a pinhole, and no analogue gain with long exposure (0.6s) was used to minimise the effect of camera noise. All other camera prepossessing (white-balance etc) was also turned off. In addition, to further improve signal to noise ratio (as the maximum precision of the cameras is 10bits per pixel), for each light, 4 raw images were captured. In addition, even though the setup was in a completely dark room, for each view, an `ambient' image was also captured as this is used to calibrate the zero level of each camera pixel.

All raw data including calibration have been saved and will be made available.

\subsection{Calibration}

\paragraph{Camera Intrinsics.} Calibration pattern from \url{https://calib.io/} is used to create checkerboard images. The intrinsics are estimated using the OpenCV calibration toolbox. 7 views of the checkerboard, 60 rotations each are captured giving a total of 420 checkerboard images. The calibration re-projection error of $0.2$px is obtained. Images are wrapped to compensate ration distortion, place camera center in middle of image and equalize focal lengths.

\paragraph{Point Light Calibration.}

The standard approach to point light source Photometric Stereo (\cite{Mecca2014near, queau2018led}) is employed. An anisotropic angular dissipation as well as inverse square distance attenuation is assumed. Thus if the light source position and orientation are  $\mathbf{P}$ and $\hat{\mathbf{D}}$ respectively, the lighting vector $\mathbf{L}$  of a world point $\mathbf{X}$ is computed as $\mathbf{L}=\mathbf{P}-\mathbf{X}$. For an angular dissipation factor $\mu$, the angular dissipation is $(\hat{\mathbf{L}}  \cdot \hat{\mathbf{D}})^{\mu }$, where $\hat{\mathbf{L}}=\frac{\mathbf{L}}{||\mathbf{L}||}$ is the normalized lighting vector. Combining the above quantities with an intrinsic brightness $\Phi$ and the inverse distance factor $1/||\mathbf{L}||^2$, gives the overall point attenuation (of the 3D world point $\mathbf{X}$) as:

\begin{equation}
\label{eq:attenuation}
a(\mathbf{X})= \Phi \frac{( \hat{\mathbf{L}} \cdot \hat{\mathbf{D}})^{\mu }}{||\mathbf{L}||^2}.
\end{equation}

The estimation of parameters $(\mathbf{P},\mathbf{\hat{D}},\mu ,\Phi )$ of Equation~\ref{eq:attenuation} is achieved using captures of diffuse reflectance calibration target (see supplementary material). Indeed, images of the target should satisfy a diffuse irradiance equation:

\begin{equation}
\label{eq:irad}
\mathbf{I}= {\Phi} a \rho {\hat{\mathbf{L}}} \cdot\hat{\mathbf{N}}.
\end{equation}
\noindent
with $\rho=0.99$ being the albedo according to the material data-sheet. Surface normal $\mathbf{\hat{N}}$ and 3D position of the planar target is easily estimated by performing manual segmentation to find the edges (and subsequently the corners) of the planar pattern followed by PnP-based pose estimation. Simple differentiable rendering (using Keras of Tensorflow v2.0) Equation~\ref{eq:irad} allows recovery of all calibration parameters in a similar manner as in \cite{Mecca21luces}. 13 stereo pair views (total 26 views x 15 lights) images are used achieving re-rendering error of $0.01$.

\subsection{3D Ground Truth Capture}

We are re-using the ground truth scans from the original LUCES \cite{Mecca21luces} dataset with an additional object: \textit{Glass}. These were made with 2 scanners. Ground truth meshes for \textit{Bowl}, \textit{Buddha}, \textit{House}, \textit{Owl}, \textit{Queen} were obtained with optical GOM ATOS Core 80/135 3D scanner (reported accuracy of 0.03mm). No spray coating or markers have been used to ease the acquisition to keep the geometry of the object consistent with the PS data. The remain objects:  \textit{Ball}, \textit{Bell}, \textit{Bunny}, \textit{Cup},\textit{Die}, \textit{Glass}, \textit{Hippo}, \textit{Jar}, \textit{Tool}, \textit{Squirrel} with the Zeiss CT scanner M1500/225 kV which provides an accuracy within the order of 9$\mu m$.

 \paragraph{Alignment.} We use the standard 2D to 3D alignment procedure as in \cite{Shi_2016_CVPR, Mecca21luces}. This involves the mutual information filter of MeshLab~\cite{CignoniCCDGR08} with  manual initialization and then repetitive refinement (until `pixel perfect' using the semi-transparent overlay). For 2D reference image, the average photometric stereo image was used, with additional manual image processing (exposure, brightness and contrast). Using the aligned meshes, ground truth normal maps and segmentation masks were rendered (with PyTorch3D).


\begin{table*}[t]
\setlength{\tabcolsep}{3.0pt} 
\begin{center}
\resizebox{1.0\textwidth}{!}{%
\hspace{-0.0125\textwidth}
\begin{tabular}{ | c | c |c c c c c c c c c c  | c|}
 \hline
Method & Error  & Bowl & Buddha & Bunny & Cup & Die & Hippo & House & Owl & Queen & Squirrel & Average \\ \hline
UniPS~\cite{ikehata2023sdmunips} (uncalibrated) & MAE  & 37.3 & 18.3 & 15.5 & 23.9 & 25.3 & 18.9 & 37.6 & 28.2 & 17.7 & 17.0 &24.0\\ 
 \hline
  Uni MS-PS~\cite{hardyunips} (uncalibrated) & MAE  & 17.5 & \textbf{13.7} & \textbf{10.6} & \textbf{12.3} & \textbf{10.3} & \textbf{9.6} & \textbf{33.0} & \textbf{14.7} & \textbf{12.9} & \textbf{11.3} & \textbf{14.6}  \\
\hline 
NF-PX-Net~\cite{Logothetis22} (calibrated) & MAE & \textbf{16.7} & 20.1 & 12.8 & 19.9 & 19.4 & 14.3 & 36.2 & 18.8 & 15.8 & 19.7  & 19.4 \\ \hline
\end{tabular}
} 
\end{center}
\caption{This figure provides the evaluation of two uncalibrated PS normal estimation methods (UniPS~\cite{ikehata2023sdmunips} and Uni MS-PS~\cite{hardyunips}) and one calibrated normal estimation method NF-PX-Net~\cite{Logothetis22}. Uni MS-PS~\cite{hardyunips} being the most recent method outperforms other competitors on all objects except Bowl. Note, however the average angular error of $14.6^{\circ}$ indicating a large room for improvement especially for objects like \textit{Bowl}, \textit{House}, \textit{Cup} and \textit{Buddha}.}
\label{tab:normalresults}
\end{table*}

\subsection{Dataset Overview}

LUCES-MV contains 10 pose aligned objects namely \textit{Bowl}, \textit{Buddha}, \textit{Bunny}, \textit{Cup}, \textit{Die}, \textit{Hippo}, \textit{House}, \textit{Owl}, \textit{Queen}, \textit{Squirrel} as well as 5 unaligned (due to rotational symmetry) objects \textit{Ball}, \textit{Bell}, \textit{Glass}, \textit{Jar}, \textit{Tool} of diverse materials and geometry as shown in Figure~\ref{fig:objects}. Noteworthy properties are extreme concavity (creating shadows and self reflections) in \textit{Bowl}, extreme metallic shininess in \textit{Cup} extreme complicated topology in  \textit{House}. See a more in depth explanation in the supplementary material.

For each object, PS images were captured at every $10^o$ leading to a total of 36 stereo pairs, (72 views x 15 PS images =1080 RGB images per object). However, due to the manual aligning processes being laborious, for each object only 12 views were aligned (all stereo pairs every $60^o$) and that constitutes the main part of the dataset which is used for evaluation experiments of Section~\ref{sec:experiments}. Nevertheless, all of the data will be released (including all 4 Bayer RAW images that are used to generate each RGB image), inviting the community to improve calibration-free methods;  standard SFM like COLMAP~\cite{schoenberger2016sfm} and self-calibrating neural reconstruction like BARF~\cite{Lin_2021_ICCV} are unreliable on texture-less object on completely black background.


\begin{figure*}[t]
\centering
 \includegraphics[width=0.86\textwidth]{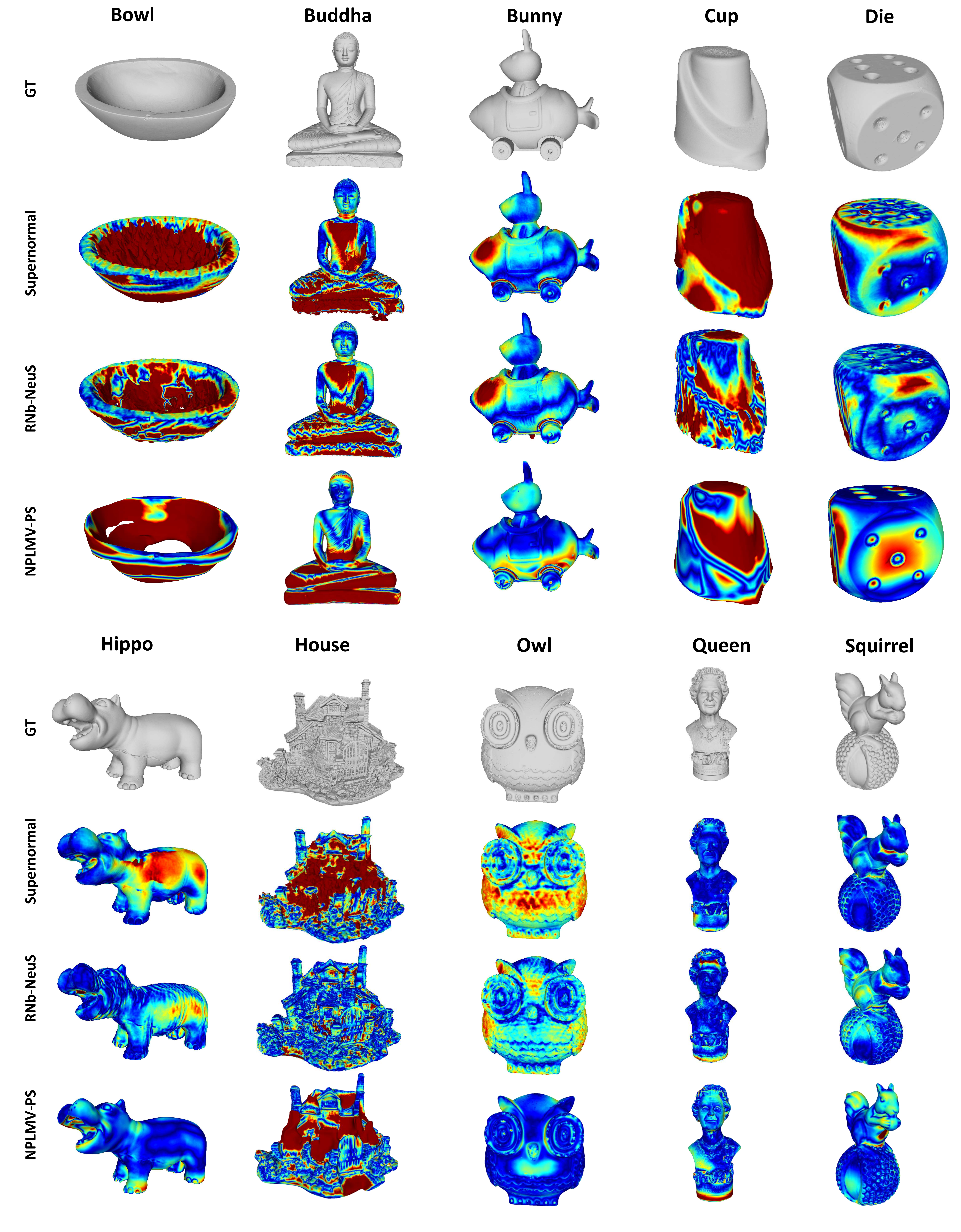}
 \caption{This figure illustrates reconstruction to GT Hausdorff distance for each point of predicted meshes for all three MVPS methods. We emphasize that although the distance for all points is visualized, the numbers reported on Table~\ref{tab:mainresults} are computed on visible points only (computing visibility for both GT and reconstructions).  As indicated by Table~\ref{tab:mainresults}, RNb-NeuS~\cite{BrumentRNb24} shows same or slightly better performance than other two methods. In particular, significantly lower error on \textit{House} object is very impressive and is likely due to implicit stereo matching of the texture through the use of albedo maps. As evident from the results shape error could be significantly improved on most of the objects. Improvements in both better normal estimation as well as neural rendering of pixel intensities are required to improve SOTA on LUCES-MV dataset. Dark red color corresponds to per point closest distance of 1 millimeter and dark blue - 0.}  
 \label{fig:mainshape}
\end{figure*}

\section{Experiment setup}
\label{sec:experiments}

We evaluated multiple different methods on three key Photometric stereo tasks: single view normal estimation, binocular photometric stereo shape estimation and multi-view photometric stereo shape estimation which have different classes of applications 

\noindent
\textbf{Single view normal estimation.} For single view normal estimation we compare two SOTA universal photometric stereo methods UniPS~\cite{ikehata2023sdmunips} and Uni MS-PS~\cite{hardyunips} as well as a calibrated normal estimation method NF-PX-Net~\cite{Logothetis22}. These 3 methods are among the current best performers on single view DiLiGenT \cite{Shi_2016_CVPR} and have public test code and network weights. UniPS~\cite{ikehata2023sdmunips} also provides an estimate of albedo maps. For all methods, ground truth segmentation maps are provided. NF-PX-Net~\cite{Logothetis22} requires light calibration parameters are well as an approximate mean depth.

We evaluate mean angular error (as it is the standard in single view PS datasets like \cite{Shi_2016_CVPR}) shown in Table~\ref{tab:normalresults}, obtained by taking an average pixel wise error for all 12 images per object. We also provide a visualization of pixelwise angular error for initial pose, viewpoint 1 for all methods and all objects in Figure~\ref{fig:normals}.

\noindent
\textbf{Binocular photometric stereo.} We also evaluate modern binocular photometric stereo method of~\cite{Logothetis24WACV}. 
Quantitively we evaluate mean and median shape errors computed as bidirectional Hausdorff distance (using the python interface of MeshLab\cite{CignoniCCDGR08}) between all \textit{visible} points on ground truth and predicted meshes. Quantitative in Table~\ref{tab:mainresults} and qualitative results are provided in supplementary.
Note that qualitative visualization shows the shape error from reconstruction to ground truth for all points, but only visible points are included on the computation in Table~\ref{tab:mainresults}.

\noindent
\textbf{Multi-view photometric stereo.} The key experiments involve evaluating SOTA multi-view photometric stereo methods of Supernormal~\cite{cao2023supernormal}, RNb-NeuS~\cite{BrumentRNb24} and NPLMV-PS~\cite{LogothetisWACV2025}. As the the latter method has versions which either uses only normal information (N) or only pixel intensities information (I) we provide three evaluation settings, including where both pieces of information are used. The same evaluation (Table~\ref{tab:mainresults}) and visualization (Figure~\ref{fig:mainshape} ) metrics are used as for binocular photometric stereo method~\cite{Logothetis24WACV}. Note we use Uni MS-PS~\cite{hardyunips} normals for all three methods and the albedo maps of UniPS~\cite{ikehata2023sdmunips} for RNb-NeuS~\cite{BrumentRNb24}. We note that normal only (N) version of NPLMV-PS~\cite{LogothetisWACV2025} is conceptually very similar to Supernormal~\cite{cao2023supernormal} and intensities only (I) version of NPLMV-PS~\cite{LogothetisWACV2025} is conceptually very similar to \cite{Brahimi_2024_CVPR}, which unfortunately is closed source.

\begin{table*}[t]
\setlength{\tabcolsep}{3.0pt} 
\begin{center}
\resizebox{1.0\textwidth}{!}{%
\hspace{-0.0125\textwidth}
\begin{tabular}{ | c | c |c c c c c c c c c c  | c|}
 \hline
Method & Error  & Bowl & Buddha & Bunny & Cup & Die & Hippo & House & Owl & Queen & Squirrel  & Average \\ \hline

\multirow{3}{*}{\vspace{0.4cm} 
Binocular-PS~\cite{Logothetis24WACV} } & MN  & 19.02 & 5.67 & 1.93 & 5.37 & 2.97 & 10.12 & 5.50 & 1.72 &2.60 &3.67 & 5.86 \\
 & MD & 18.68 & 4.54 & 1.78 & 5.39 & 2.84 & 9.19 & 5.00 & 0.81 & 2.54 & 1.60 & 5.24\\ \hline \hline

 \multirow{3}{*}{\vspace{0.4cm} RNb-NeuS~\cite{BrumentRNb24} (GT-N)}  & MN  & 0.17 & 0.17 & 0.07 & 0.51 & 0.07 & 0.10 & 0.13 & 0.03 & 0.04 & 0.04 & 0.13 \\
 & MD  & 0.09 & 0.05 & 0.04 & 0.27 & 0.07 & 0.08 & 0.04 & 0.03 & 0.03 & 0.03 & 0.07 \\ \hline 
 \multirow{3}{*}{\vspace{0.4cm} RNb-NeuS-\cite{BrumentRNb24}} & MN  & \textbf{1.37} & \textbf{0.68} & \textbf{0.39} & \textbf{1.10} & \textbf{0.41} & 0.28 & \textbf{0.47} & 0.37 & \textbf{0.21} & 0.26 & \textbf{0.55} \\
 & MD  & 0.81  & 0.45  &  0.27  & 0.80  & 0.28  & 0.22  & 0.31  &0.34  & 0.16  & 0.23  &0.39 \\ \hline 
\multirow{3}{*}{\vspace{0.4cm} Supernormal\cite{cao2023supernormal}} & MN & 4.04  & 1.08  & 0.41  & 1.41  & 0.61  & 0.35  & 1.26  & 0.48  & 0.22  & \textbf{0.22}  & 1.01\\
 & MD  & 2.69 & 0.87 & 0.33 & 1.32 & 0.52 & 0.32 &0.92 & 0.46 &0.20 &0.20 & 0.78 \\ \hline 
\multirow{2}{*}{ NPLMV-PS~\cite{LogothetisWACV2025} (N)} & MN  & 2.14  & 1.06  & 0.50  & 1.22  & 0.79  & 0.28  & 1.66  & 0.26  & 0.33  & \textbf{0.22}  & 0.85  \\
& MD  & 1.81 & 0.63 & 0.33 & 0.98 & 0.51 & 0.23 & 0.67 & 0.19 & 0.24 & 0.17 & 0.57 \\ \hline
\multirow{2}{*}{ NPLMV-PS~\cite{LogothetisWACV2025} (I)} & MN  & 4.43 & 1.27 & 0.42 & 1.56 & 0.53 & 1.61 & 1.36 & 0.27 & 0.50 & 0.47 & 1.24 \\
& MD & 2.56 & 0.79 & 0.27 & 1.19 & 0.49 & 1.10 & 0.71 & 0.21 & 0.41 & 0.42 & 0.82  \\ \hline
\multirow{2}{*}{ NPLMV-PS~\cite{LogothetisWACV2025} (N+I)} & MN  & 2.03  & 0.91  & 0.45  & 1.22 & 0.43 & \textbf{0.25} & 1.06 & \textbf{0.25} & 0.33 & 0.24 & 0.72 \\
& MD  & 1.59 & 0.55 & 0.32 & 0.98 & 0.32 & 0.26 & 0.51 & 0.18& 0.28 & 0.19 &0.52  \\ \hline

\end{tabular}
} 
\end{center}
\caption{This table shows the computed results for a binocular photometric stereo method of~\cite{Logothetis24WACV} and three SOTA multi-view photometric stereo methods. RNb-Neus~\cite{BrumentRNb24} achieves the best mean shape error of 0.55mm and significantly (by 0.1mm or more) outperforms all of the competing approaches except for \textit{Hippo}, \textit{Owl} and \textit{Squirrel}. Unlike in DiLiGenT-MV~\cite{LiZWSDT20} benchmark, as reported by~\cite{LogothetisWACV2025} the  the error ratio (best error achieved on DiLiGent-MV~\cite{LiZWSDT20} when using ground truth normals is 0.11mm) of best method  run on predicted normals and ground truth normals is significantly higher (e.g. 4.2 times vs 1.8), indicating that LUCES-MV is a significantly more challenging benchmark. While all three methods evaluated on DiLiGenT-MV~\cite{LiZWSDT20} bechmark achieve errors within 0.04mm (ie. 0.2mm, 0.2mm and 0.24mm for NPLMV-PS~\cite{LogothetisWACV2025}, Supernormal~\cite{cao2023supernormal} and RNb-NeuS~\cite{BrumentRNb24}, on LUCES-MV there are clear differences in performance between all three methods. Also at least 6 of 10 of the objects have error larger than 0.3mm indicating that there is plenty of space for improving multi-view photometric stereo methods.}
\label{tab:mainresults}
\end{table*}

\section{Results}
\label{sec:results}
\noindent
We discuss the result of three types of experiments:

\noindent
\textbf{Single view normal estimation.} We observe that UniPS~\cite{ikehata2023sdmunips} has significantly worse performance than other two methods. This is not very surprising as it was optimized for far-field Photometric Stereo and thus cannot directly cope with near-light attenuation and perspective distortion. Uni MS-PS~\cite{hardyunips} shows best performance (both quantitatively and qualitatively) on all objects except \textit{Bowl}, despite NF-PX-Net~\cite{logothetis2021pxnet} benefiting from a calibrated normal estimation network. Specifically, NF-PX-Net~\cite{Logothetis22} has sightly poorer performance than uncalibrated Uni MS-PS~\cite{hardyunips} as it includes less material augmentation of its training dataset (e.g. see a very good performance of NF-PX-Net~\cite{logothetis2021pxnet} on \textit{Queen} object) and also, since it has been trained on independent samples of observational maps it has inferior ability to correctly predict normals along the edges and in regions suffering from severe reflections). Finally, NF-PX-Net~\cite{Logothetis22} also shows significantly poor ability to predict normals at oblique angles but is competitive to Uni MS-PS~\cite{hardyunips} on smooth regions (e.g. \textit{Bunny}, body of \textit{Hippo}, face of \textit{Squirrel}). Poor performance on \textit{Bowl}, \textit{House}, \textit{Cup} of Uni MS-PS~\cite{hardyunips} hints that significant research efforts are required to better material modeling for single view Photometric Stereo, which we hope our dataset will contribute to fostering.

\noindent
\textbf{Binocular photometric stereo.} While binocular photometric stereo~\cite{Logothetis24WACV} has many applications in robotics, there are yet to appear a significant number of competing methods. Binocular PS result of \cite{Logothetis24WACV} (first section in Table~\ref{tab:mainresults}  ) is significantly worse that the respective multi-view results, as well as the original result that it was obtained in LUCES-ST \cite{Logothetis24WACV} and 2 views on DiLiGenT-MV \cite{LiZWSDT20}. The qualitative visualization (in supplementary) reveals that although the surfaces are locally correct, the overall shape suffers due to over smoothing of occlusion boundaries. This loss of accuracy (especially compared to LUCES-ST \cite{Logothetis24WACV}) is explained by the fact that the object distance is more that twice in LUCES-MV than LUCES-ST ($\sim35$cm to $\sim15$cm) and thus the reduction of parallax is reducing the value of stereo. DiLiGenT-MV also contains circular motion so the parallax between views is significant.

\noindent
\textbf{Multi-view photometric stereo.} The worst performing method is the intensity only (I)  version of NPLMV-PS~\cite{LogothetisWACV2025}, which is the only method not utilizing the normal estimates. This signifies that there is significant scope for improvement on the neural rendering for challenging PS objects. 

Normal only (N) version of  NPLMV-PS~\cite{LogothetisWACV2025} is also outperforming the functionally similar Supernormal~\cite{cao2023supernormal} method on most of the objects with exception of the very smooth ones \textit{Bunny}, \textit{Die} and \textit{Queen}. This signifies that the patch rendering procedure of Supernormal~\cite{cao2023supernormal} is potentially a loss of accuracy in case of high details (with \textit{House} being an outlier).

Additionally, including N+I on NPLMV-PS~\cite{LogothetisWACV2025} significantly improves it compared to its other 2 versions showing that the 2 sources of information act synergistically. Not surprising, the albedo information also allows RNb-NeuS~\cite{BrumentRNb24} to show very good performance, on most objects with exception of \textit{Buddha} and \textit{Cup}, yet still being the overall winner with a significant margin (0.55mm mean vs 0.72mm of NPLMV-PS~\cite{LogothetisWACV2025}). Nevertheless, this performance  is far from the performance observed in DiLiGenT-MV bechmark as per evaluation of~\cite{LogothetisWACV2025}.

Finally, we provide evaluation of RNb-NeuS~\cite{BrumentRNb24} (the best overall approach) with ground truth normals as a way to estimate potential performance that could be expected with future improvements on normal estimation methods. Note that this is an imperfect estimate as predicted albedos are also used (as there is no way to compute ground truth albedos) and thus providing relatively mediocre results on the \textit{Cup} (0.51mm mean, 0.27mm median)  were the albedo estimation is challenging. Nevertheless, all other objects achieve around 0.05mm to 0.15mm which is considerably better than any actual fair predictions, showing that there is significant scope for future improvements.

\section{Conclusion}

In this paper, we propose LUCES-MV, the first real-world, multi-view dataset designed for near-field point light source photometric stereo. Our dataset includes 15 objects with diverse materials, each imaged under varying light conditions from an array of 15 LEDs positioned 30 to 40 centimeters from the camera center. 

We evaluate state-of-the-art near-field photometric stereo algorithms, highlighting their strengths and limitations across different material and shape complexities. LUCES-MV dataset offers an important benchmark for developing more robust, accurate and scalable real-world Photometric Stereo based 3D reconstruction methods.

{\small
\bibliographystyle{ieee_fullname}
\bibliography{egbib}
}

\appendix

\begin{figure*}[h]
\centering
\includegraphics[width=\textwidth]{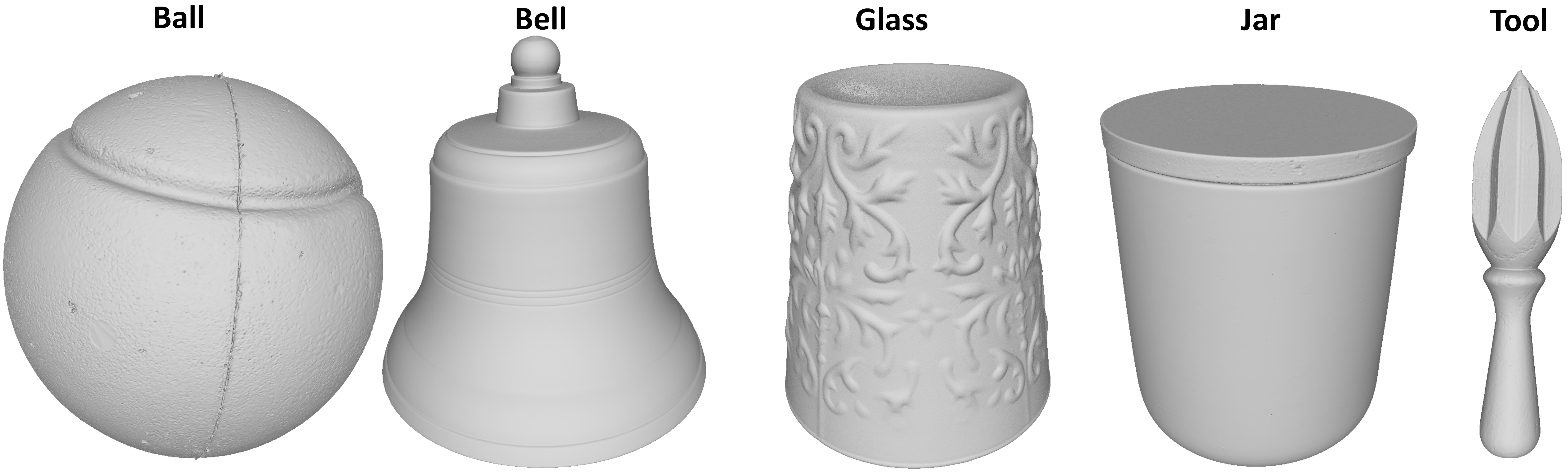}
\caption{Ground truth meshes for the additional, \textit{unposed} objects namely \textit{Ball}, \textit{Bell}, \textit{Glass}, \textit{Jar}, \textit{Tool}. The respective average photometric stereo images are shown in Figure~\ref{fig:obs3}. }
\label{fig:additionalgt}
\end{figure*}

\section{Additional Dataset Details}
\label{sec:data_extra}

In this section we provide additional details of LUCES-MV dataset which were omitted from the main paper. In particular, we discuss the unposed objects, and images used for camera intrinsic calibration and light calibration.

\subsection{Ground truth meshes for the objects}

\begin{figure*}[t]
\centering
\includegraphics[height=0.12\textwidth]{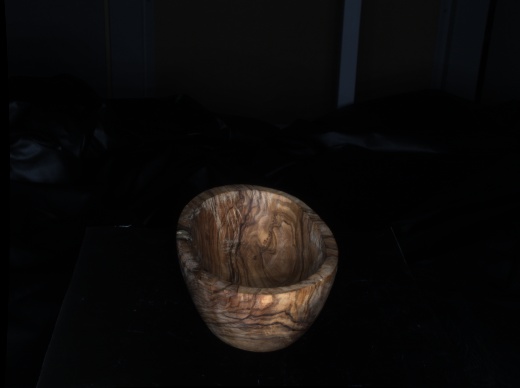}
\includegraphics[height=0.12\textwidth]{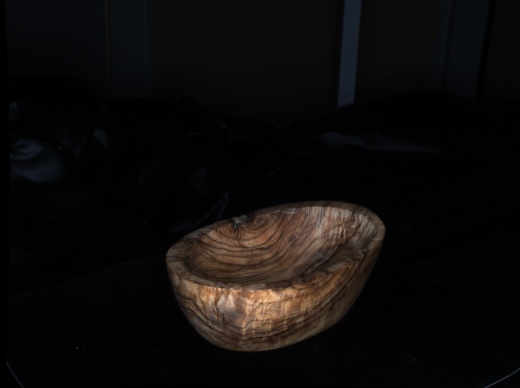}
\includegraphics[height=0.12\textwidth]{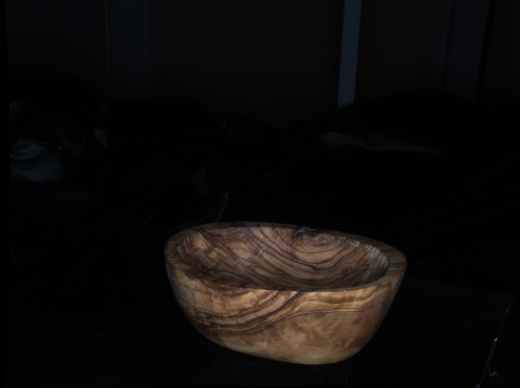}
\includegraphics[height=0.12\textwidth]{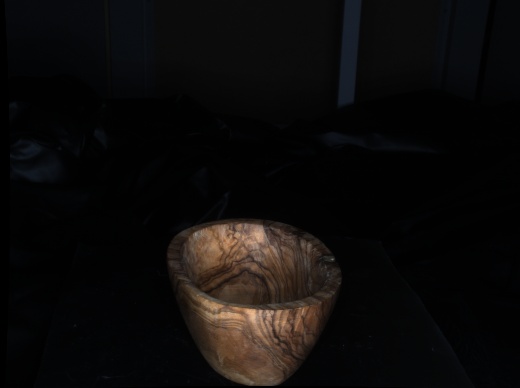}
\includegraphics[height=0.12\textwidth]{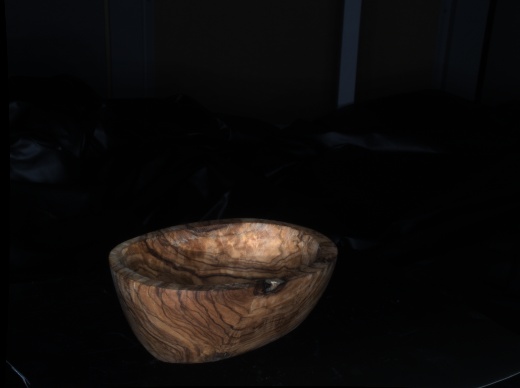}
\includegraphics[height=0.12\textwidth]{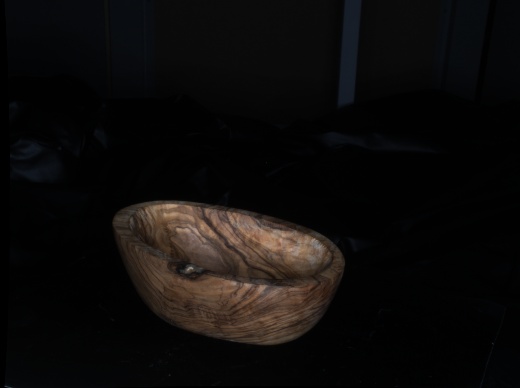}
\includegraphics[height=0.12\textwidth]{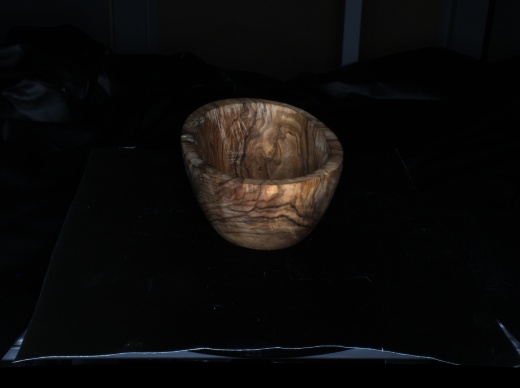}
\includegraphics[height=0.12\textwidth]{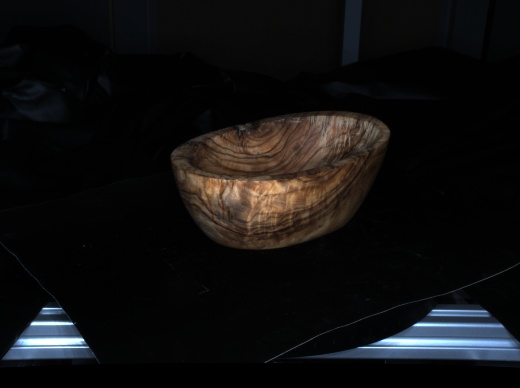}
\includegraphics[height=0.12\textwidth]{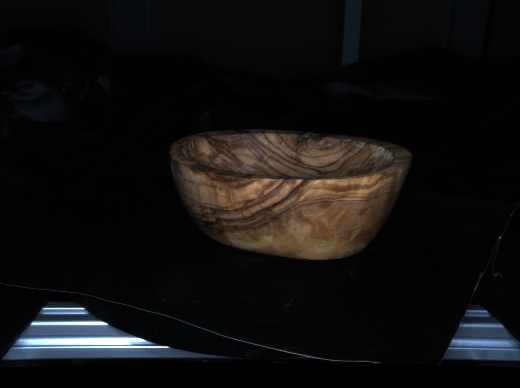}
\includegraphics[height=0.12\textwidth]{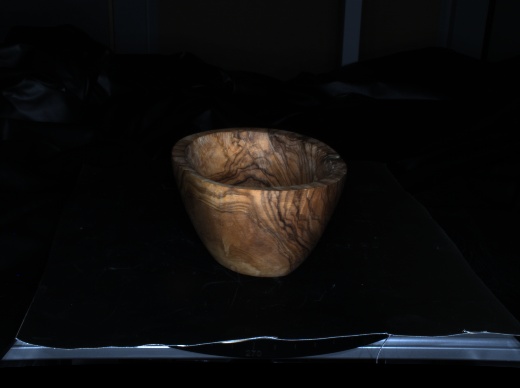}
\includegraphics[height=0.12\textwidth]{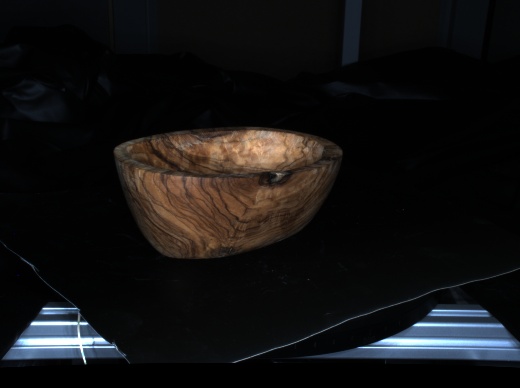}
\includegraphics[height=0.12\textwidth]{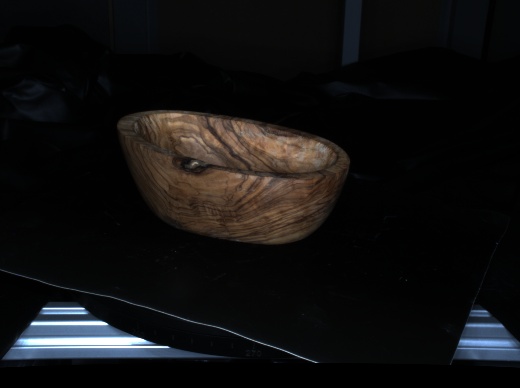}
\includegraphics[height=0.12\textwidth]{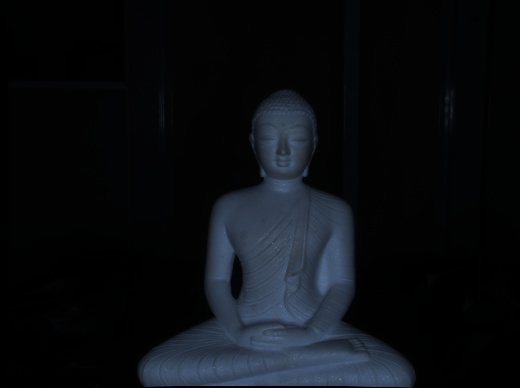}
\includegraphics[height=0.12\textwidth]{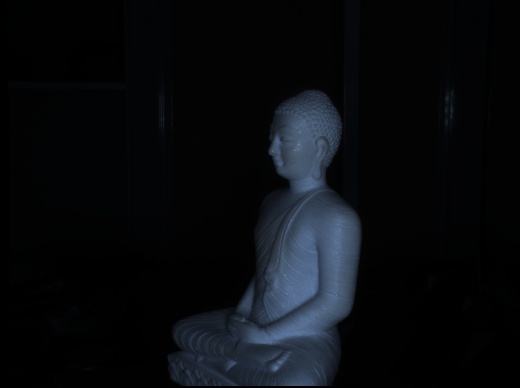}
\includegraphics[height=0.12\textwidth]{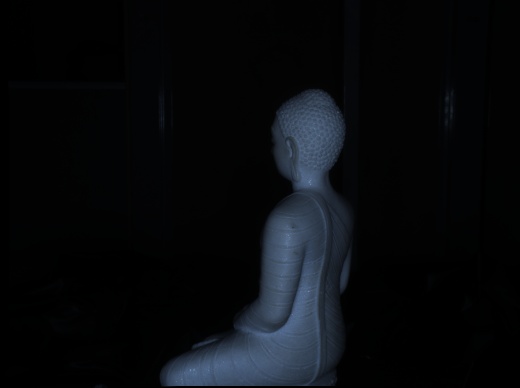}
\includegraphics[height=0.12\textwidth]{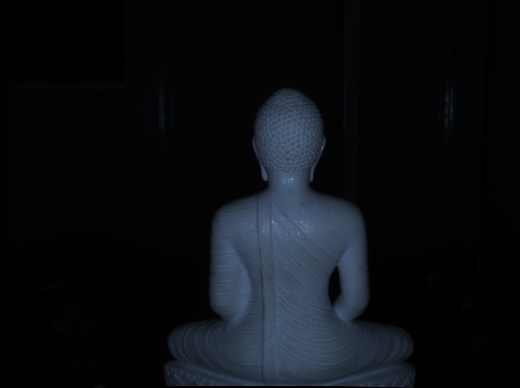}
\includegraphics[height=0.12\textwidth]{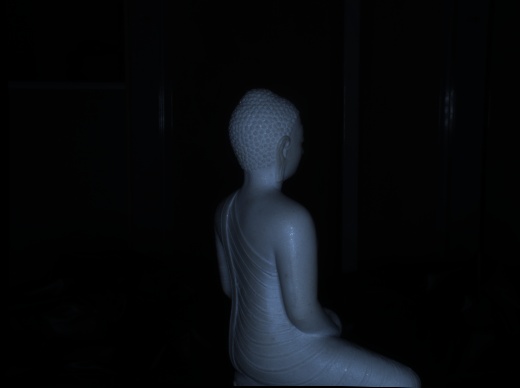}
\includegraphics[height=0.12\textwidth]{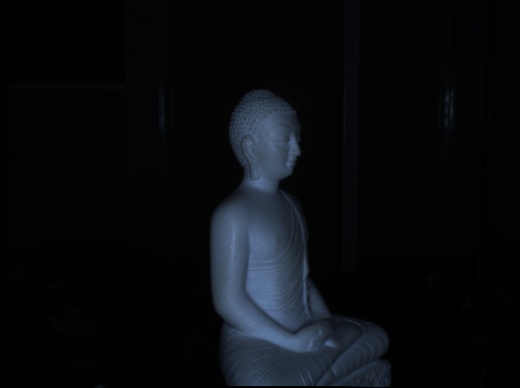}
\includegraphics[height=0.12\textwidth]{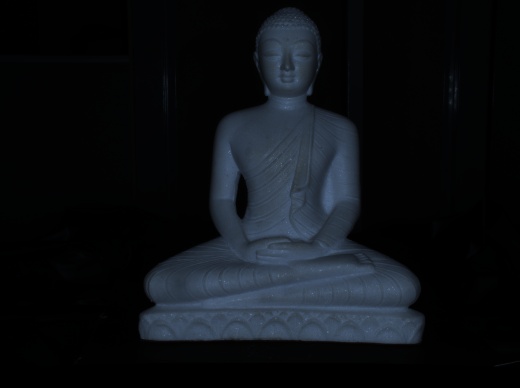}
\includegraphics[height=0.12\textwidth]{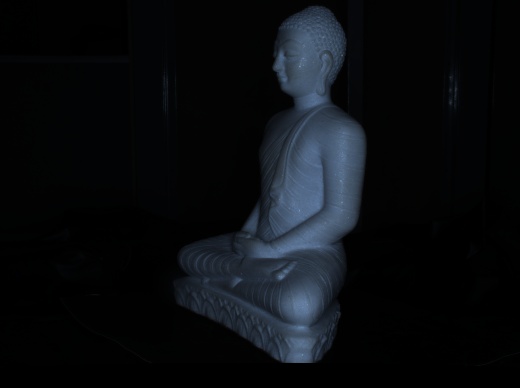}
\includegraphics[height=0.12\textwidth]{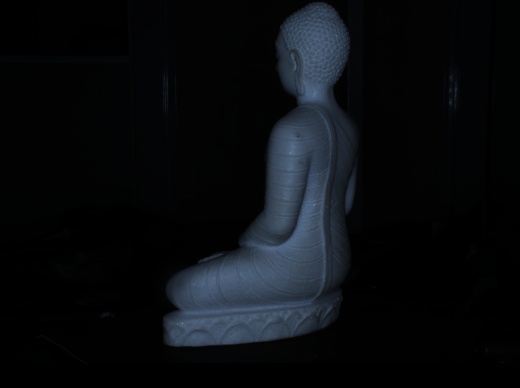}
\includegraphics[height=0.12\textwidth]{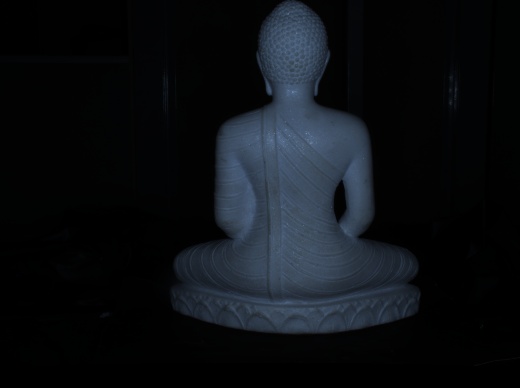}
\includegraphics[height=0.12\textwidth]{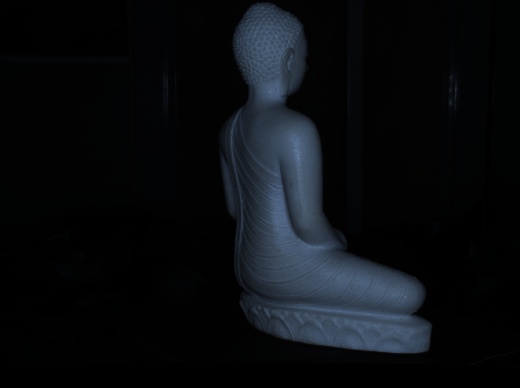}
\includegraphics[height=0.12\textwidth]{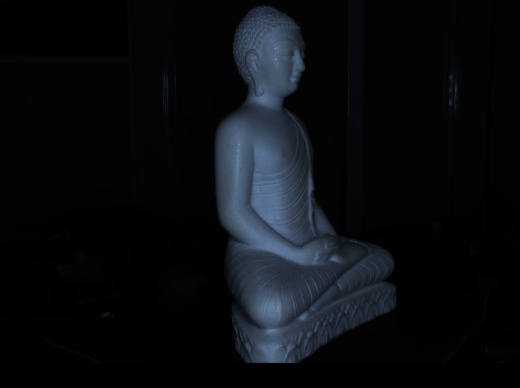}
\includegraphics[height=0.12\textwidth]{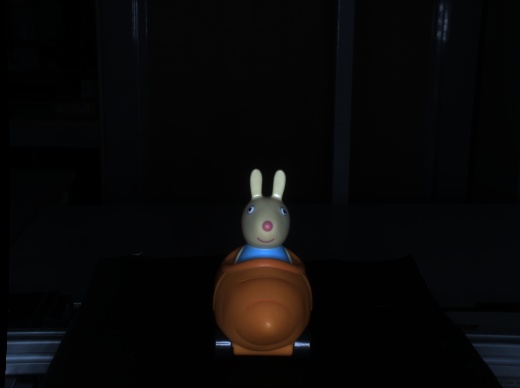}
\includegraphics[height=0.12\textwidth]{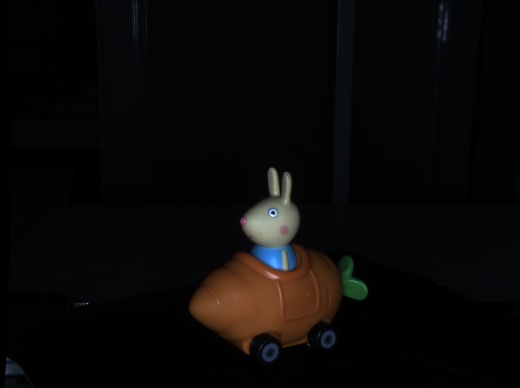}
\includegraphics[height=0.12\textwidth]{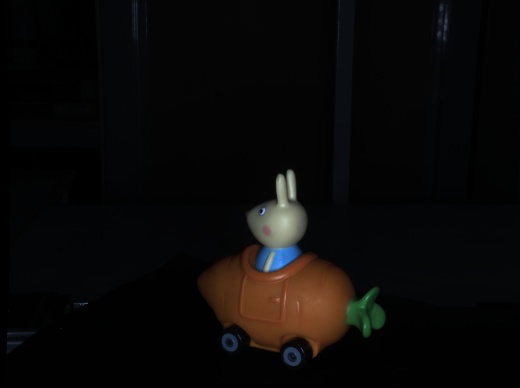}
\includegraphics[height=0.12\textwidth]{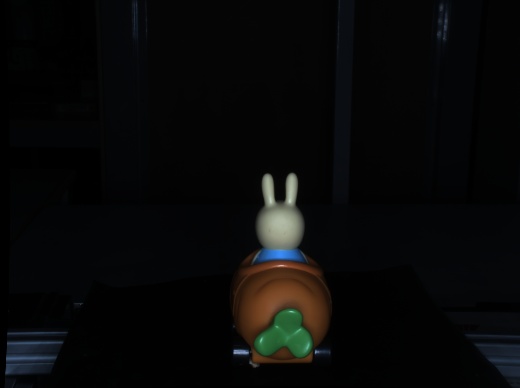}
\includegraphics[height=0.12\textwidth]{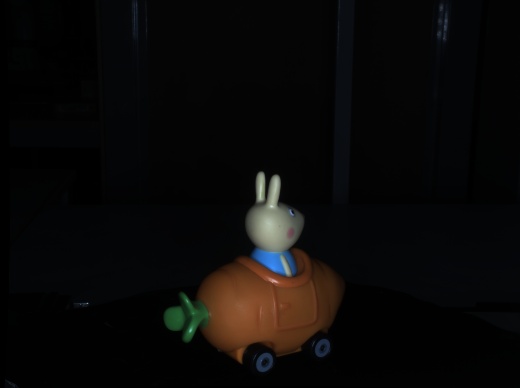}
\includegraphics[height=0.12\textwidth]{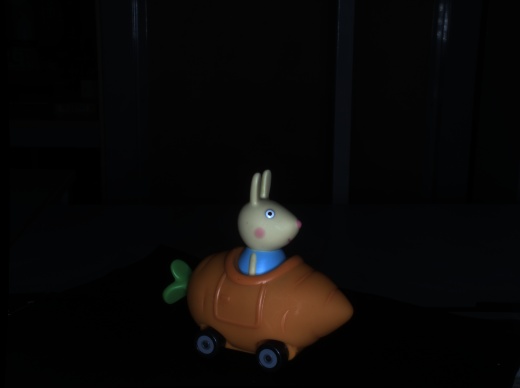}
\includegraphics[height=0.12\textwidth]{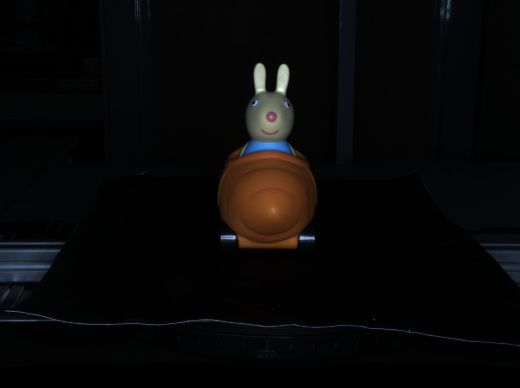}
\includegraphics[height=0.12\textwidth]{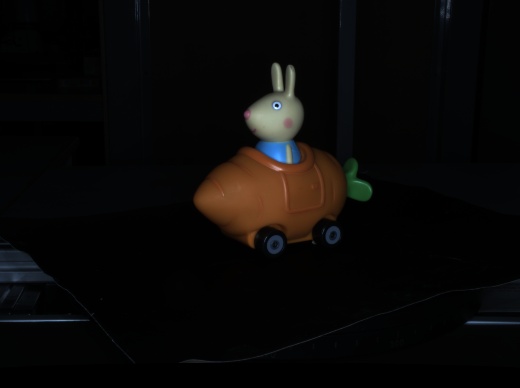}
\includegraphics[height=0.12\textwidth]{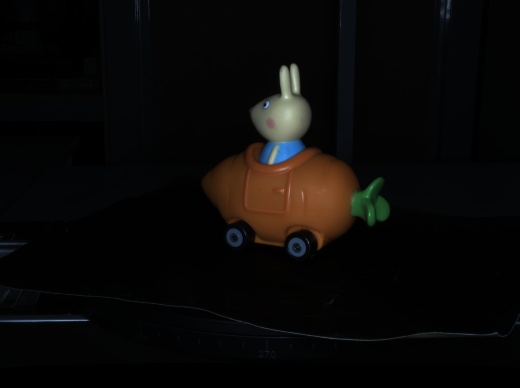}
\includegraphics[height=0.12\textwidth]{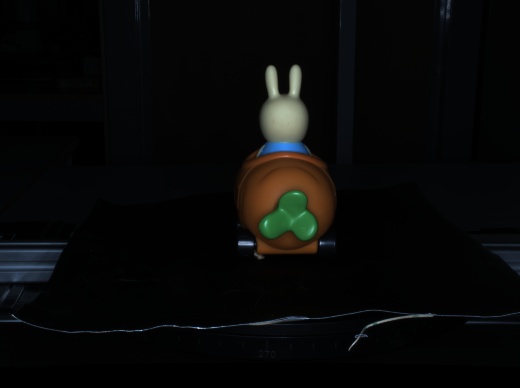}
\includegraphics[height=0.12\textwidth]{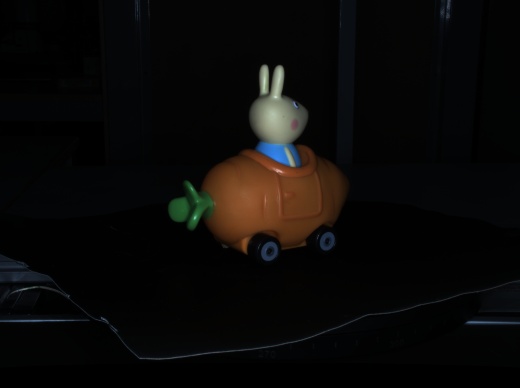}
\includegraphics[height=0.12\textwidth]{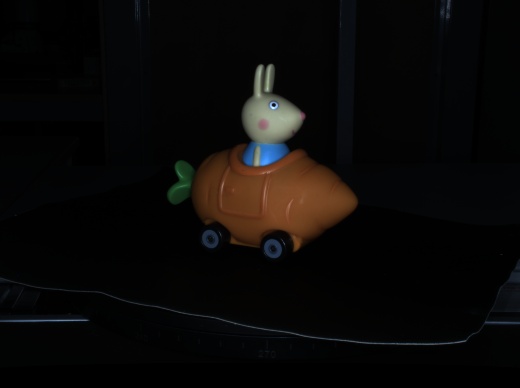}
\includegraphics[height=0.12\textwidth]{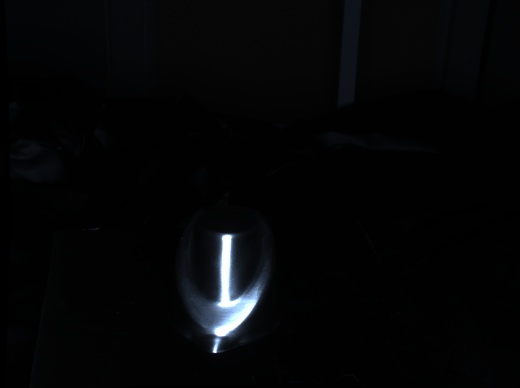}
\includegraphics[height=0.12\textwidth]{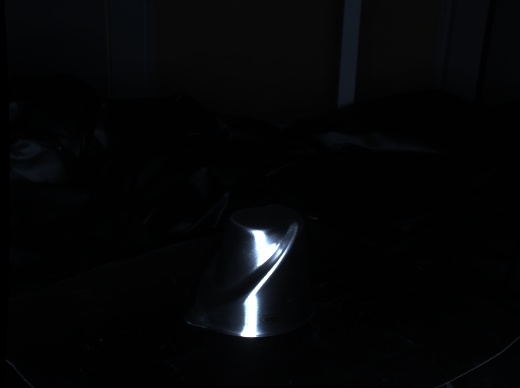}
\includegraphics[height=0.12\textwidth]{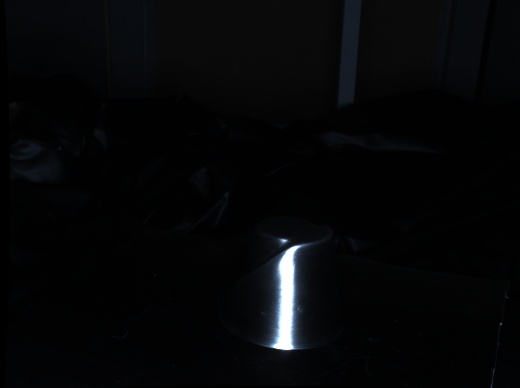}
\includegraphics[height=0.12\textwidth]{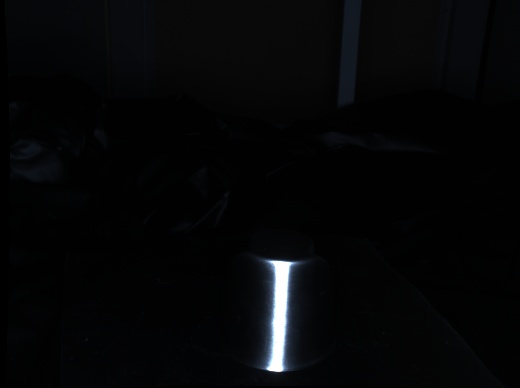}
\includegraphics[height=0.12\textwidth]{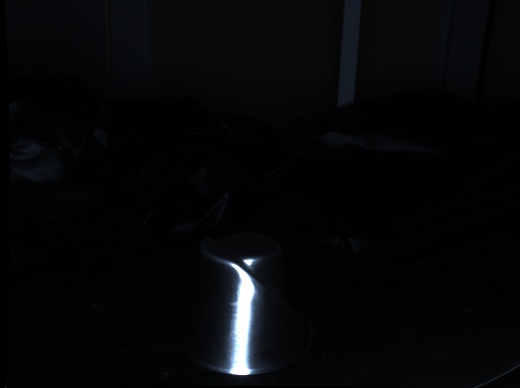}
\includegraphics[height=0.12\textwidth]{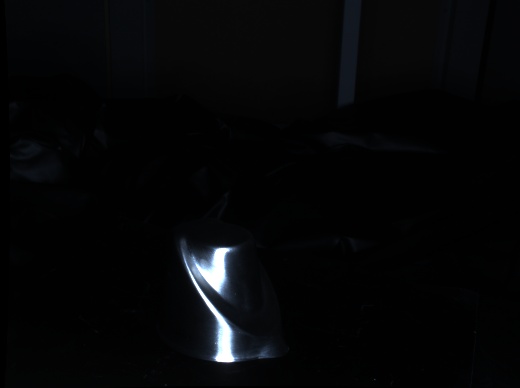}
\includegraphics[height=0.12\textwidth]{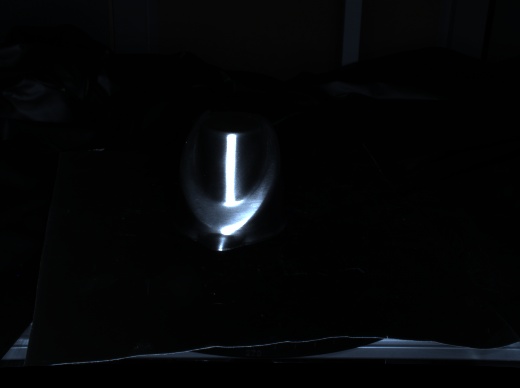}
\includegraphics[height=0.12\textwidth]{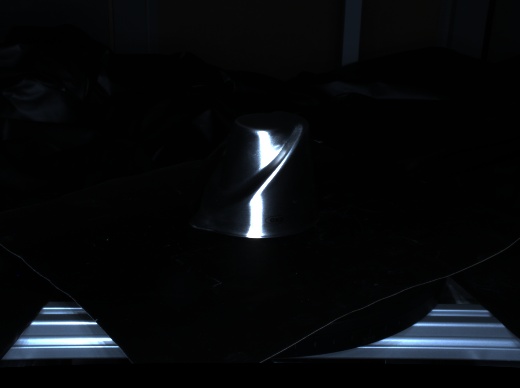}
\includegraphics[height=0.12\textwidth]{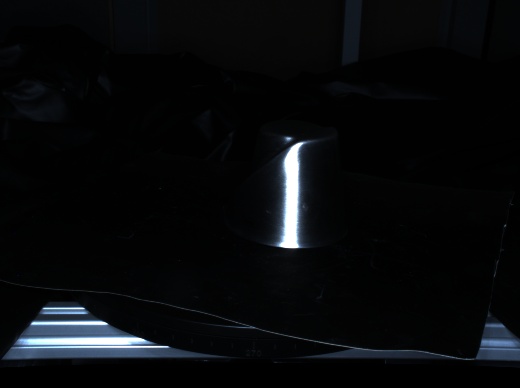}
\includegraphics[height=0.12\textwidth]{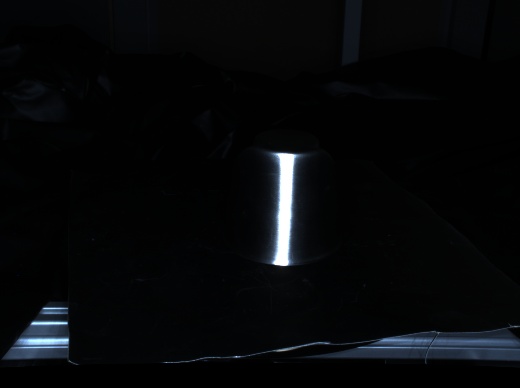}
\includegraphics[height=0.12\textwidth]{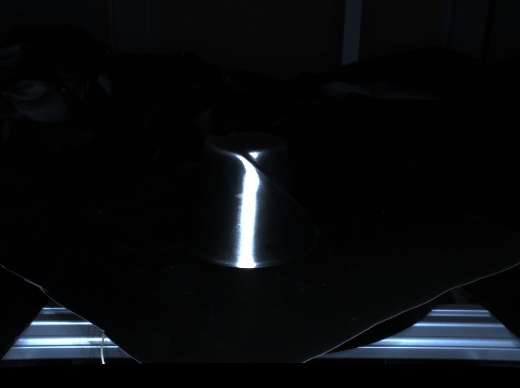}
\includegraphics[height=0.12\textwidth]{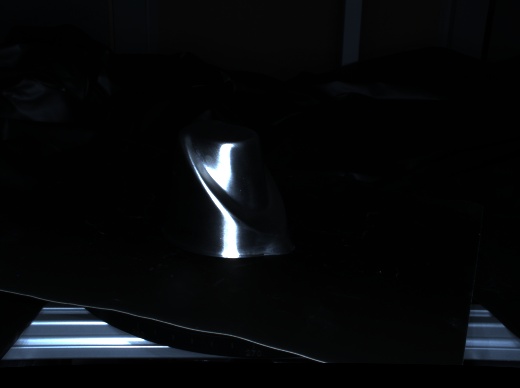}
\includegraphics[height=0.12\textwidth]{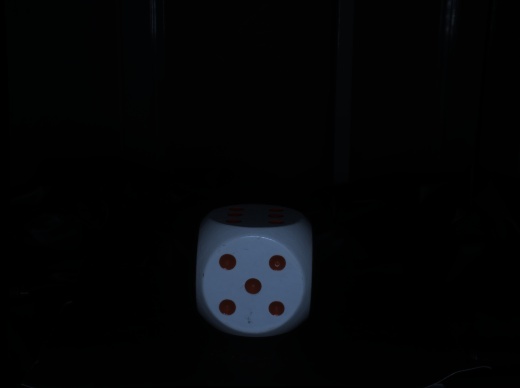}
\includegraphics[height=0.12\textwidth]{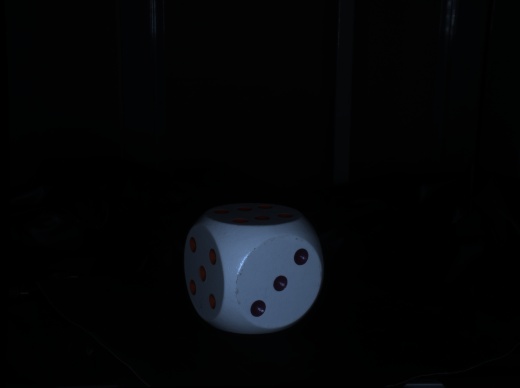}
\includegraphics[height=0.12\textwidth]{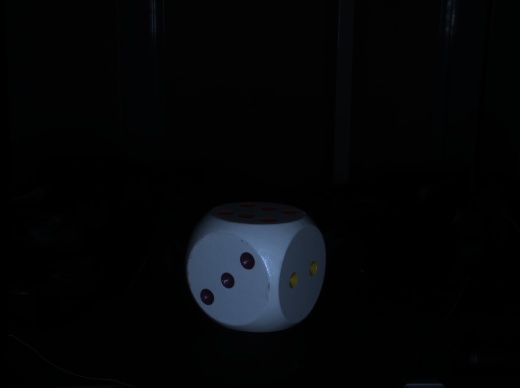}
\includegraphics[height=0.12\textwidth]{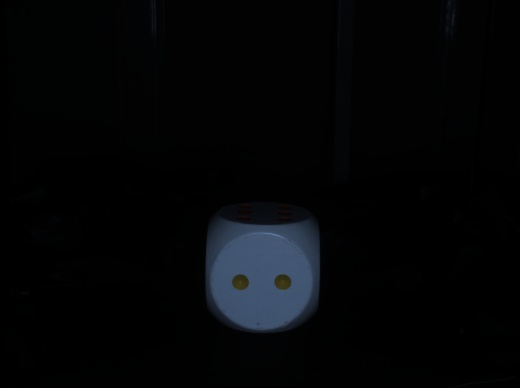}
\includegraphics[height=0.12\textwidth]{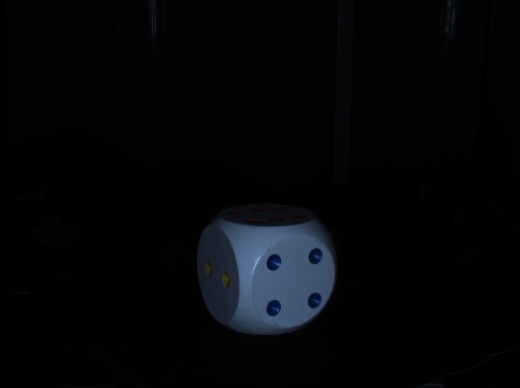}
\includegraphics[height=0.12\textwidth]{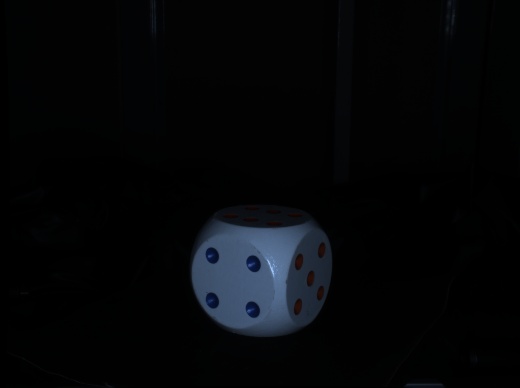}
\includegraphics[height=0.12\textwidth]{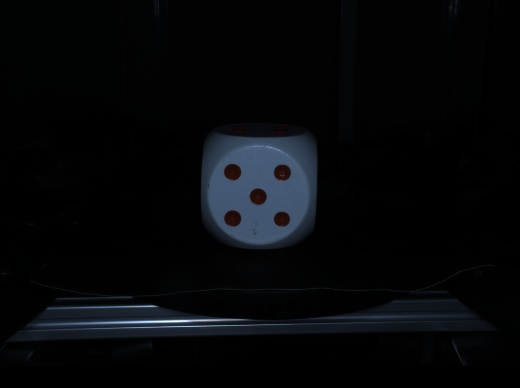}
\includegraphics[height=0.12\textwidth]{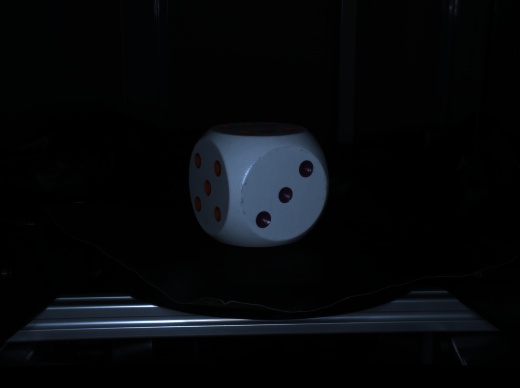}
\includegraphics[height=0.12\textwidth]{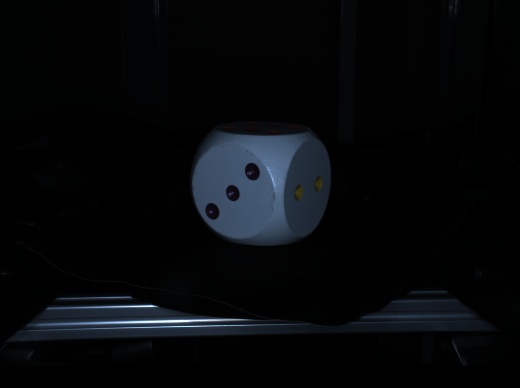}
\includegraphics[height=0.12\textwidth]{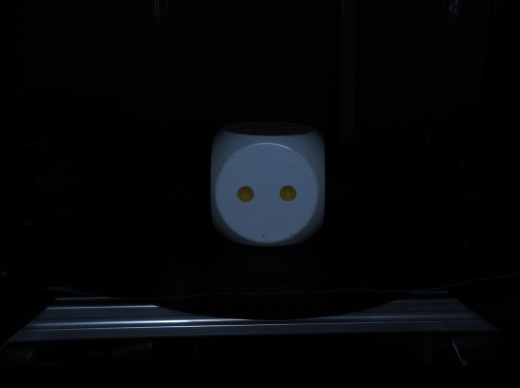}
\includegraphics[height=0.12\textwidth]{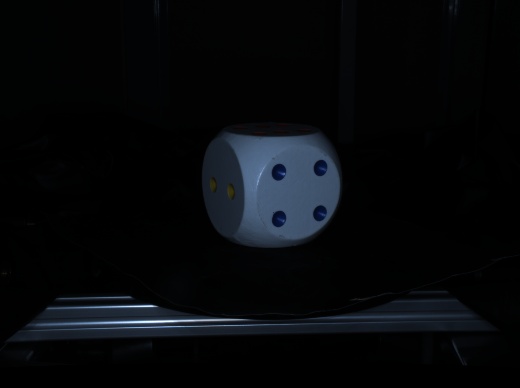}
\includegraphics[height=0.12\textwidth]{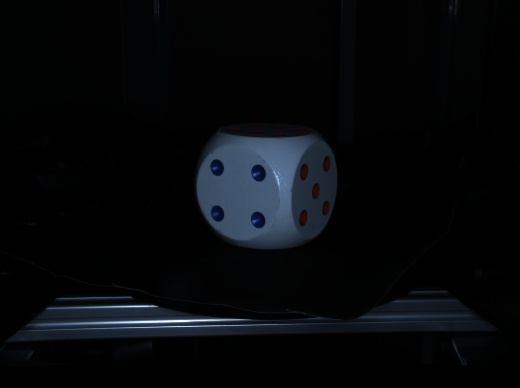}
 \caption{Average PS image for all views for \textit{Bowl}, \textit{Buddha}, \textit{Bunny}, \textit{Cup}, \textit{Dice}. }
 \label{fig:obs1}
\end{figure*}

\begin{figure*}[t]
\centering
\includegraphics[height=0.12\textwidth]{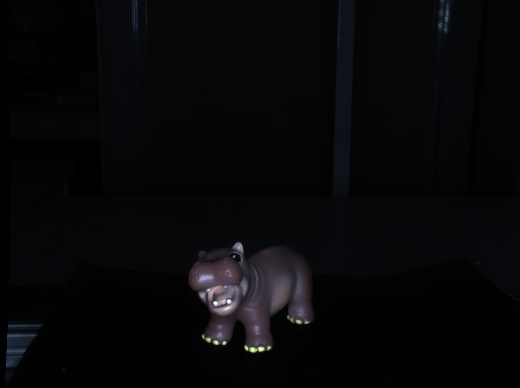}
\includegraphics[height=0.12\textwidth]{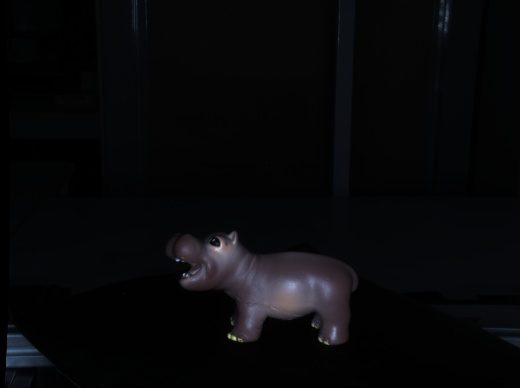}
\includegraphics[height=0.12\textwidth]{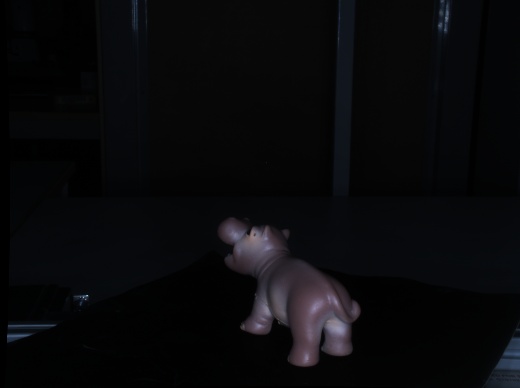}
\includegraphics[height=0.12\textwidth]{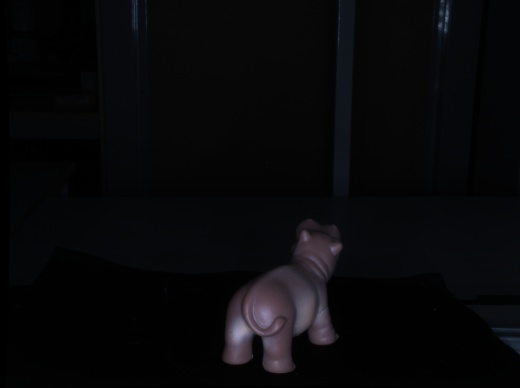}
\includegraphics[height=0.12\textwidth]{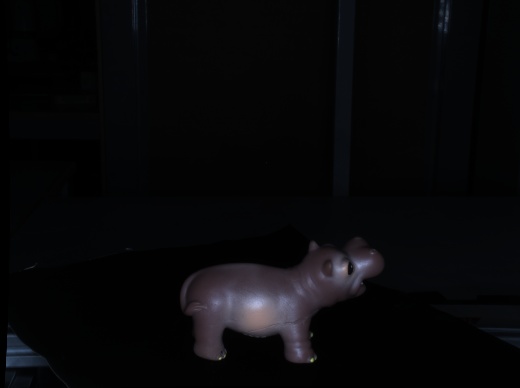}
\includegraphics[height=0.12\textwidth]{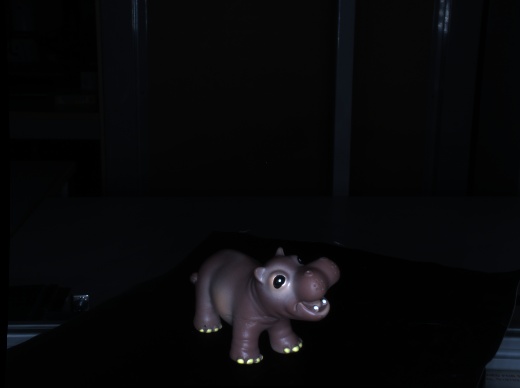}
\includegraphics[height=0.12\textwidth]{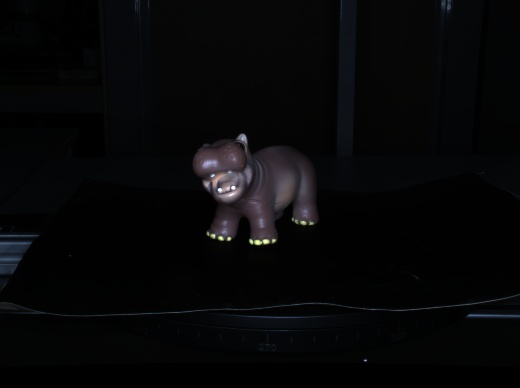}
\includegraphics[height=0.12\textwidth]{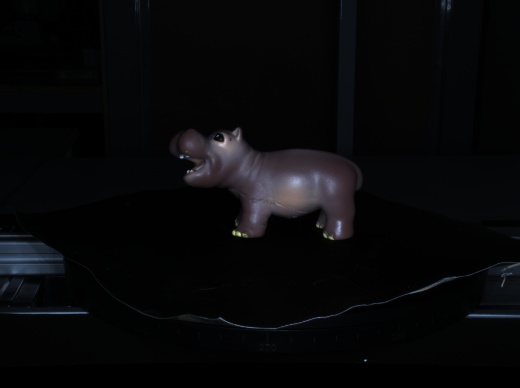}
\includegraphics[height=0.12\textwidth]{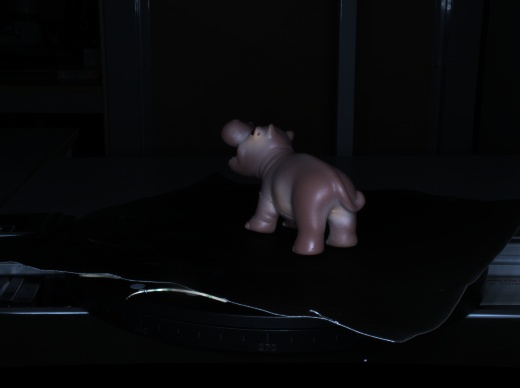}
\includegraphics[height=0.12\textwidth]{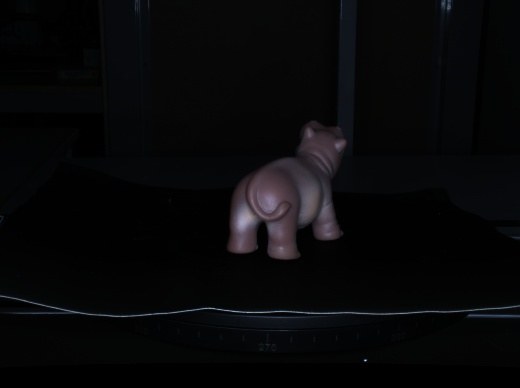}
\includegraphics[height=0.12\textwidth]{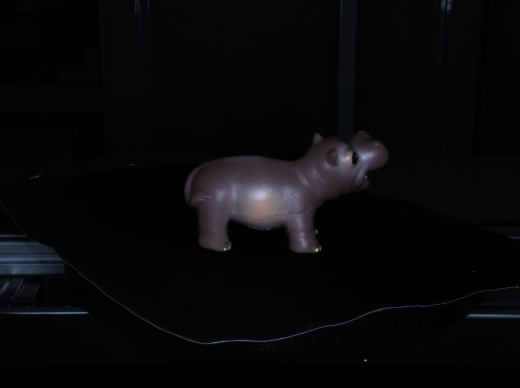}
\includegraphics[height=0.12\textwidth]{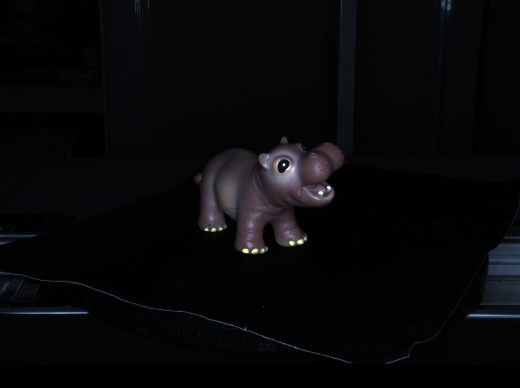}
\includegraphics[height=0.12\textwidth]{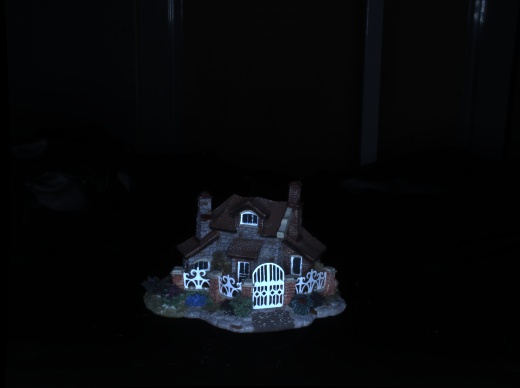}
\includegraphics[height=0.12\textwidth]{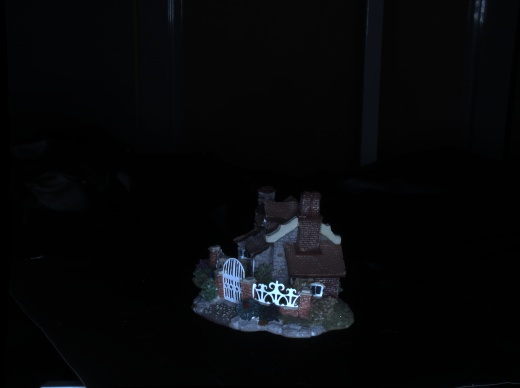}
\includegraphics[height=0.12\textwidth]{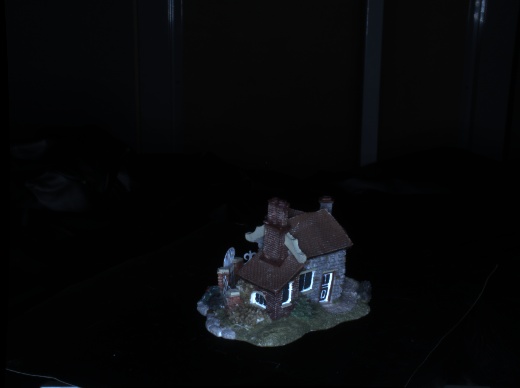}
\includegraphics[height=0.12\textwidth]{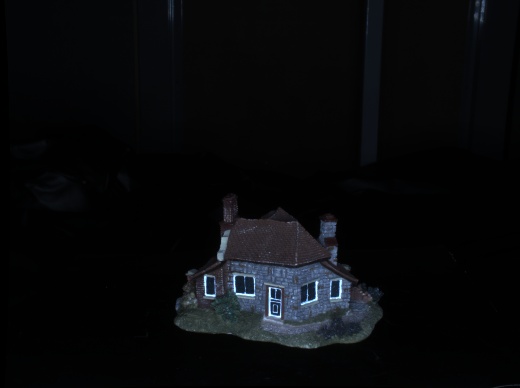}
\includegraphics[height=0.12\textwidth]{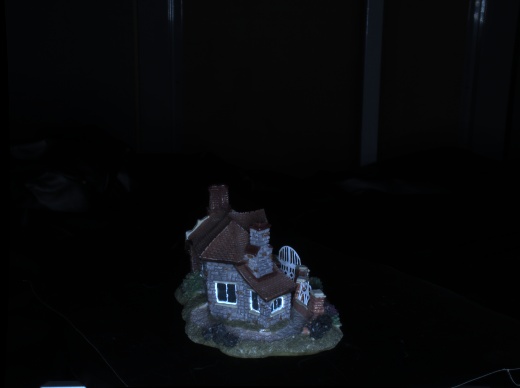}
\includegraphics[height=0.12\textwidth]{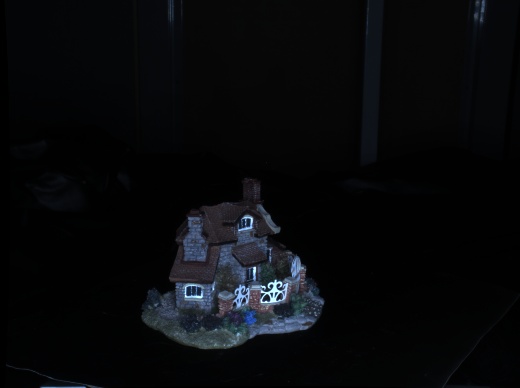}
\includegraphics[height=0.12\textwidth]{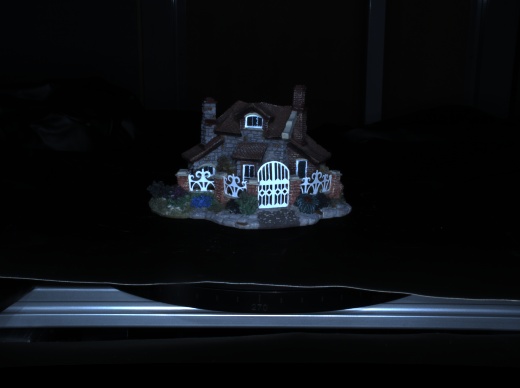}
\includegraphics[height=0.12\textwidth]{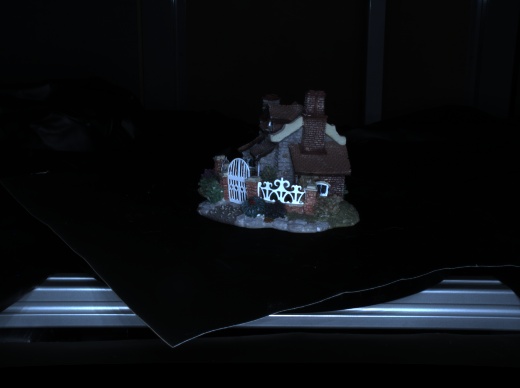}
\includegraphics[height=0.12\textwidth]{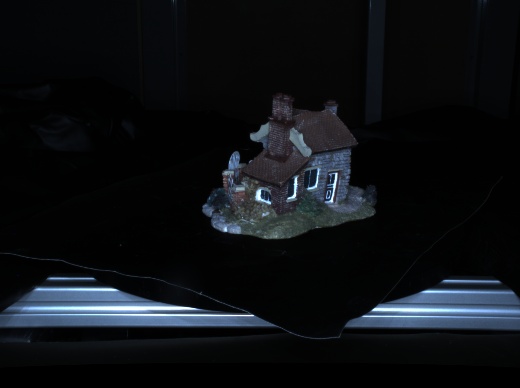}
\includegraphics[height=0.12\textwidth]{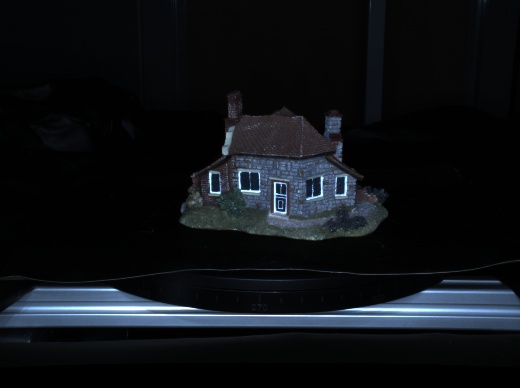}
\includegraphics[height=0.12\textwidth]{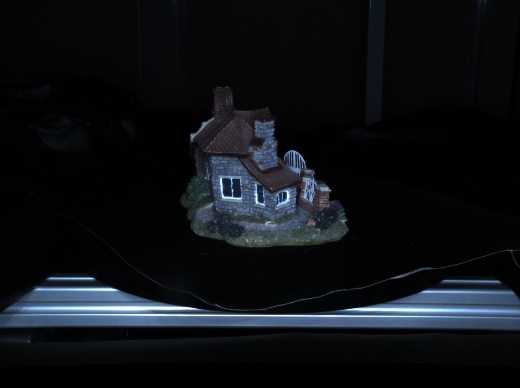}
\includegraphics[height=0.12\textwidth]{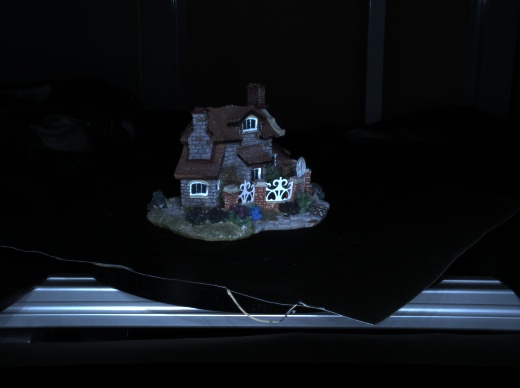}
\includegraphics[height=0.12\textwidth]{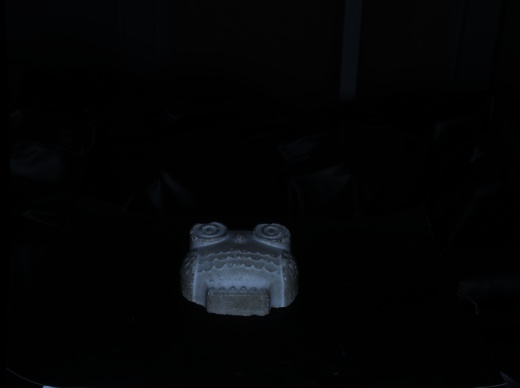}
\includegraphics[height=0.12\textwidth]{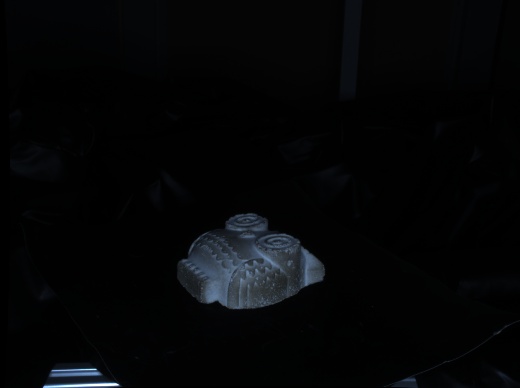}
\includegraphics[height=0.12\textwidth]{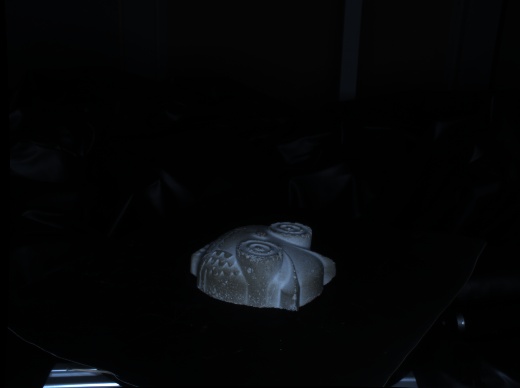}
\includegraphics[height=0.12\textwidth]{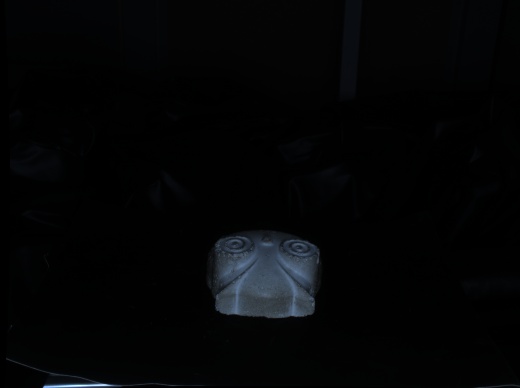}
\includegraphics[height=0.12\textwidth]{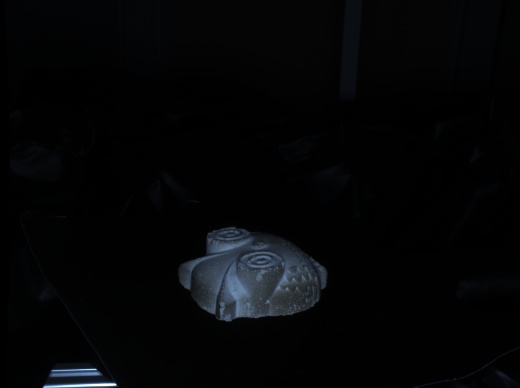}
\includegraphics[height=0.12\textwidth]{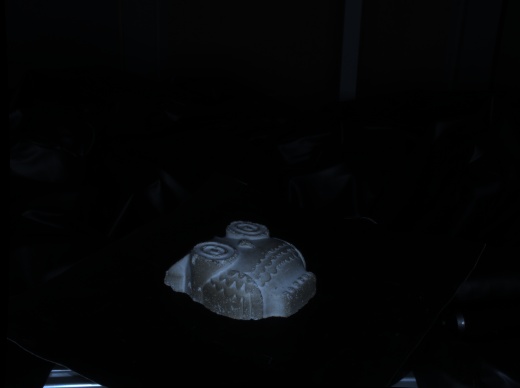}
\includegraphics[height=0.12\textwidth]{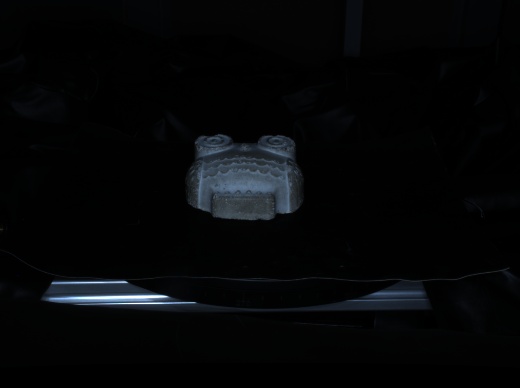}
\includegraphics[height=0.12\textwidth]{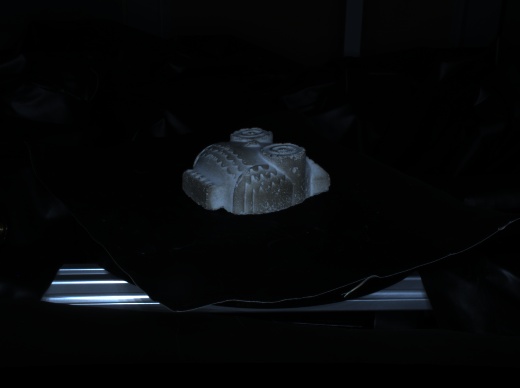}
\includegraphics[height=0.12\textwidth]{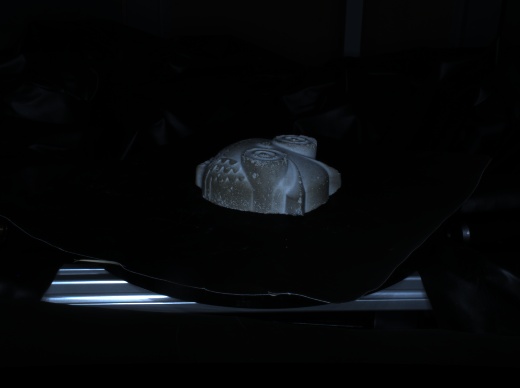}
\includegraphics[height=0.12\textwidth]{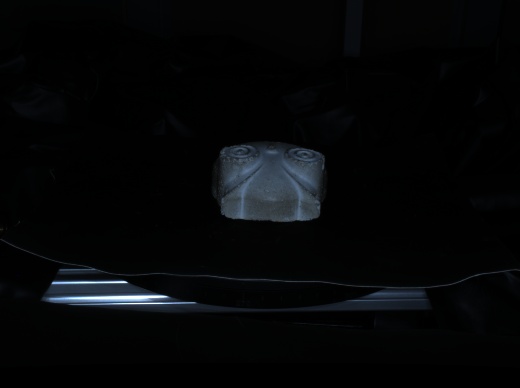}
\includegraphics[height=0.12\textwidth]{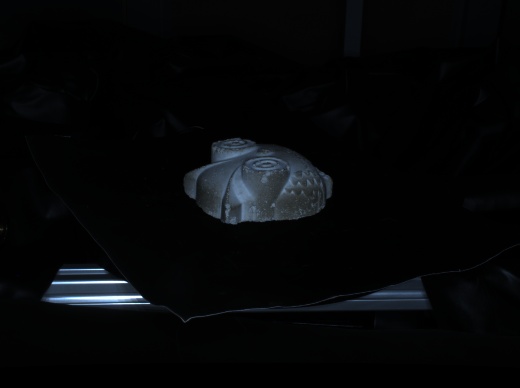}
\includegraphics[height=0.12\textwidth]{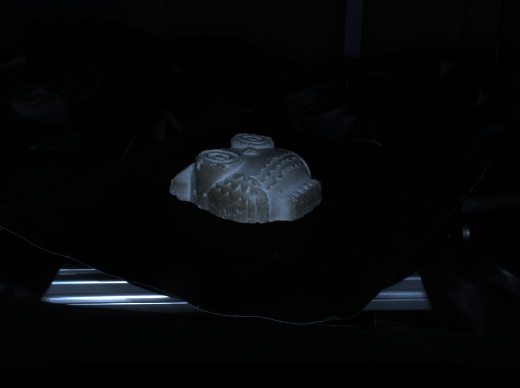}
\includegraphics[height=0.12\textwidth]{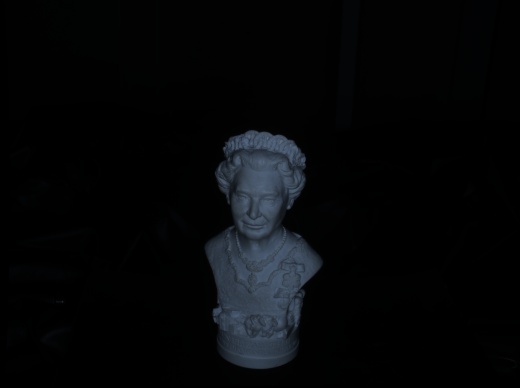}
\includegraphics[height=0.12\textwidth]{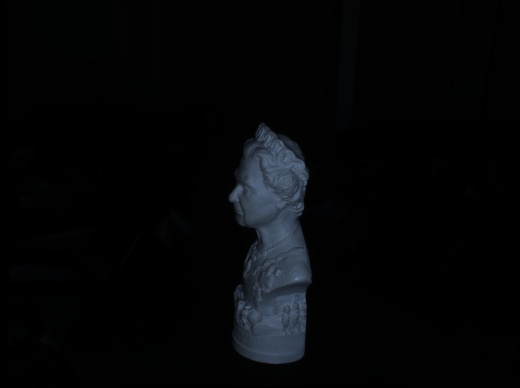}
\includegraphics[height=0.12\textwidth]{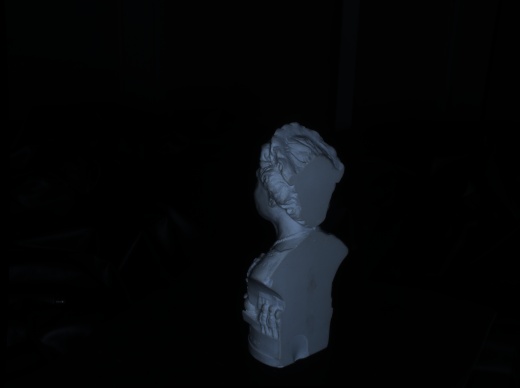}
\includegraphics[height=0.12\textwidth]{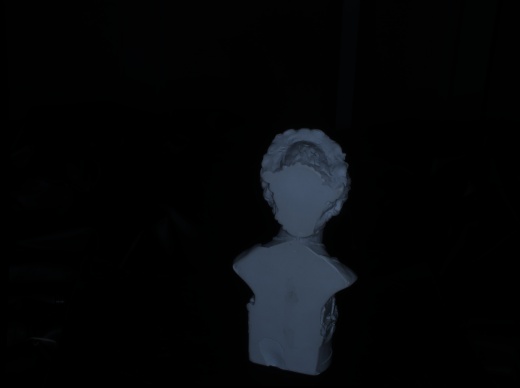}
\includegraphics[height=0.12\textwidth]{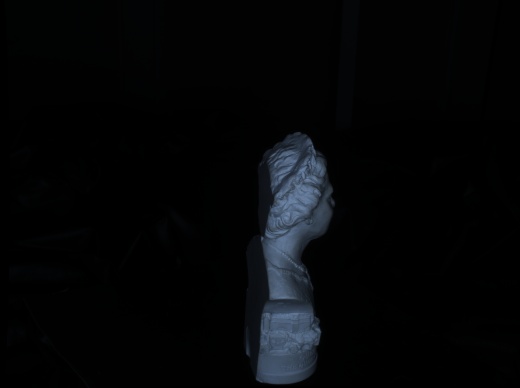}
\includegraphics[height=0.12\textwidth]{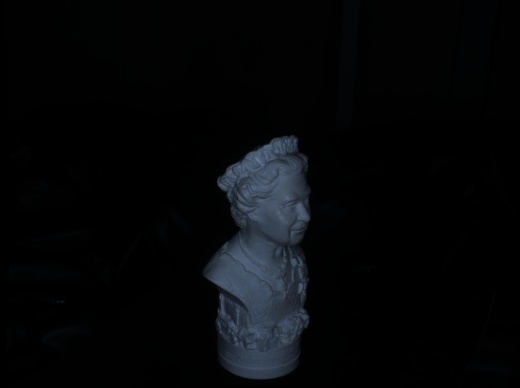}
\includegraphics[height=0.12\textwidth]{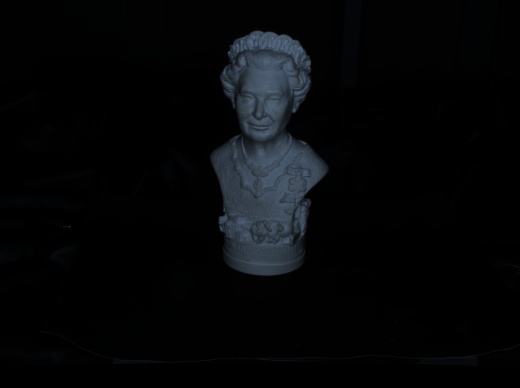}
\includegraphics[height=0.12\textwidth]{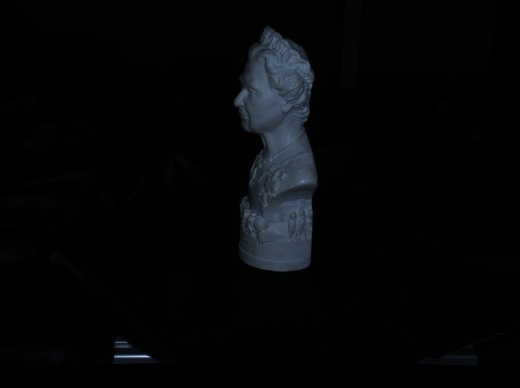}
\includegraphics[height=0.12\textwidth]{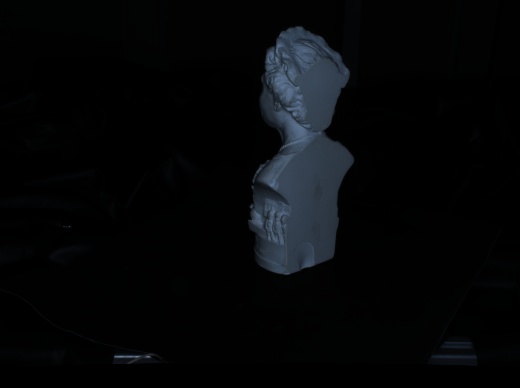}
\includegraphics[height=0.12\textwidth]{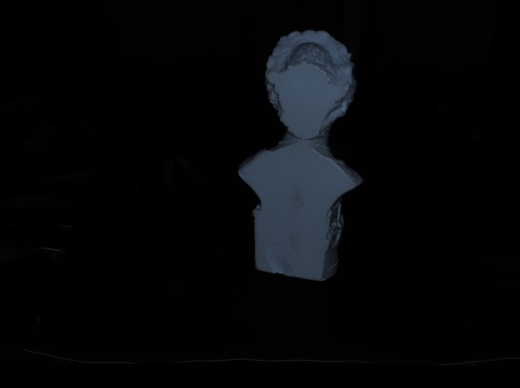}
\includegraphics[height=0.12\textwidth]{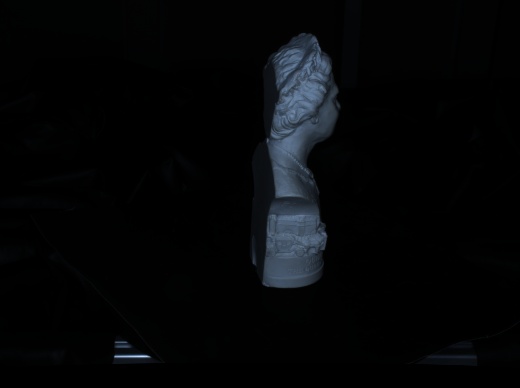}
\includegraphics[height=0.12\textwidth]{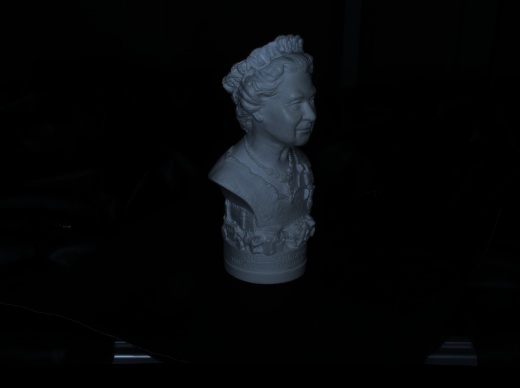}
\includegraphics[height=0.12\textwidth]{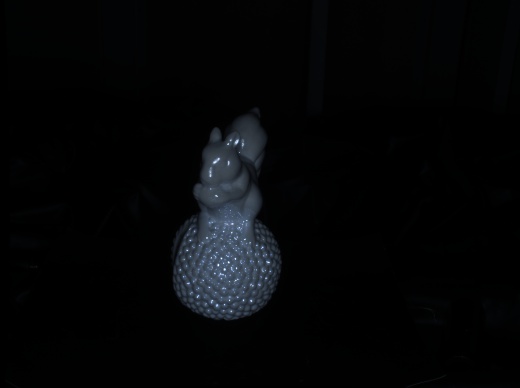}
\includegraphics[height=0.12\textwidth]{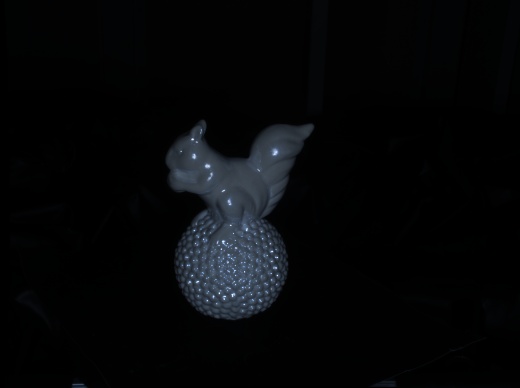}
\includegraphics[height=0.12\textwidth]{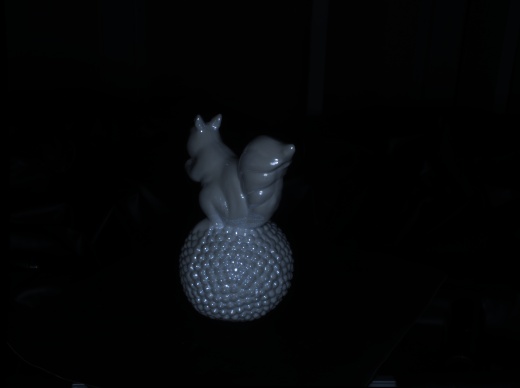}
\includegraphics[height=0.12\textwidth]{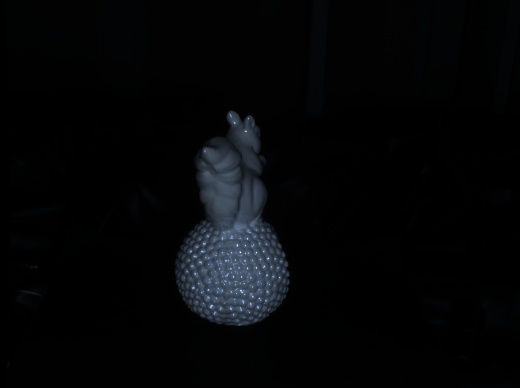}
\includegraphics[height=0.12\textwidth]{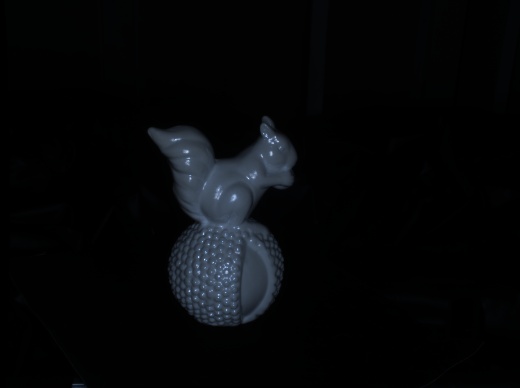}
\includegraphics[height=0.12\textwidth]{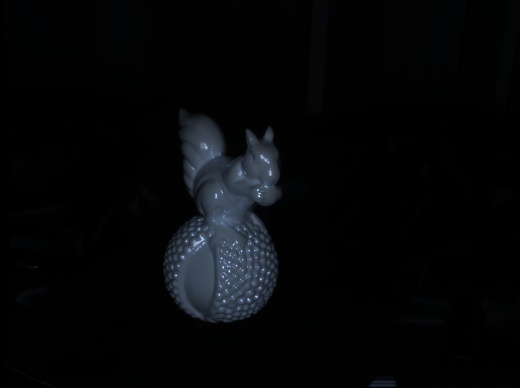}
\includegraphics[height=0.12\textwidth]{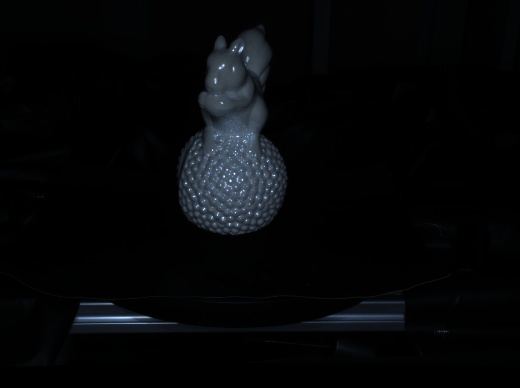}
\includegraphics[height=0.12\textwidth]{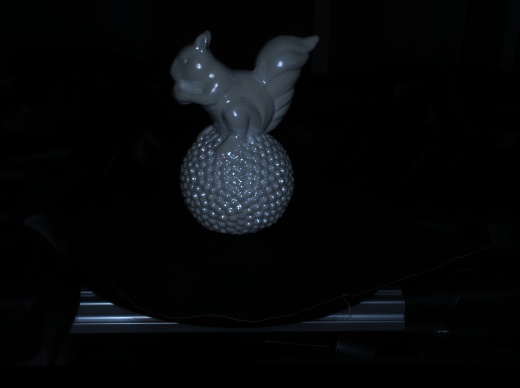}
\includegraphics[height=0.12\textwidth]{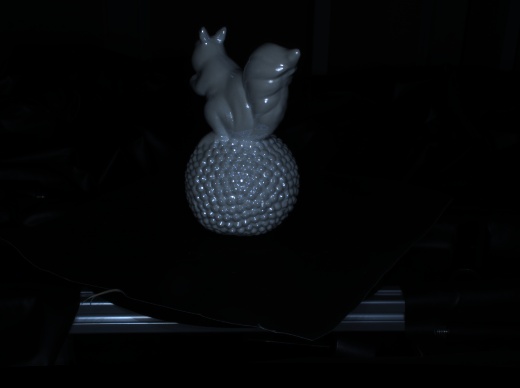}
\includegraphics[height=0.12\textwidth]{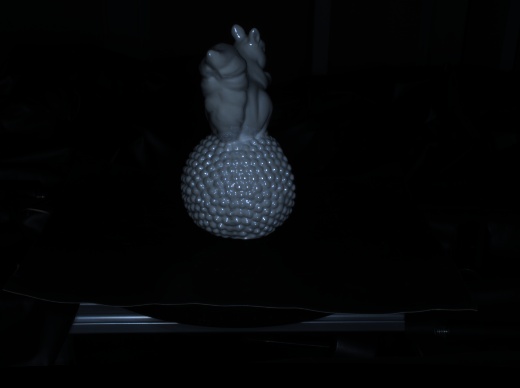}
\includegraphics[height=0.12\textwidth]{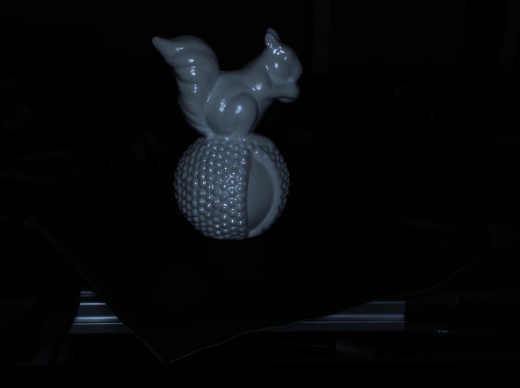}
\includegraphics[height=0.12\textwidth]{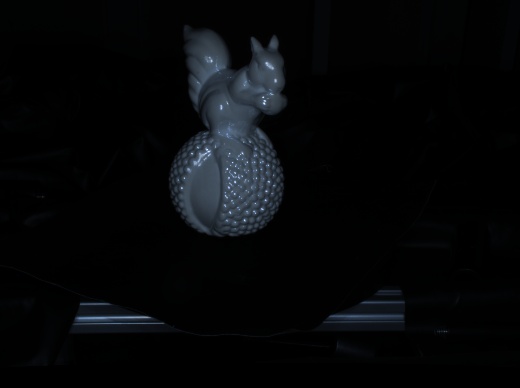}

 \caption{Average PS image for all views for \textit{Hippo}, \textit{House}, \textit{Owl}, \textit{Queen}, \textit{Squirrel}.  }
 \label{fig:obs2}
\end{figure*}

\begin{figure*}[t]
\centering
\includegraphics[height=0.12\textwidth]{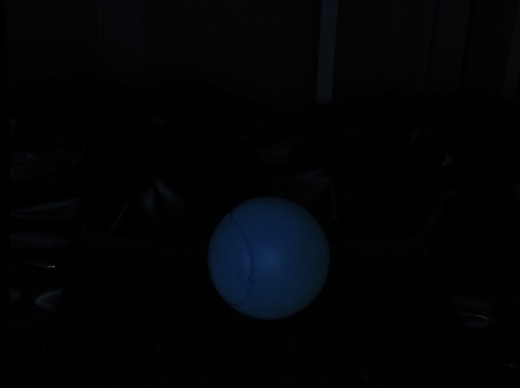}
\includegraphics[height=0.12\textwidth]{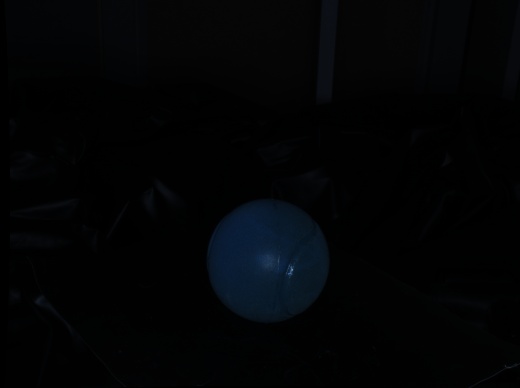}
\includegraphics[height=0.12\textwidth]{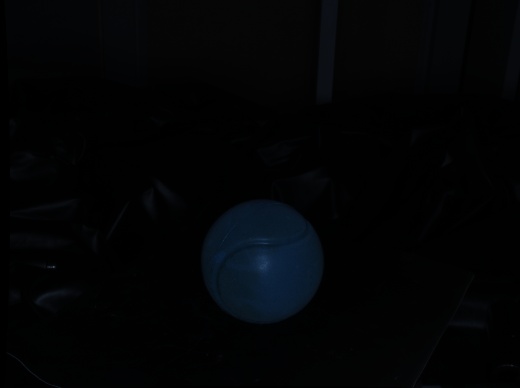}
\includegraphics[height=0.12\textwidth]{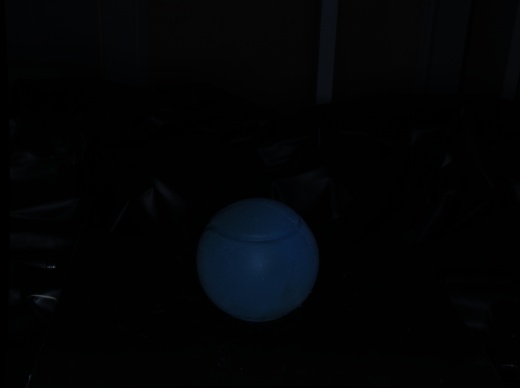}
\includegraphics[height=0.12\textwidth]{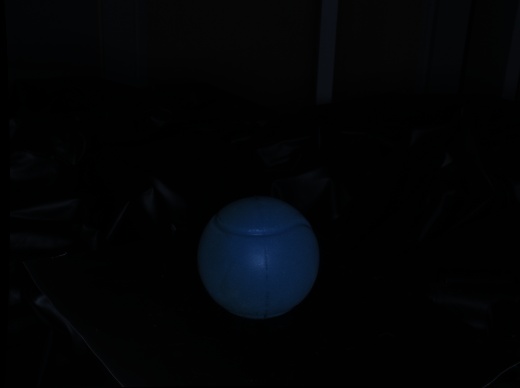}
\includegraphics[height=0.12\textwidth]{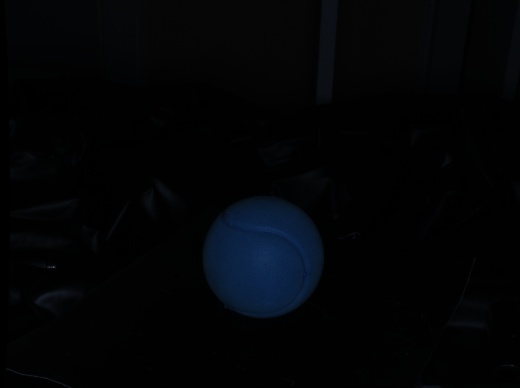}
\includegraphics[height=0.12\textwidth]{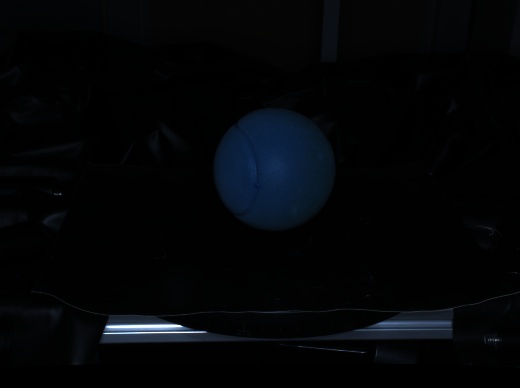}
\includegraphics[height=0.12\textwidth]{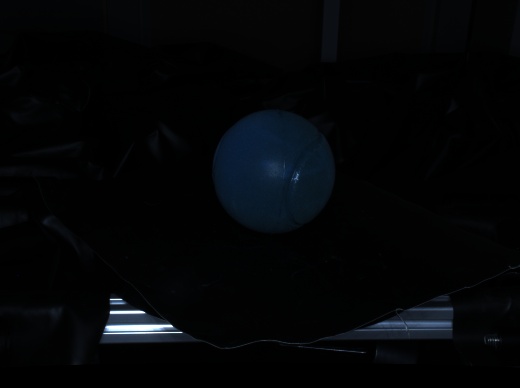}
\includegraphics[height=0.12\textwidth]{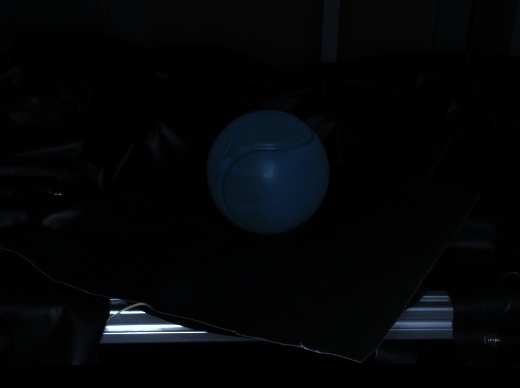}
\includegraphics[height=0.12\textwidth]{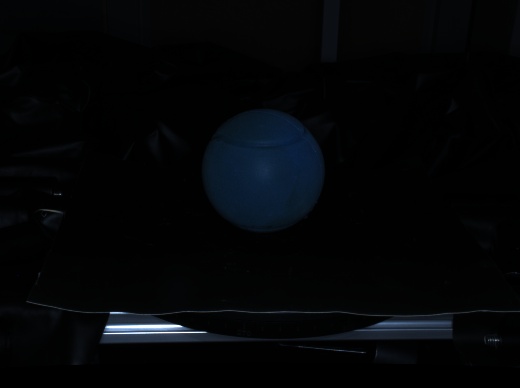}
\includegraphics[height=0.12\textwidth]{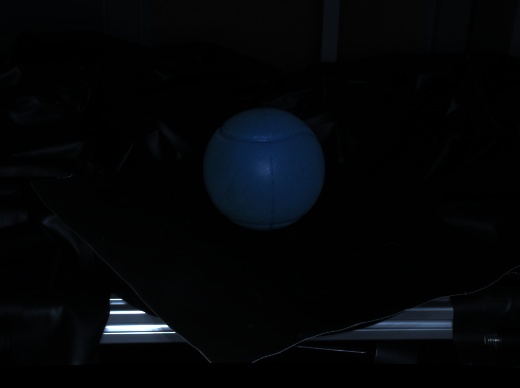}
\includegraphics[height=0.12\textwidth]{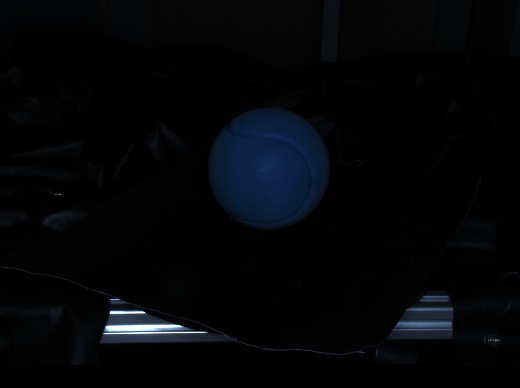}
\includegraphics[height=0.12\textwidth]{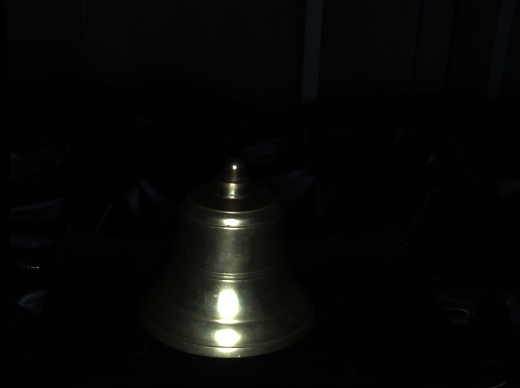}
\includegraphics[height=0.12\textwidth]{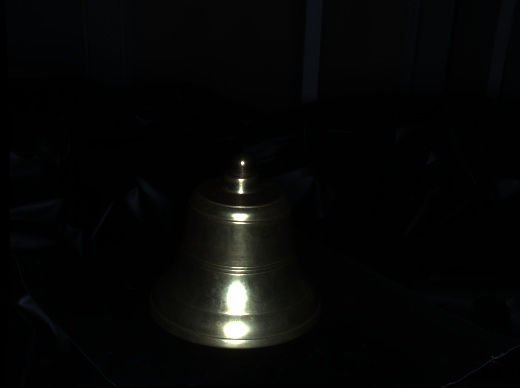}
\includegraphics[height=0.12\textwidth]{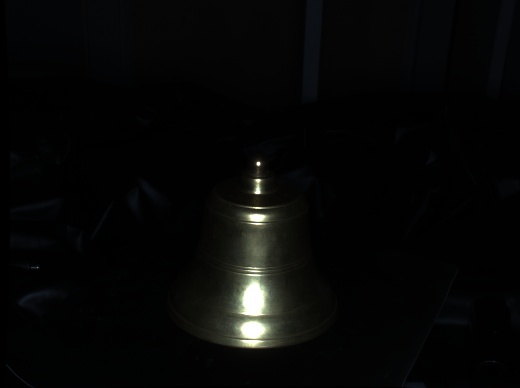}
\includegraphics[height=0.12\textwidth]{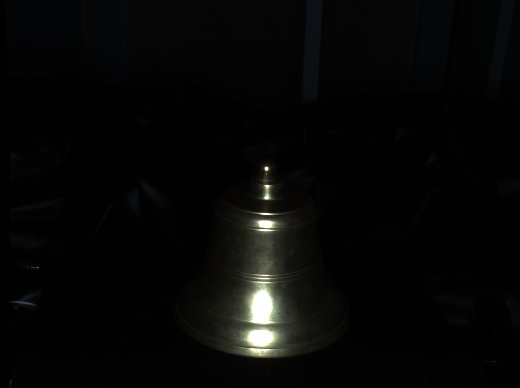}
\includegraphics[height=0.12\textwidth]{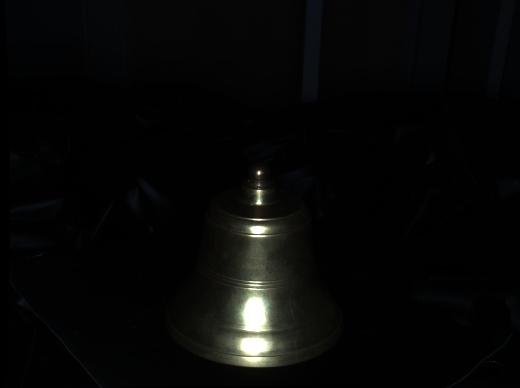}
\includegraphics[height=0.12\textwidth]{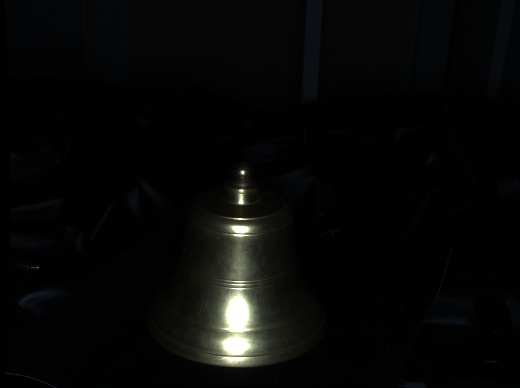}
\includegraphics[height=0.12\textwidth]{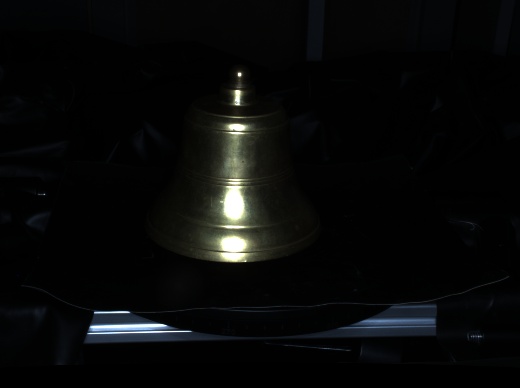}
\includegraphics[height=0.12\textwidth]{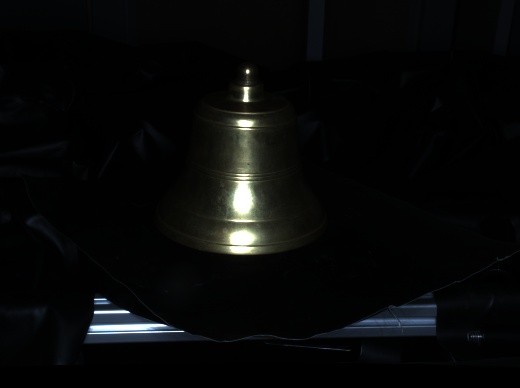}
\includegraphics[height=0.12\textwidth]{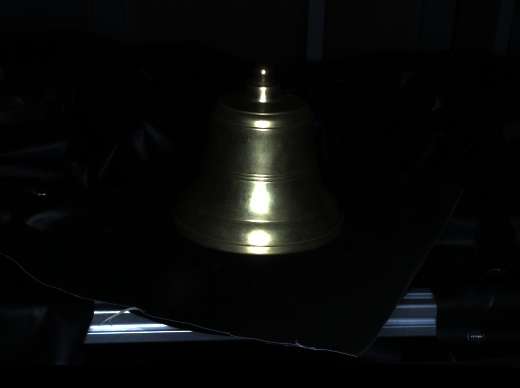}
\includegraphics[height=0.12\textwidth]{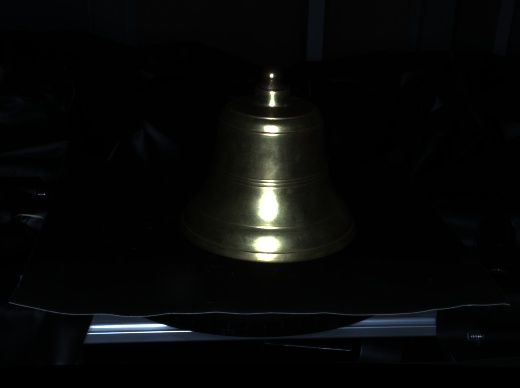}
\includegraphics[height=0.12\textwidth]{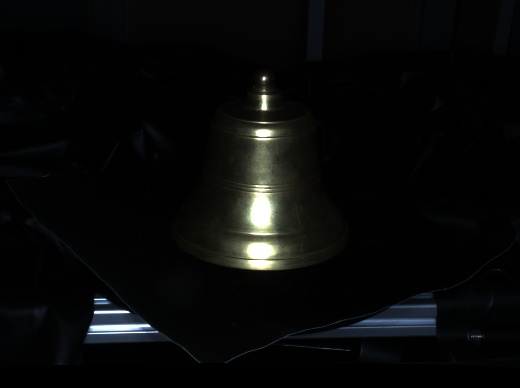}
\includegraphics[height=0.12\textwidth]{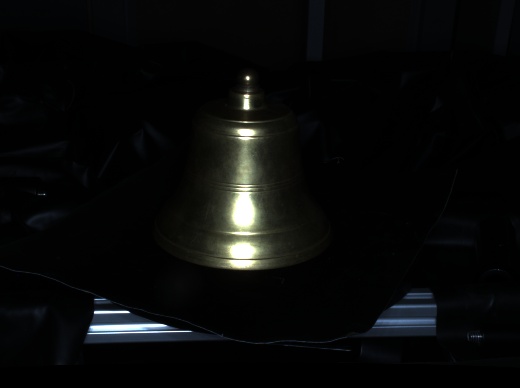}
\includegraphics[height=0.12\textwidth]{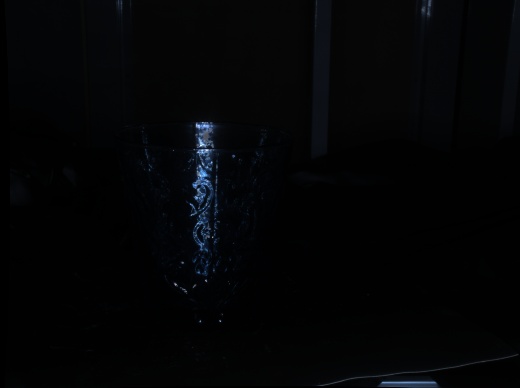}
\includegraphics[height=0.12\textwidth]{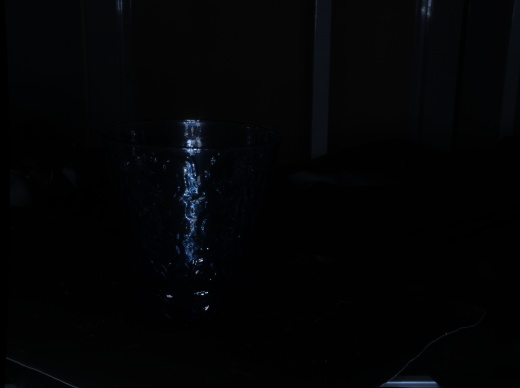}
\includegraphics[height=0.12\textwidth]{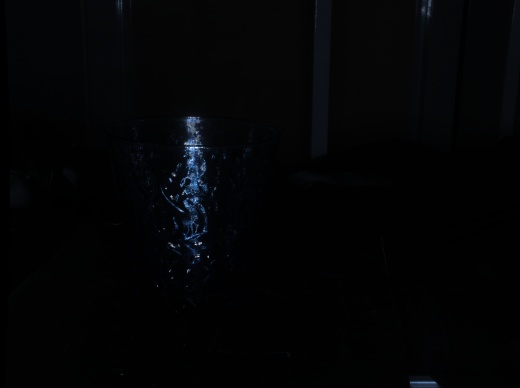}
\includegraphics[height=0.12\textwidth]{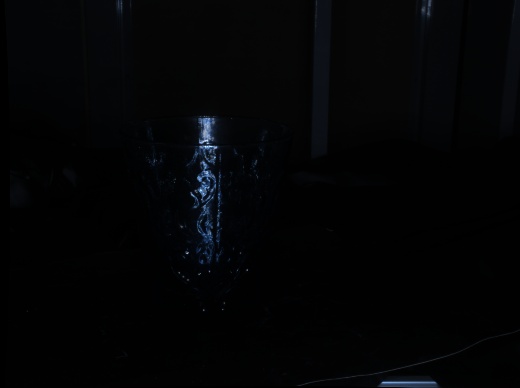}
\includegraphics[height=0.12\textwidth]{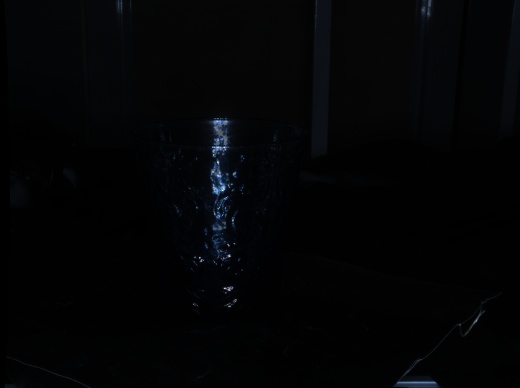}
\includegraphics[height=0.12\textwidth]{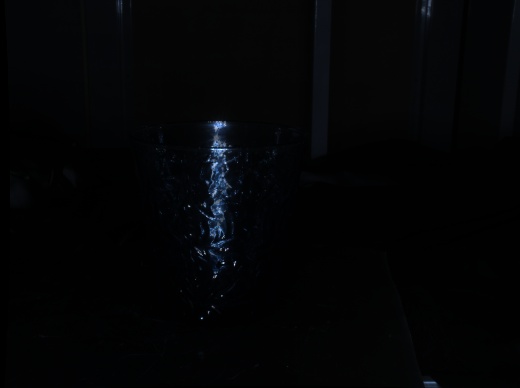}
\includegraphics[height=0.12\textwidth]{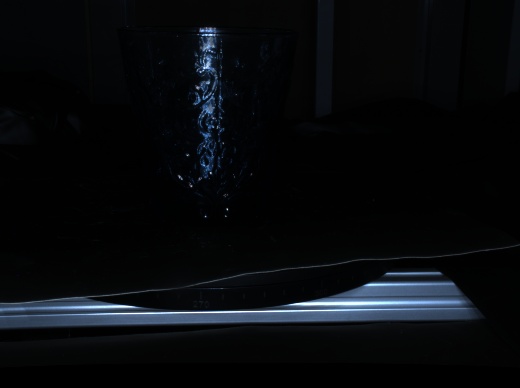}
\includegraphics[height=0.12\textwidth]{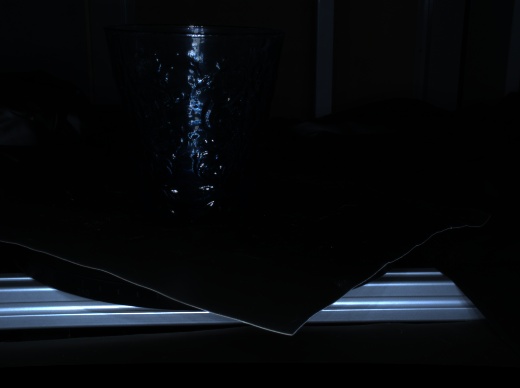}
\includegraphics[height=0.12\textwidth]{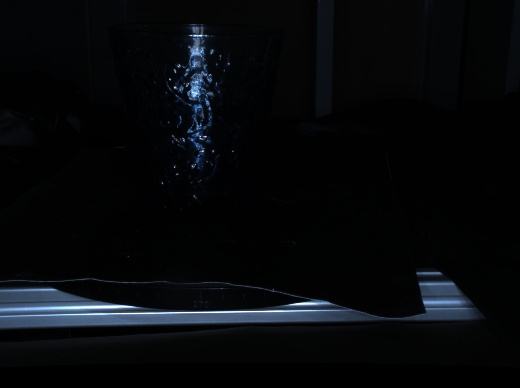}
\includegraphics[height=0.12\textwidth]{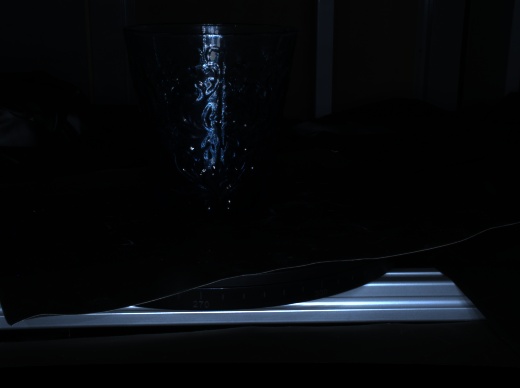}
\includegraphics[height=0.12\textwidth]{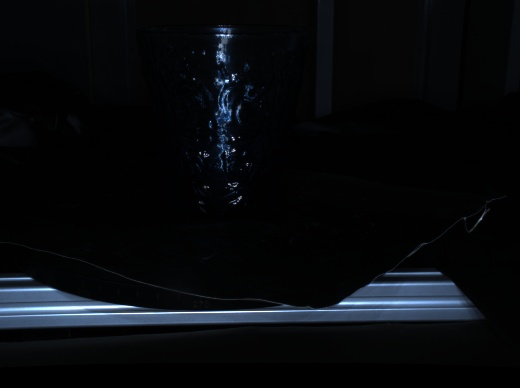}
\includegraphics[height=0.12\textwidth]{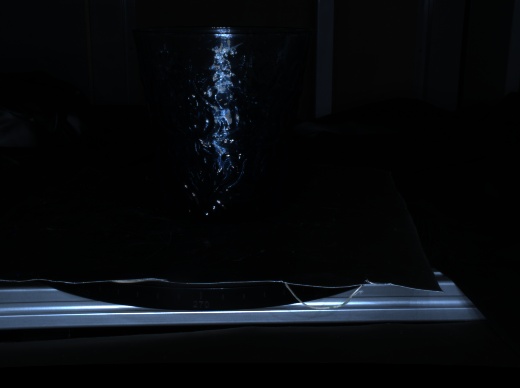}
\includegraphics[height=0.12\textwidth]{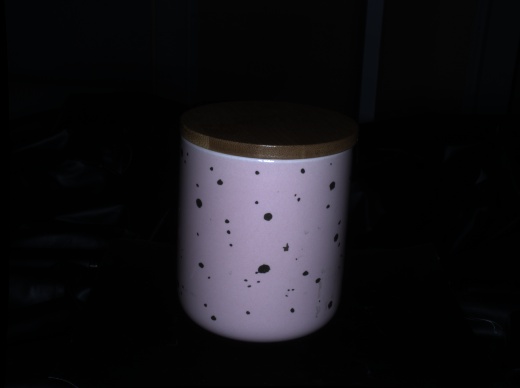}
\includegraphics[height=0.12\textwidth]{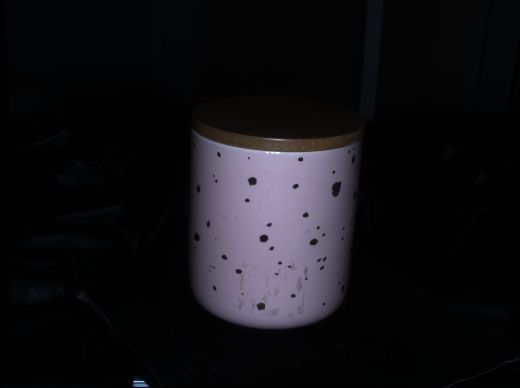}
\includegraphics[height=0.12\textwidth]{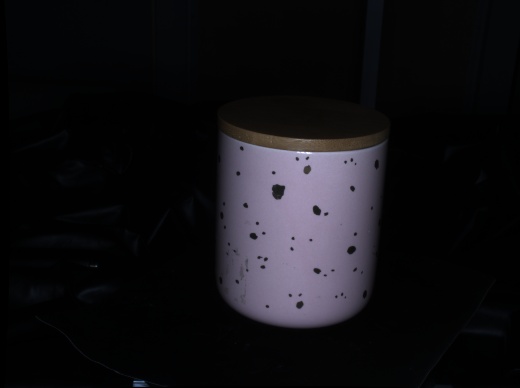}
\includegraphics[height=0.12\textwidth]{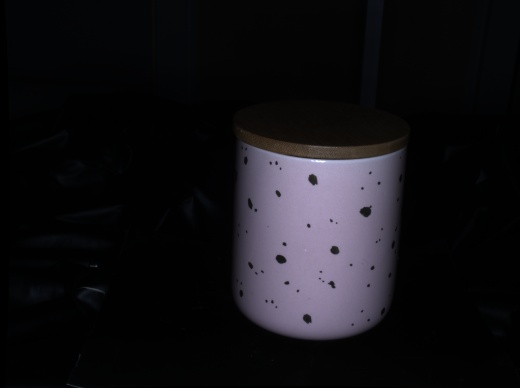}
\includegraphics[height=0.12\textwidth]{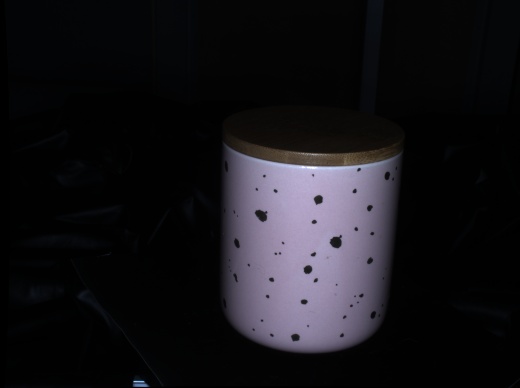}
\includegraphics[height=0.12\textwidth]{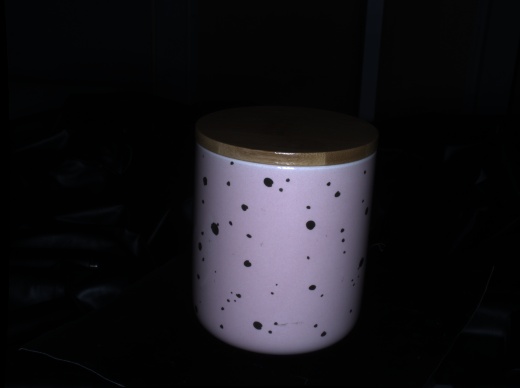}
\includegraphics[height=0.12\textwidth]{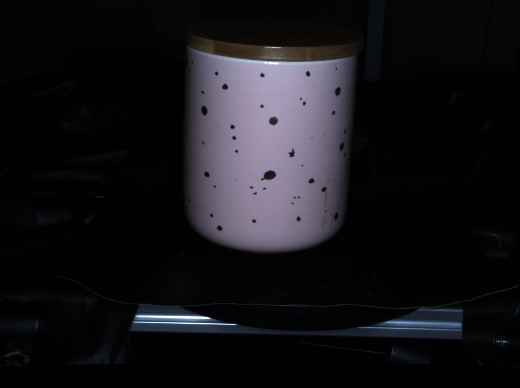}
\includegraphics[height=0.12\textwidth]{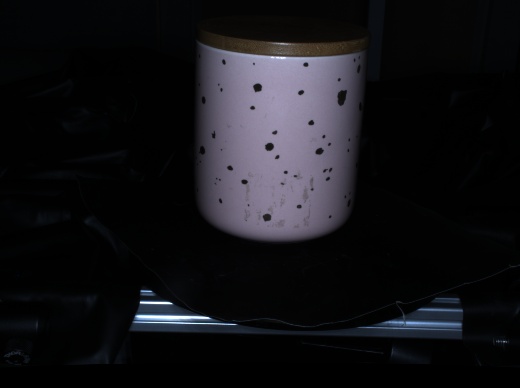}
\includegraphics[height=0.12\textwidth]{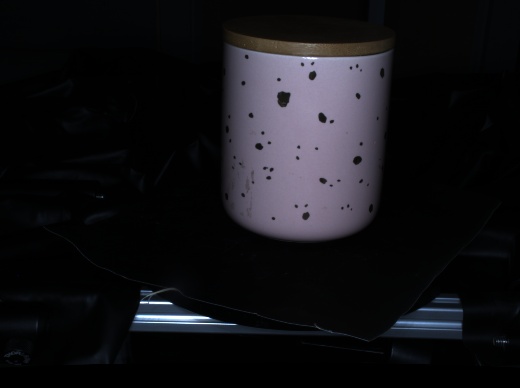}
\includegraphics[height=0.12\textwidth]{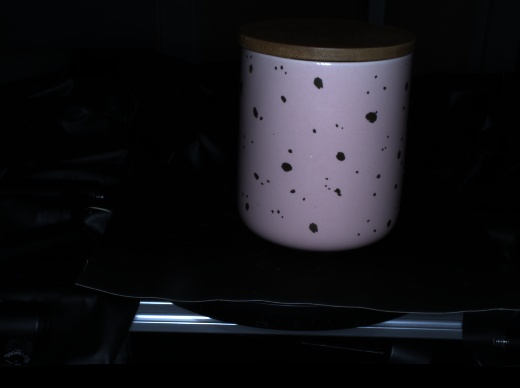}
\includegraphics[height=0.12\textwidth]{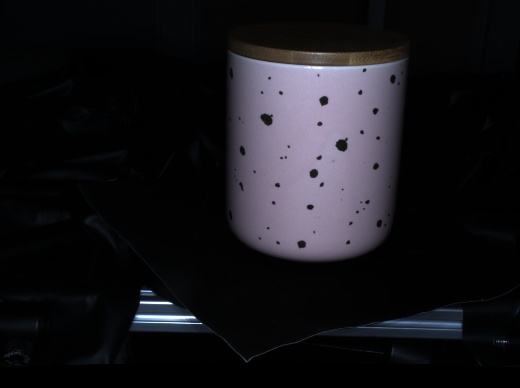}
\includegraphics[height=0.12\textwidth]{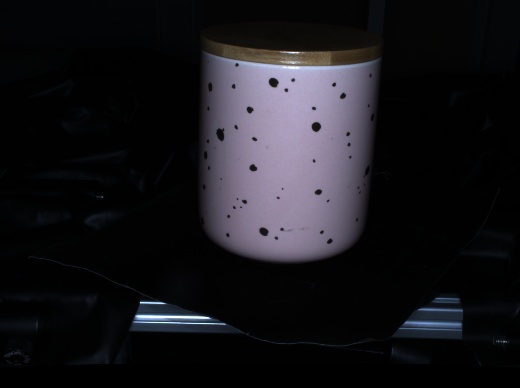}
\includegraphics[height=0.12\textwidth]{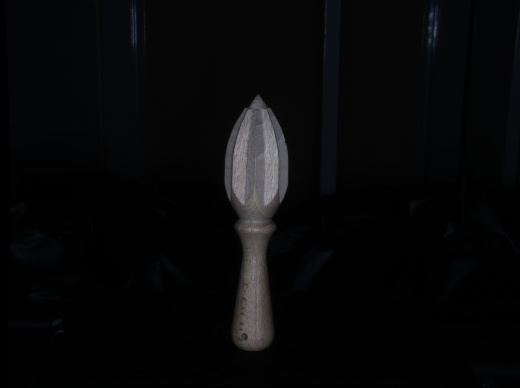}
\includegraphics[height=0.12\textwidth]{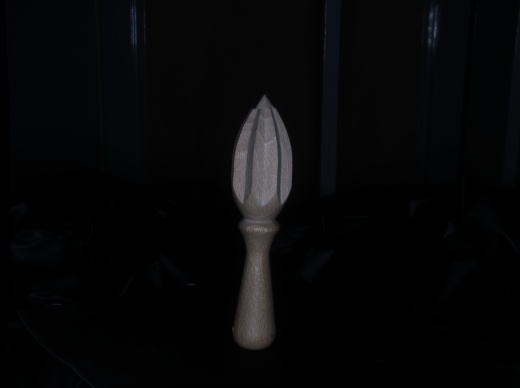}
\includegraphics[height=0.12\textwidth]{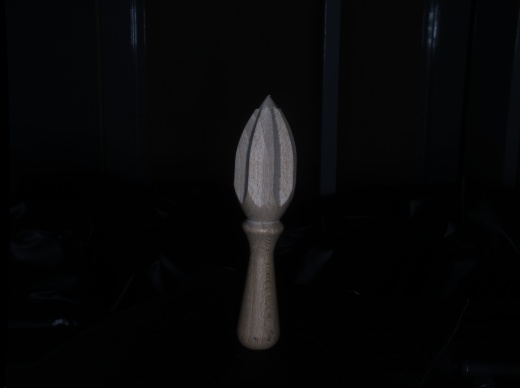}
\includegraphics[height=0.12\textwidth]{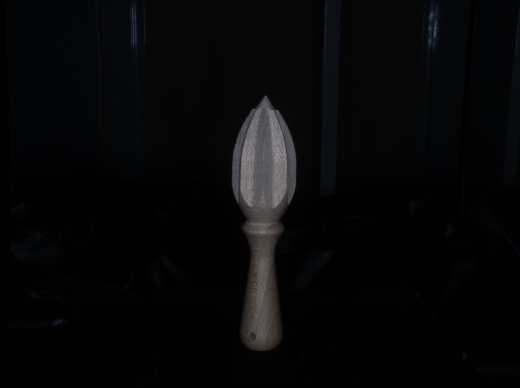}
\includegraphics[height=0.12\textwidth]{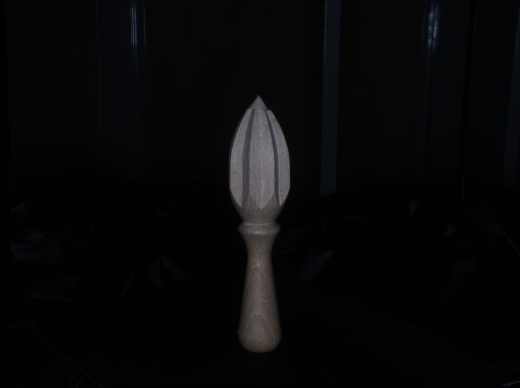}
\includegraphics[height=0.12\textwidth]{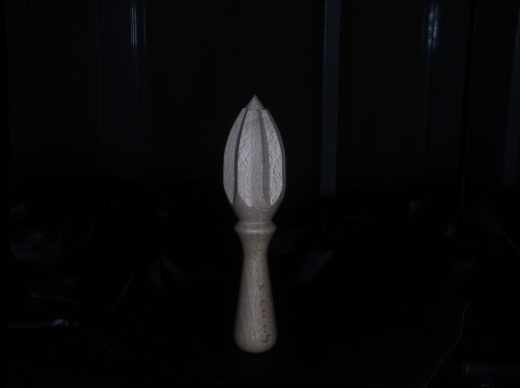}
\includegraphics[height=0.12\textwidth]{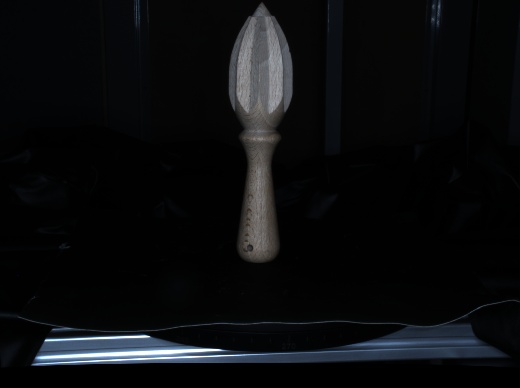}
\includegraphics[height=0.12\textwidth]{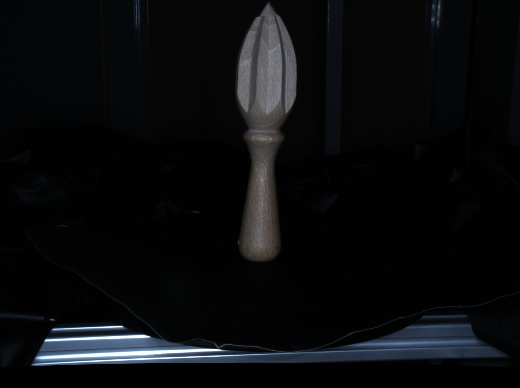}
\includegraphics[height=0.12\textwidth]{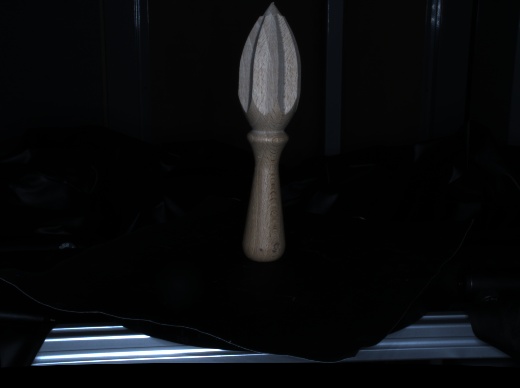}
\includegraphics[height=0.12\textwidth]{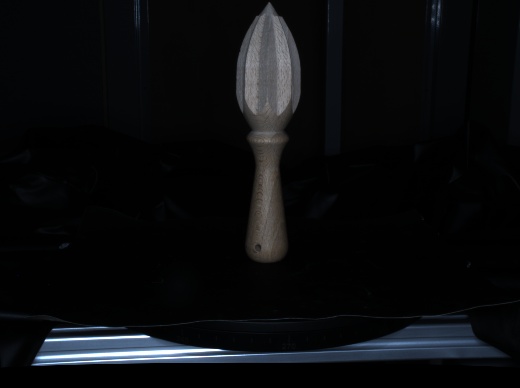}
\includegraphics[height=0.12\textwidth]{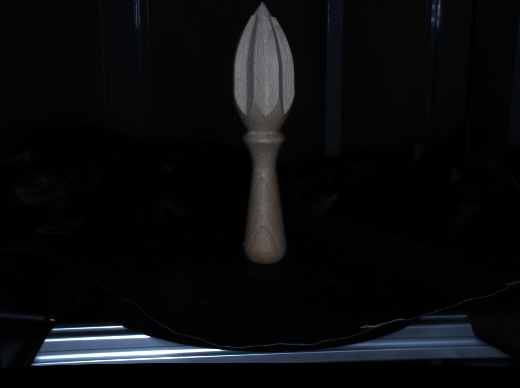}
\includegraphics[height=0.12\textwidth]{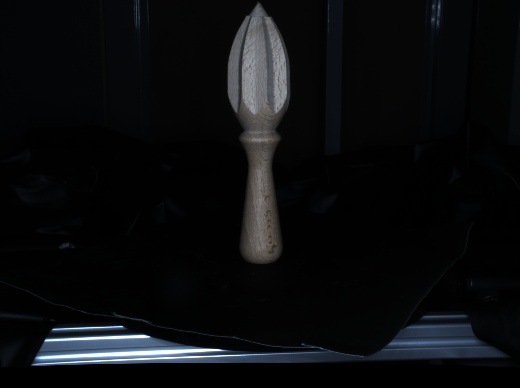}
 \caption{Average PS image for all views for \textit{Ball}, \textit{Bell}, \textit{Glass}, \textit{Jar}, \textit{Tool}.  }
 \label{fig:obs3}
\end{figure*}

Figure~\ref{fig:additionalgt} shows the ground truth meshes of the unposed objects namely \textit{Ball}, \textit{Bell}, \textit{Glass}, \textit{Jar},\textit{Tool}. Preview images (average photometric stereo image) for all objects in all views are shown in Figures~\ref{fig:obs1}~\ref{fig:obs2} and ~\ref{fig:obs3}. For all objects, the stereo pair views are grouped vertically for a total of 6 views (every $60^0$). The approximate real world dimensions of all objects are shown in Table~\ref{tab:objsizes}. 

\begin{table*}[t]
\begin{center}
\resizebox{1.0\textwidth}{!}{%
\begin{tabular}{ | c |c c c c c  |} 
 \hline
Object  & Bowl & Buddha & Bunny & Cup & Die \\  \hline
Dimensions (mm)  & 50$\times$30$\times$70 & 85$\times$200$\times$170 & 50$\times$85$\times$90 &  60$\times$70$\times$70 &  65$\times$65$\times$65 \\  \hline
Object  &  Hippo & House & Owl & Queen & Squirrel \\  \hline
Dimensions (mm)  & 40$\times$50$\times$95 &105$\times$80$\times$70 & 65$\times$45$\times$80 & 60$\times$60$\times$135 & 50$\times$70$\times$115 \\  \hline
Object  & Ball & Bell & Glass &  Jar & Tool \\  \hline
Dimensions (mm)  & 70$\times$70$\times$70 & 105$\times$105$\times$110 & 90$\times$90$\times$100 & 100$\times$100$\times$115 & 30$\times$30$\times$140 \\  \hline
\end{tabular}
} 
\end{center}
\caption{This table provides approximate dimensions of the objects used in LUCES-MV. Note that precise sizes of objects can be computed directly from the ground truth meshes provided.}
\label{tab:objsizes} 
\end{table*}





\subsection{Intrinsics calibration images}
Intrinsic parameter calibration was performed using a the standard OpenCV caliration toolbox and a $10\times9$ calibration pattern from \url{https://calib.io/}. 420 images are captured to ensure orientation, location and scale variation as shown in Figure~\ref{fig:calib}. These images will be made available as a part of files representing the dataset.
\subsection{Light calibration images}
\begin{figure*}[t]
\centering
\includegraphics[height=0.14\textwidth]{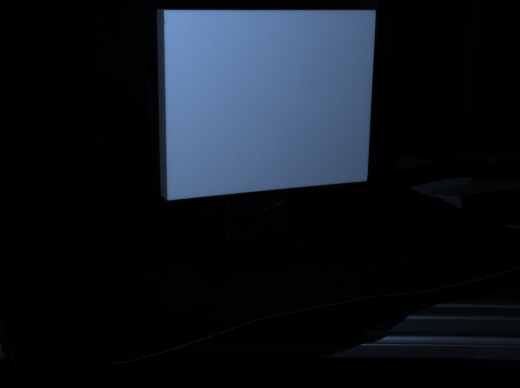}
\includegraphics[height=0.14\textwidth]{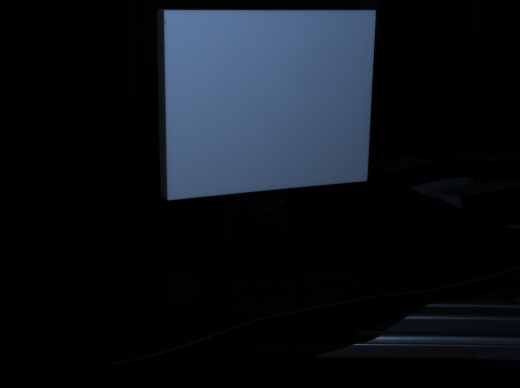}
\includegraphics[height=0.14\textwidth]{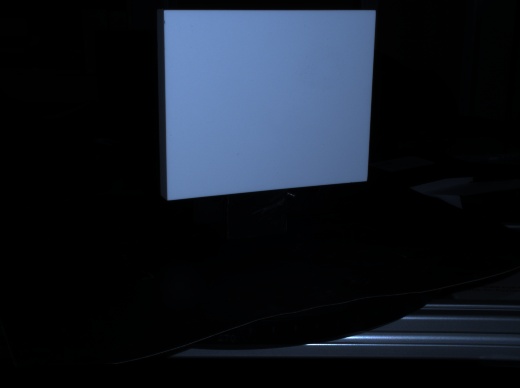}
\includegraphics[height=0.14\textwidth]{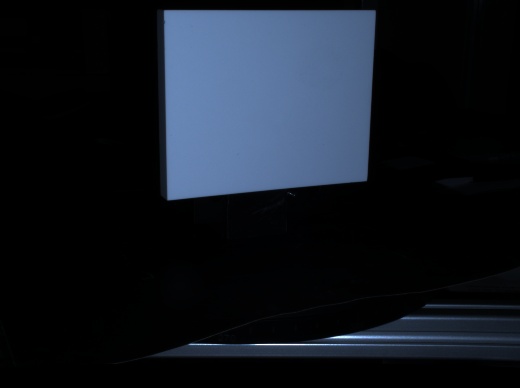}
\includegraphics[height=0.14\textwidth]{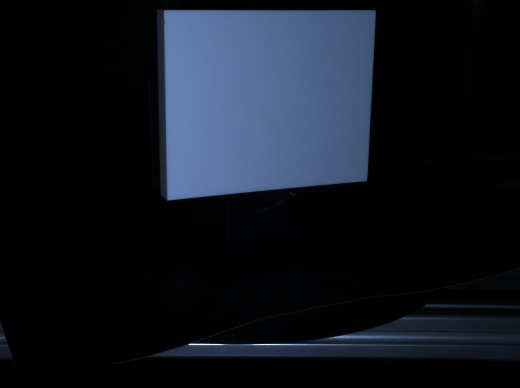}
\includegraphics[height=0.14\textwidth]{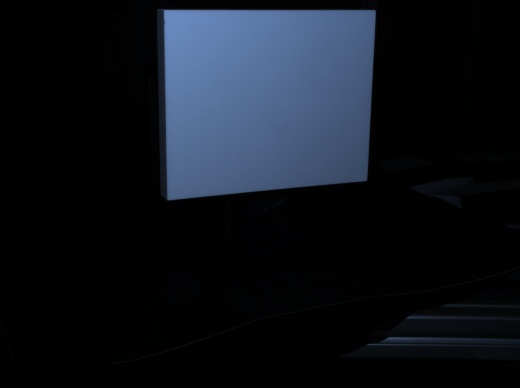}
\includegraphics[height=0.14\textwidth]{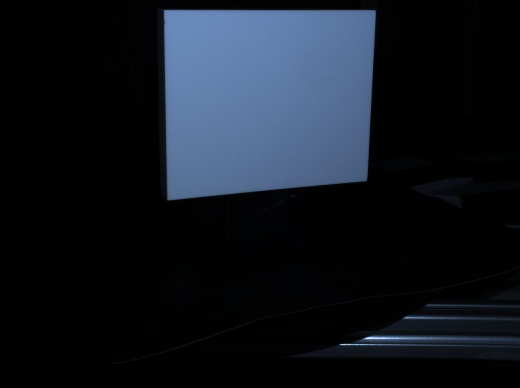}
\includegraphics[height=0.14\textwidth]{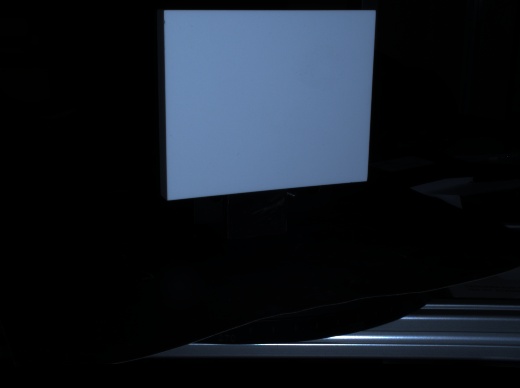}
\includegraphics[height=0.14\textwidth]{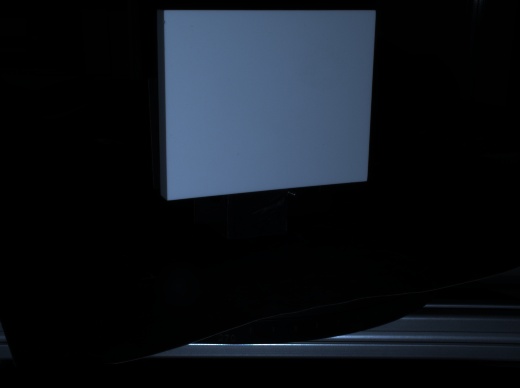}
\includegraphics[height=0.14\textwidth]{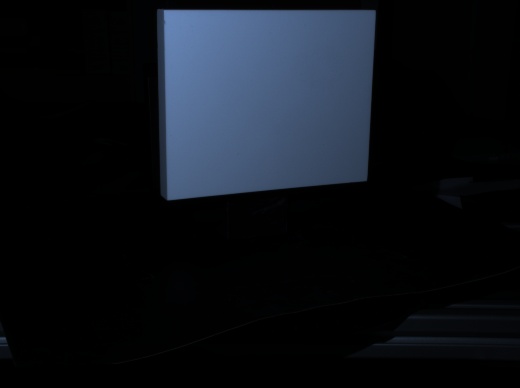}
\includegraphics[height=0.14\textwidth]{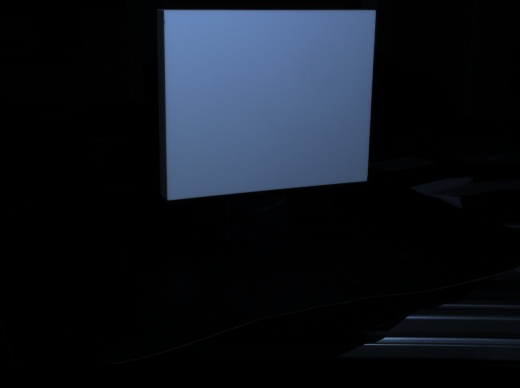}
\includegraphics[height=0.14\textwidth]{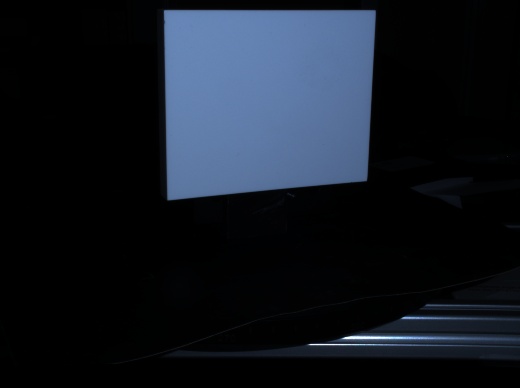}
\includegraphics[height=0.14\textwidth]{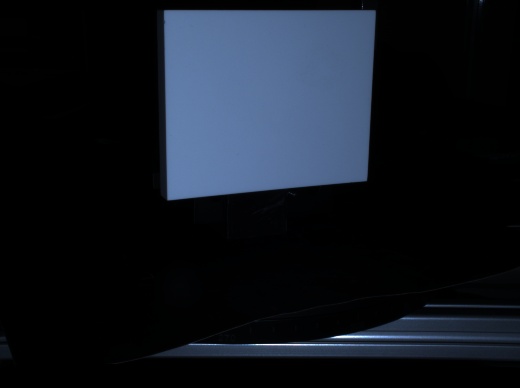}
\includegraphics[height=0.14\textwidth]{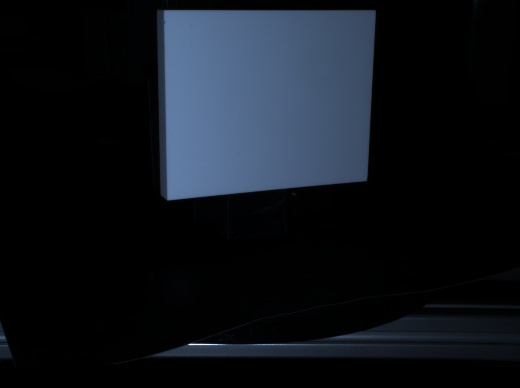}
\includegraphics[height=0.14\textwidth]{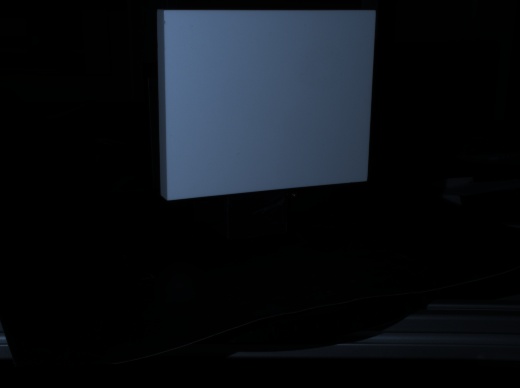}
\includegraphics[height=0.14\textwidth]{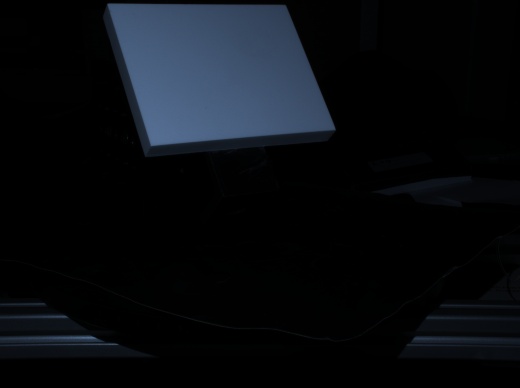}
\includegraphics[height=0.14\textwidth]{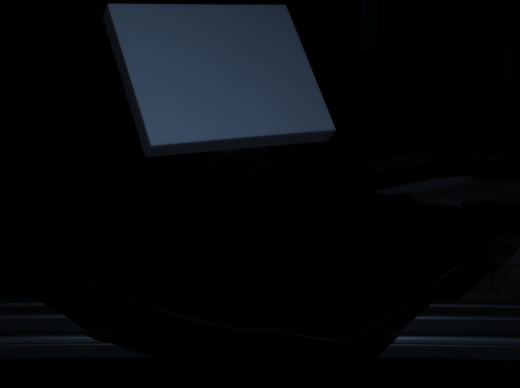}
\includegraphics[height=0.14\textwidth]{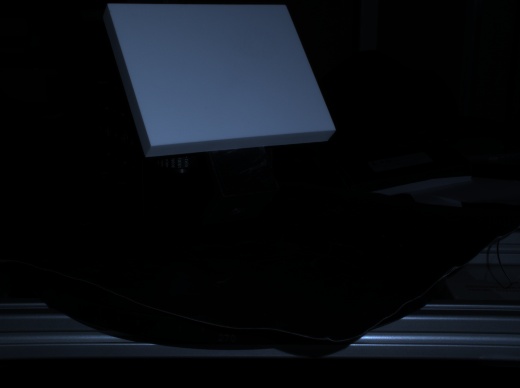}
\includegraphics[height=0.14\textwidth]{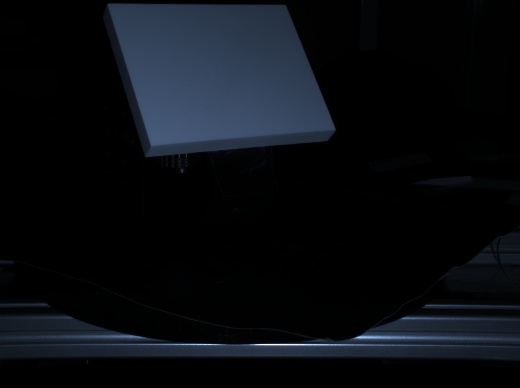}
\includegraphics[height=0.14\textwidth]{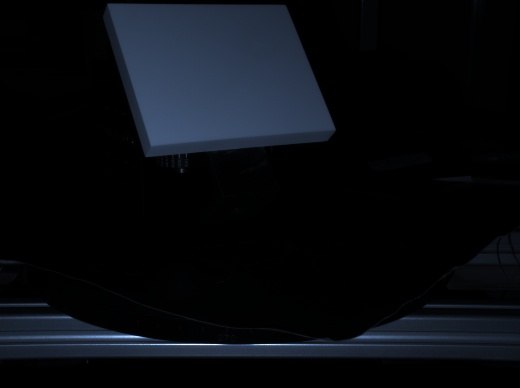}
\includegraphics[height=0.14\textwidth]{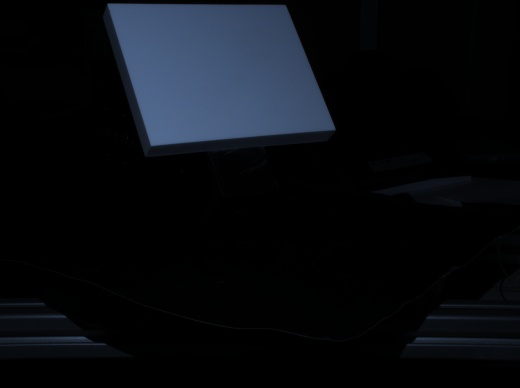}
\includegraphics[height=0.14\textwidth]{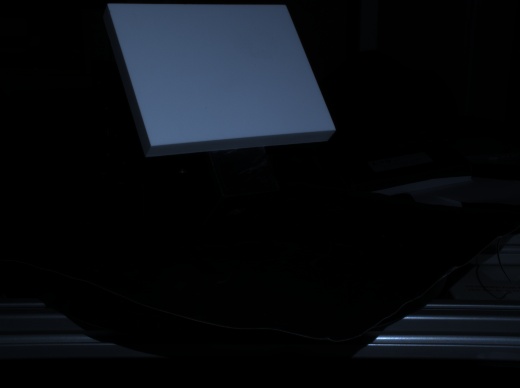}
\includegraphics[height=0.14\textwidth]{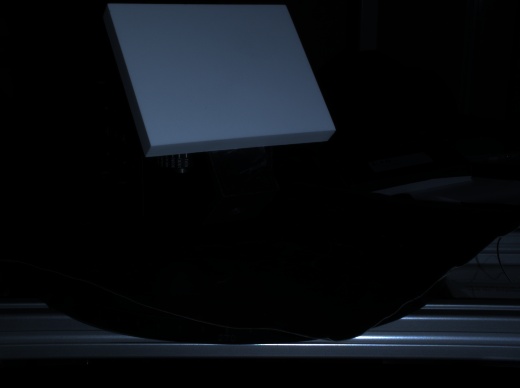}
\includegraphics[height=0.14\textwidth]{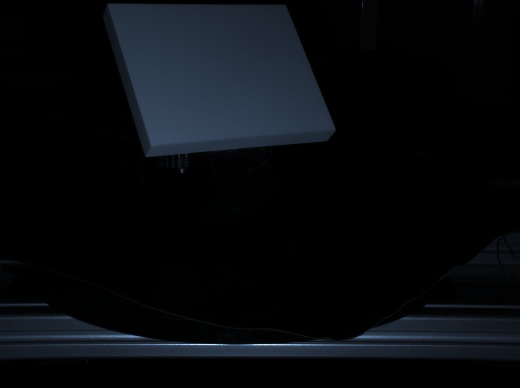}
\includegraphics[height=0.14\textwidth]{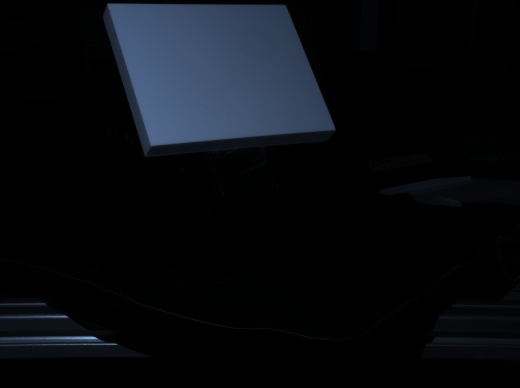}
\includegraphics[height=0.14\textwidth]{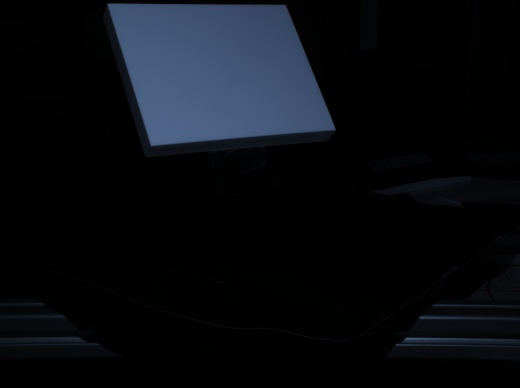}
\includegraphics[height=0.14\textwidth]{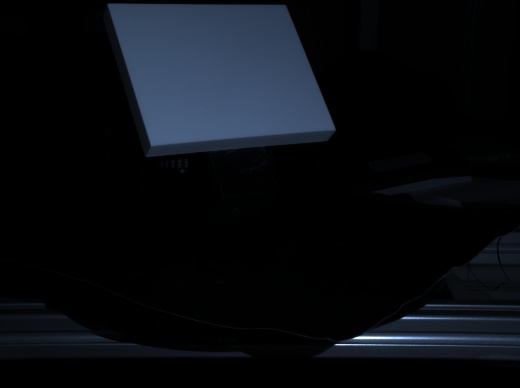}
\includegraphics[height=0.14\textwidth]{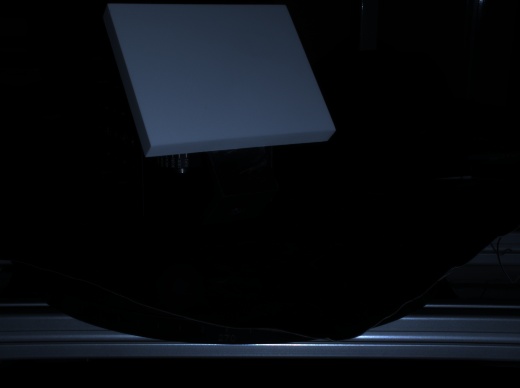}
\includegraphics[height=0.14\textwidth]{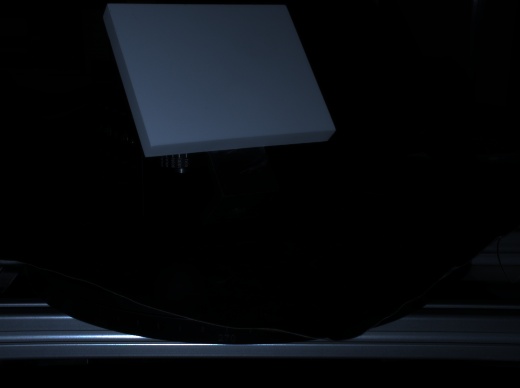}
\includegraphics[height=0.14\textwidth]{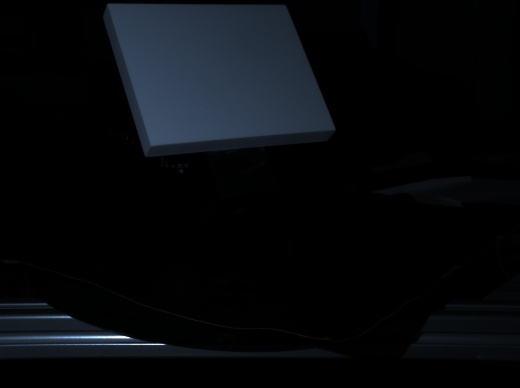}
\includegraphics[height=0.14\textwidth]{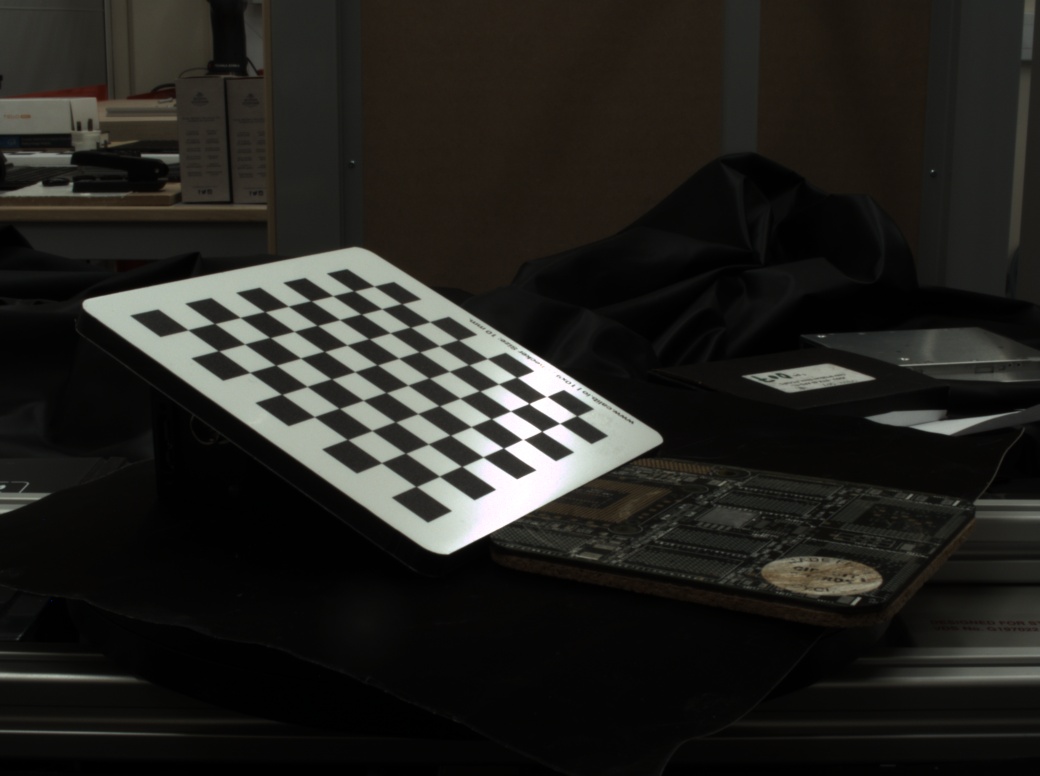}
\includegraphics[height=0.14\textwidth]{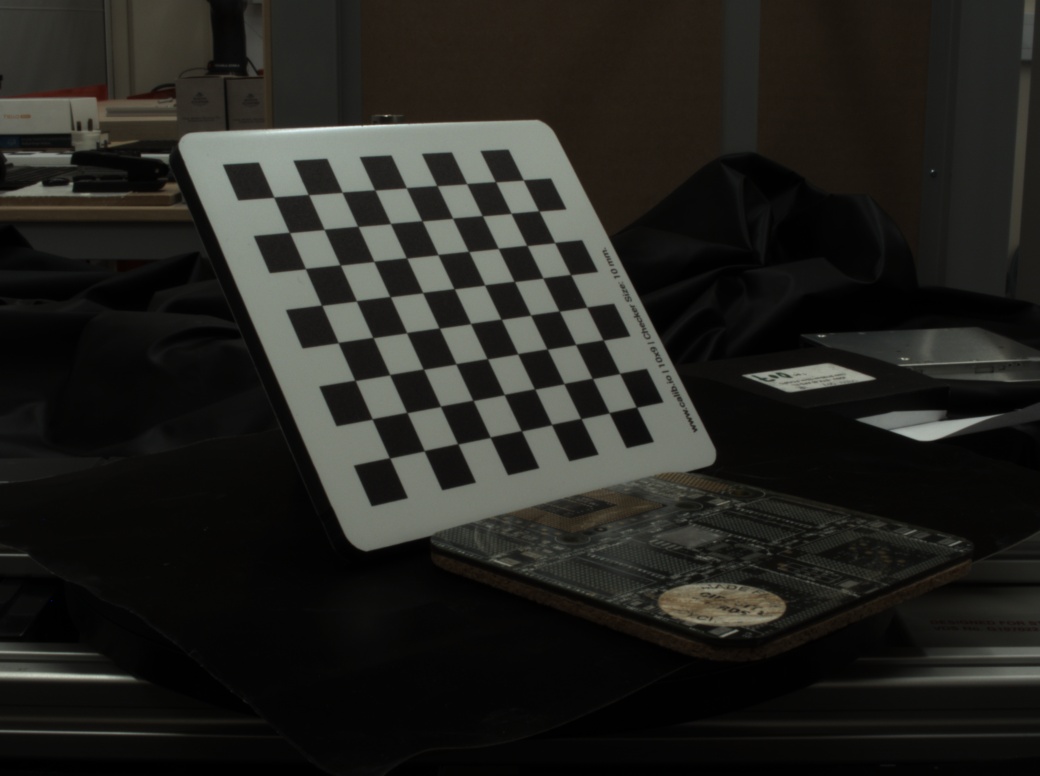}
\includegraphics[height=0.14\textwidth]{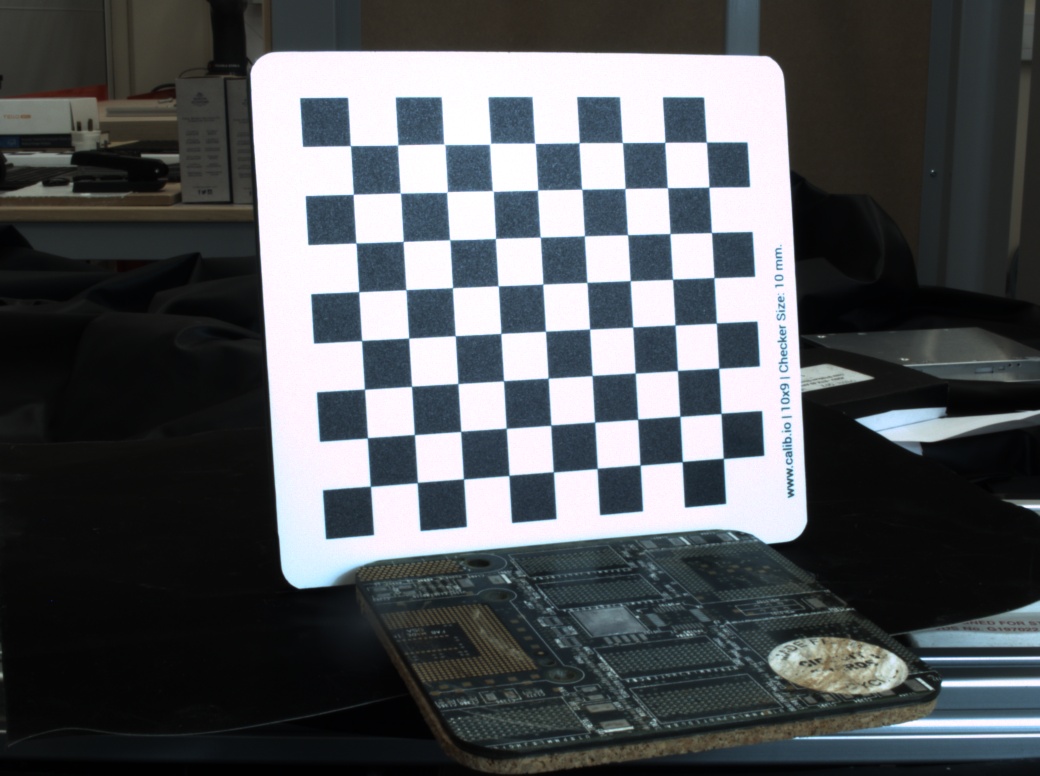}
\includegraphics[height=0.14\textwidth]{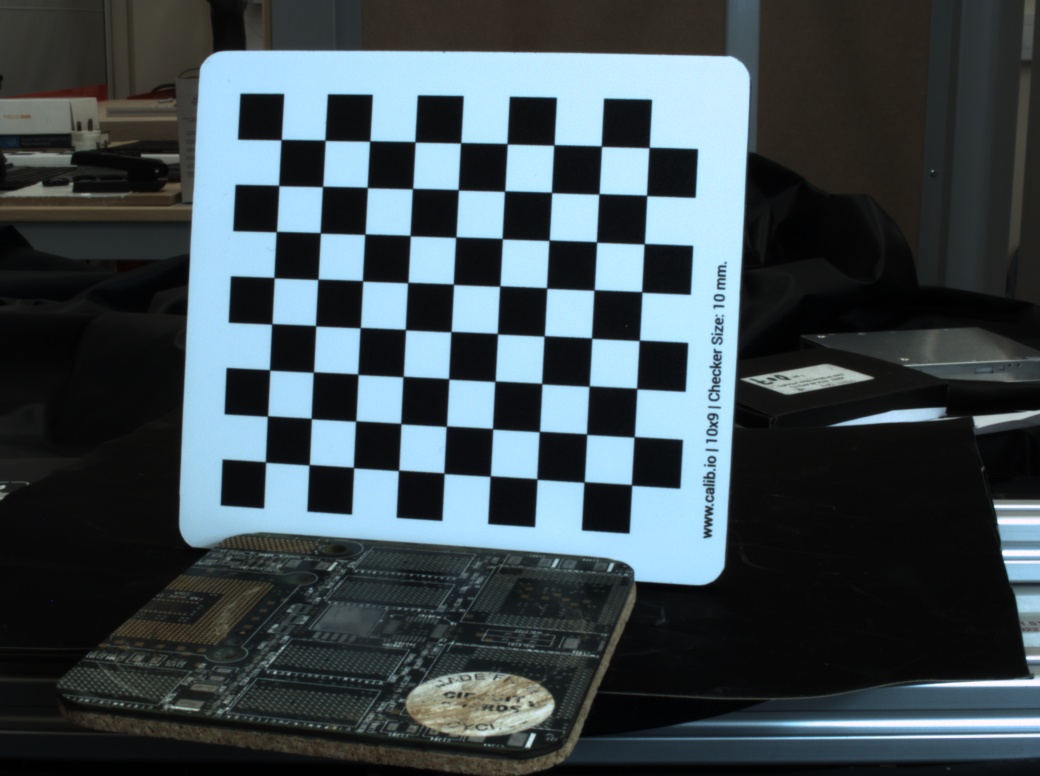}
`\includegraphics[height=0.14\textwidth]{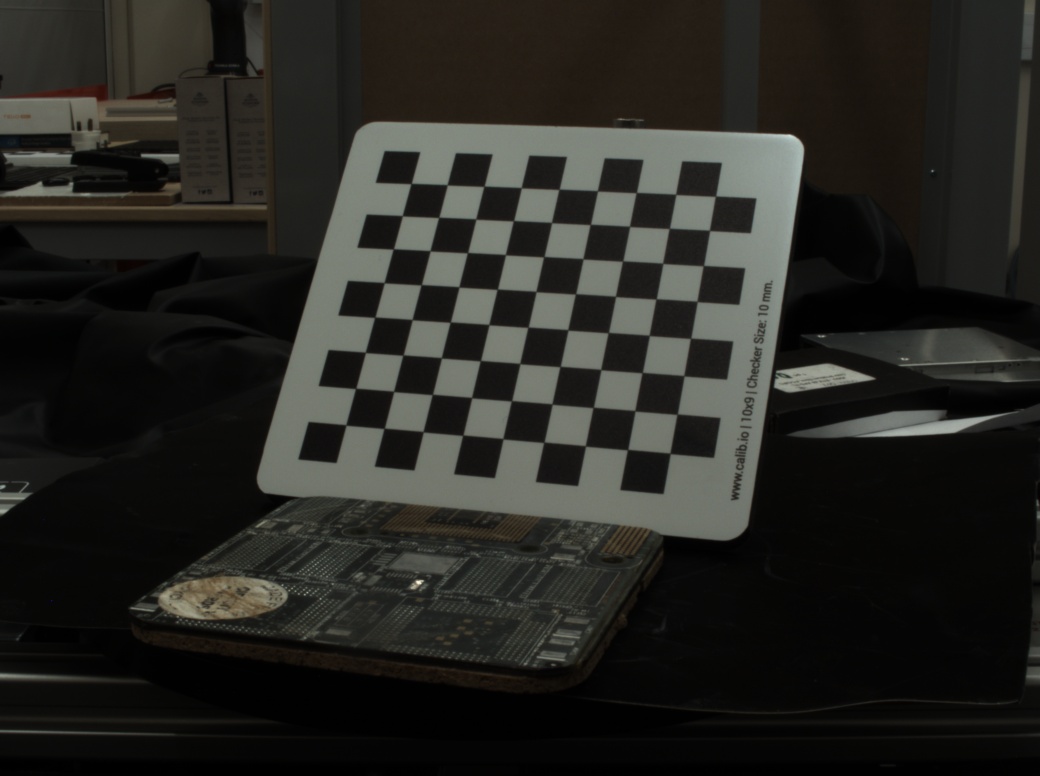}
 \caption{This figure shows all 15 images from 2 out of the 26 views used for calibration. It also shows 5 of 420 checkerboard images. Note that to maximize the image quality of the checkerboard images, additional external illumination was used.}
 \label{fig:calib}
\end{figure*}

Light calibration was performed using the same point-wise differentiable renderer from the original LUCES~\cite{Mecca21luces} dataset. In sort, the flat object is approximate Lambertian, with albedo $0.99$ (according to manufactured specifications) with no shadows and interefection, thus following irradiance Equation 2 of main text. 
390 images (2 cameras, 15 lights, 13 view-points) are captured to ensure orientation and scale variation as shown in Figure~\ref{fig:calib}. These images will be made available as a part of files representing the dataset. We hope that these calibration images will encourage future research into neural light attenuation models/point light radiance fields, that should be more accurate than the SOTA analytic calibrated LED model of \cite{Mecca2014near}.

\section{Additional results}

The section contains additional results not reported in the main paper. In particular, it provides a discussion on why certain objects used in LUCES-MV are significantly more challenging than objects used in DiLiGenT-MV~\cite{LiZWSDT20}. It also visualises results of binocular photometric stereo results of~\cite{Logothetis24WACV}, provides predicted albedo image visualisations of UniPS~\cite{ikehata2023sdmunips} (used by RNb-NeuS~\cite{BrumentRNb24}), lists running times of competing multi-view photometric stereo methods and shows additional results of non-PS reconstruction method, Neuroangelo~\cite{li2023neuralangelo}.

\subsection{Object choices}
The objects in LUCES-MV were chosen to have a varied selection of shapes and materials. \textit{Ball} is made of sponge which is missing in competing datasets.  \textit{Bowl} has an extreme concavity and a specular material hence experiencing challenging shadows and intereflections.  \textit{Buddha} is made of marble which is mostly diffuse but contains high frequency details that are also highly specular. \textit{Bell} and \textit{Cup} are metallic with \textit{Cup} being much shinier (and thus harder) than \textit{Bell}. \textit{Bunny} and \textit{Hippo} are made of shiny plastic but are contrasted as \textit{Bunny}  is more textured than \textit{Hippo}. Indeed, having multiple object of similar material can be help disentangle the effect of geometry vs material on the difficulty. Thus the painted wood \textit{Die} as well as the wooden \textit{Tool} and top of the \textit{Jar} are aimed to compare different wood configurations. The \textit{House} has very complicated geometry but high amount of texture. The \textit{Squirrel} is shiny porcelain (the main body of the \textit{Jar} is also porcelain) with a lot of self reflection in the bottom part. Plaster objects \textit{Owl} and \textit{Queen} are fairly diffuse and thus are aimed to be relatively easy; however, cast shadows at concave regions did pose some challenge to current SOTA methods.

Finally, we also include the semi-transparent \textit{Glass} which is currently unsolvable for most PS and neural reconstruction methods to encourage future research in that directions.

\subsection{Binocular photometric stereo}
\label{sec:stereo_extra}

\begin{figure*}[h]
\centering
\includegraphics[width=\textwidth]{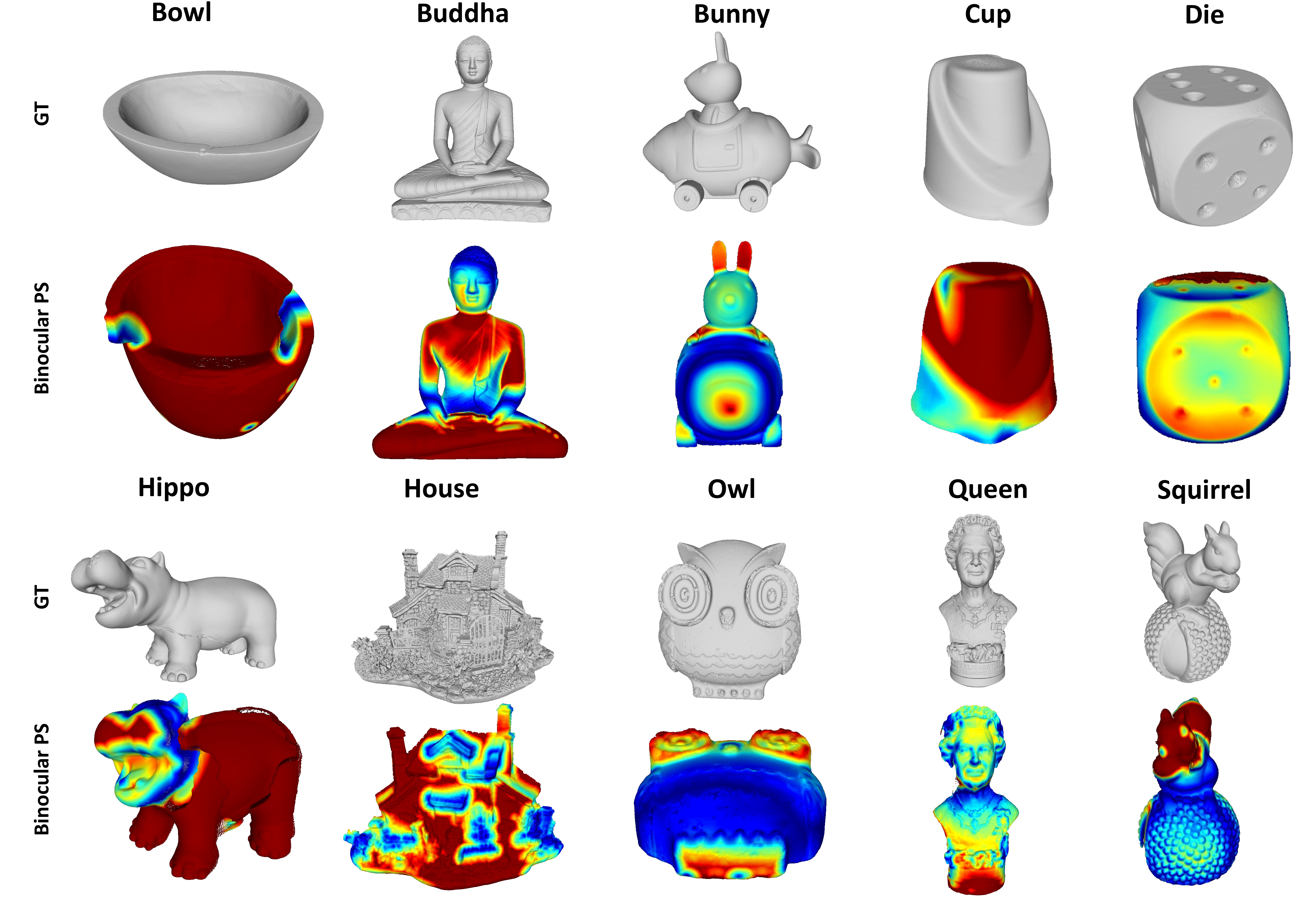}\caption{This figure shows results of binocular photometric stereo of~\cite{Logothetis24WACV}. First row shows the ground truth shape and the second row shows the shape error. Unlike in the similar figure in the main paper here red color corresponds to a 5mm error (not 1mm).}
\label{fig:binocularshaperesults}
\end{figure*}

Quantitative visualization of the results of the binocular photometric stere method \cite{Logothetis24WACV} is shown in Figure~\ref{fig:binocularshaperesults}. It is noted that the recovered shape exhibits significant bending and smoothing of the occlusion boundaries, similar to monocular PS (i.e. \cite{Logothetis22}), and that explains the relatively low quantitative performance. Thus, for applications requiring fast data capture (e.g. grasping point estimation for robotics), additional future research is required to achieve sub-millimeter precision. 

\subsection{Albedo images from Uni-PS~\cite{ikehata2023sdmunips}}

\begin{figure*}[h]
\centering
\includegraphics[width=\textwidth]{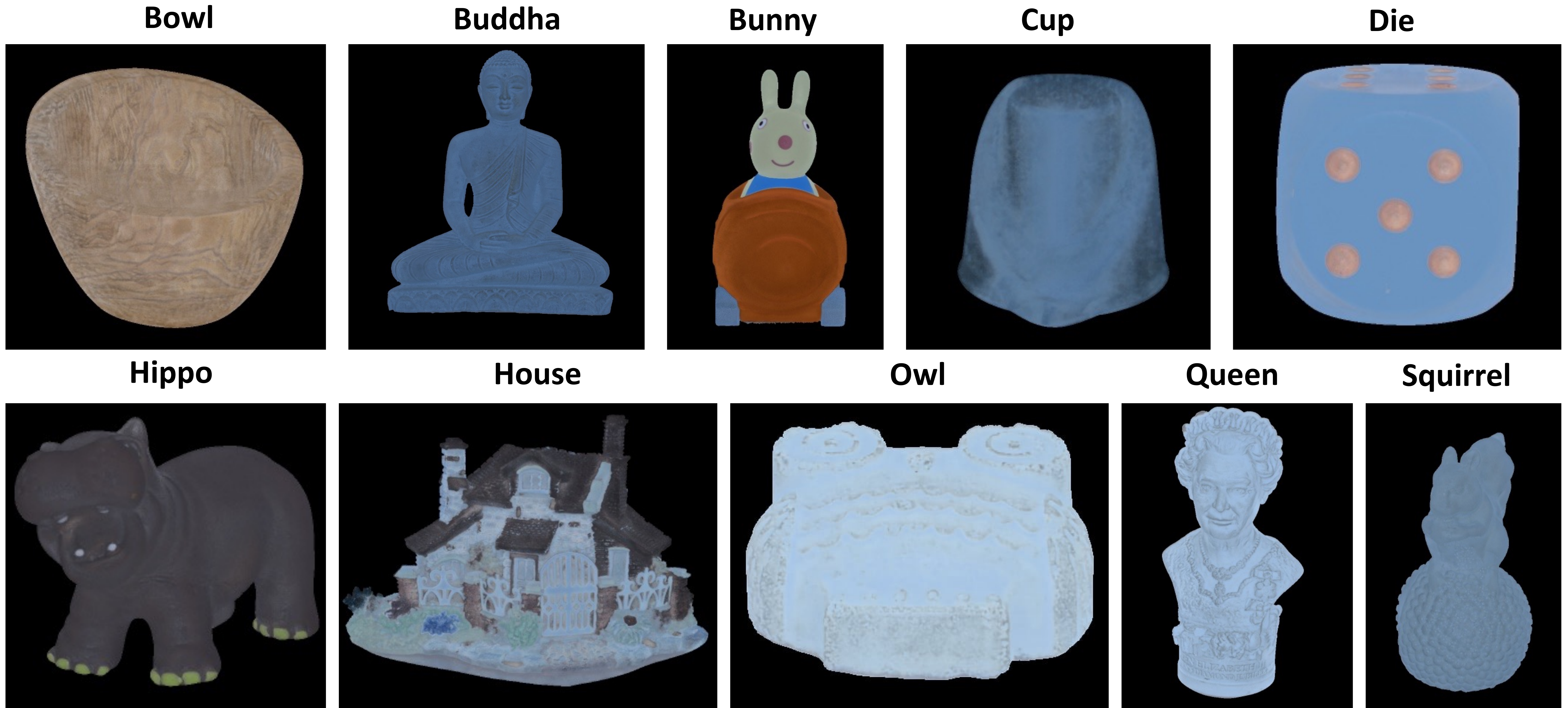}
\caption{Albedo predictions from UniPS~\cite{ikehata2023sdmunips}, used by RNb-NeuS~\cite{BrumentRNb24}.}
\label{fig:albedoimages}
\end{figure*}

Figure~\ref{fig:albedoimages} shows albedo images estimated by Uni-PS~\cite{ikehata2023sdmunips}. These images are used by the SOTA multi-view PS method RNb-NeuS~\cite{BrumentRNb24}. These albedo predictions look reasonable (since there is no easy way to obtain ground truth albedos, the can only be evaluated qualitatively) with notable exceptions the metallic \textit{Cup} and the black numbers on the \textit{Die}. Additionally, concavities on \textit{Buddha} and \textit{Queen} appear darker than the surrounding object, however in reality the real objects should be more uniform.

\subsection{Running times}

Table~\ref{tab:runningresources} shows approximate running times are GPU memory usage for all methods.  Note that RNb-NeuS~\cite{BrumentRNb24}, Supernormal~\cite{cao2023supernormal} and Neuralangelo~\cite{li2023neuralangelo} (using `instant' version, see Section~\ref{sec:Neuroangelo}) are using direct GPU acceleration\footnote{Using CUDNN~\cite{tinycudann} NERFACC~\cite{li2023nerfacc} and other speed-up techniques. Additionally,  RNb-NeuS~\cite{BrumentRNb24} has a full CUDA version. } and thus are significantly faster than Binocular \cite{Logothetis24WACV} and NPLMV-PS~\cite{LogothetisWACV2025} which have pure tensorflow implementations. Additionally, the rendering mode of NPLMV-PS~\cite{LogothetisWACV2025} significantly increases the computational cost as well. Note that all methods were run with the default settings and no effort to optimize their computational time and GPU memory usage.



\begin{table*}[t]
\begin{center}
\resizebox{1.0\textwidth}{!}{%
\begin{tabular}{ | c |c c c c c c c c c |} 
 \hline
Method  & UniPS~\cite{ikehata2023sdmunips} & Uni MS-PS~\cite{hardyunips} & NF-PX-Net~\cite{Logothetis22} & Binocular \cite{Logothetis24WACV} &  Neuralangelo~\cite{li2023neuralangelo} & RNb-NeuS~\cite{BrumentRNb24} & Supernormal~\cite{cao2023supernormal} & NPLMV-PS(N)~\cite{LogothetisWACV2025} &   NPLMV-PS~\cite{LogothetisWACV2025} (N+I)\\  \hline
Time & 2m & 20m & 5m & 5h & 40m & 5m & 5m & 3h & 8h \\  \hline
Memory & 20GB & 3GB & 1GB & 10GB & 10GB & 15GB & 5GB & 10GB & 25GB \\  \hline
\end{tabular}
} 
\end{center}
\caption{Approximate running times and GPU memory usage on single NVIDIA RTX A6000. Note that single view methods UniPS~\cite{ikehata2023sdmunips}, Uni MS-PS~\cite{hardyunips} and NF-PX-Net~\cite{Logothetis22} are reported per single view and are significantly varied depending on the number of foreground pixels; the reported numbers are an approximate overall average.}
\label{tab:runningresources}
\end{table*}

\subsection{Additional results of Neuroangelo~\cite{li2023neuralangelo}}
\label{sec:Neuroangelo}

\begin{figure*}[h]
\centering
\includegraphics[width=\textwidth]{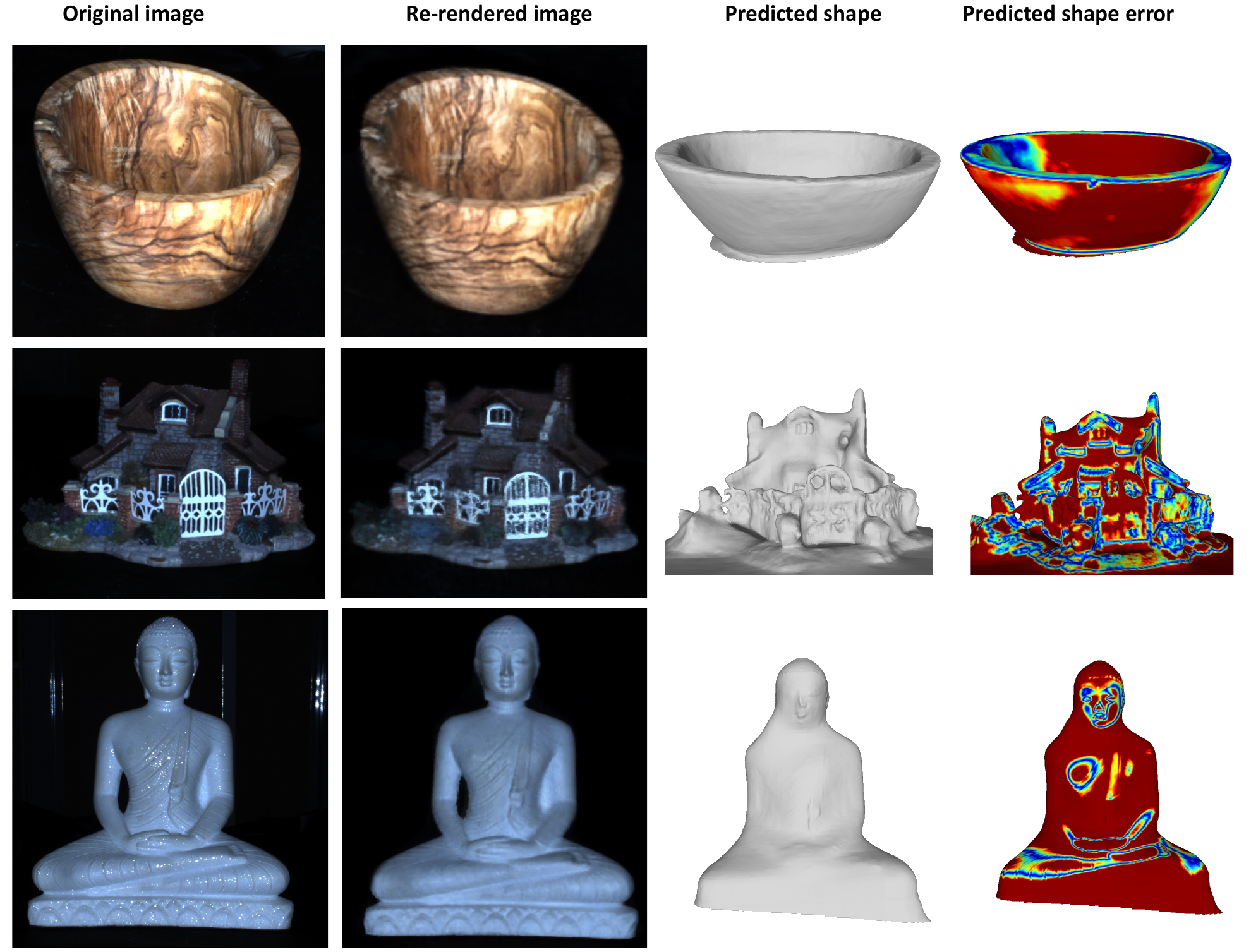}
\caption{Additional results of classical (non-PS) NeRF method 
of Neuralangelo~\cite{li2023neuralangelo}. Red color corresponds to a 1mm error.}
\label{fig:additionalneuralangelo}
\end{figure*}

Figure~\ref{fig:additionalneuralangelo} shows additional results (on objects \textit{Bowl}, \textit{Buddha}, \textit{House}) of non-PS reconstruction method, Neuroangelo~\cite{li2023neuralangelo}. Note, that to perform reconstruction, average PS images from all 36$\times$2 viewpoints and poses and sparse point cloud estimated by SFM library Colmap were used. For other objects pose estimation failed. This is not meant to be a fair comparison with the PS competitors (that use less views, more images per view, segmentation masks but not sparse point cloud) but rather a demonstration that non-PS reconstruction methods, despite their high re-rendering quality, their estimated shape is clearly worse than one estimated by multi-view photometric stere methods. 

It should be noted though that we used the fast version from \url{https://github.com/hugoycj/Instant-angelo/} which may be potentially worse than the original one with several orders of magnitude of computational speed-up.

\section{Limitations}

Since this is a dataset paper the limitations are mainly to do with the size of the dataset. Ideally, it would be scaled to 100s and 1000s of objects as a future work. It also has materials which are generally smooth and do not contain micro structures (e.g. fabrics). While the methods evaluated are SOTA for their respective tasks, we hope that this dataset will be used to evaluated more of the past and future works.

\end{document}